\documentclass[mnsc]{informs3b} 

\OneAndAHalfSpacedXI 

\usepackage{natbib}
 \bibpunct[, ]{(}{)}{,}{a}{}{,}%
 \def\bibfont{\small}%
\usepackage{booktabs, caption, makecell}

\usepackage{threeparttable}
\usepackage{subcaption}

\usepackage[resetlabels]{multibib}
\newcites{sec}{References}

\TheoremsNumberedThrough     
\ECRepeatTheorems

\EquationsNumberedThrough    

\usepackage{amsmath}

\usepackage[bottom]{footmisc}
\usepackage{afterpage}

\usepackage{xcolor}

\usepackage[bookmarks, hidelinks]{hyperref}
\hypersetup{
	bookmarksnumbered=true,     
	bookmarksopen=true,
	bookmarksopenlevel=2,    
	colorlinks=true,
	linkcolor={red!50!black},
	citecolor={blue!50!black},
	urlcolor={blue!80!black},
}

\usepackage{bookmark}
\usepackage{etoolbox}
\usepackage{tocloft}
\makeatletter
\pretocmd\endfigure{%
\addtocontents{lof}{\protect{%
    \bookmark[
    rellevel=1,
    keeplevel,
    dest=\@currentHref,
    ]{Figure \thefigure: \@currentlabelname}}}%
\bookmark[
rellevel=1,
keeplevel,
dest=\@currentHref,
]{Figure \thefigure: \@currentlabelname}%
}{}{\errmessage{Patching \noexpand\endfigure failed}}
\pretocmd\endtable{%
\addtocontents{lof}{\protect{%
    \bookmark[
    rellevel=1,
    keeplevel,
    dest=\@currentHref,
    ]{Table \thetable: \@currentlabelname}}}%
\bookmark[
rellevel=1,
keeplevel,
dest=\@currentHref,
]{Table \thetable: \@currentlabelname}%
}{}{\errmessage{Patching \noexpand\endtable failed}}
\makeatother

\usepackage{tabularx}
\usepackage{lscape}

\usepackage{cases}
\usepackage{threeparttable}

\usepackage{comment}

\usepackage{algorithm} 
\usepackage[noend]{algpseudocode} 

\usepackage{blkarray}
\usepackage{bigstrut}

\usepackage[toc]{appendix}

\usepackage[nomessages]{fp}

\usepackage{enumitem}

\DeclareMathOperator{\dplus}{+\kern -0.4em+}

\usepackage{float}
\usepackage{graphicx}

\begin{document}

\newcolumntype{L}[1]{>{\raggedright\arraybackslash}p{#1}}
\newcolumntype{C}[1]{>{\centering\arraybackslash}p{#1}}
\newcolumntype{R}[1]{>{\raggedleft\arraybackslash}p{#1}}



\RUNTITLE{Care for the Mind Amid Chronic Diseases}


\TITLE{Care for the Mind Amid Chronic Diseases: \\An Interpretable AI Approach Using IoT}




\begin{center}
\textbf{\Large Care for the Mind Amid Chronic Diseases: An Interpretable AI Approach Using IoT}    
\end{center}

\hspace{0.2cm}
\begin{center}

Jiaheng Xie$^{1,**}$, Xiaohang Zhao$^{2,*,**}$, Xiang Liu$^{1}$, Xiao Fang$^{1}$ \\
\vspace{0.2cm}
$^1$ Lerner College of Business and Economics, \\
University of Delaware, Newark, DE, USA \\

$^2$ School of Information Management \& Engineering, \\ Shanghai University of Finance and Economics, Shanghai, China \\

\vspace{0.1cm}
$*$ Corresponding Author: Xiaohang Zhao, \href{mailto:xiaohangzhao@mail.shufe.edu.cn}{xiaohangzhao@mail.shufe.edu.cn} \\
$**$ Equal Contribution \\
\end{center}
\vspace{0.3cm}

\noindent \textbf{Abstract:} Health sensing for chronic disease management creates immense benefits for social welfare. Existing health sensing studies primarily focus on the prediction of physical chronic diseases. Depression, a widespread complication of chronic diseases, is however understudied. 
We draw on the medical literature to support depression detection using motion sensor data. 
To connect humans in this decision-making, safeguard trust, and ensure algorithm transparency, we develop an interpretable deep learning model: \underline{Temp}oral \underline{P}rototype \underline{Net}work (TempPNet). 
TempPNet is built upon the emergent prototype learning models.
To accommodate the temporal characteristic of sensor data and the progressive property of depression, TempPNet differs from existing prototype learning models in its capability of capturing temporal progressions of prototypes. 
Extensive empirical analyses using real-world motion sensor data show that TempPNet outperforms state-of-the-art benchmarks in depression detection. Moreover, TempPNet interprets its decision by visualizing the temporal progression of depression and its corresponding symptoms detected from sensor data. 
We further employ a user study and a medical expert panel to
demonstrate its superiority over the benchmarks in interpretability.
This study offers an algorithmic solution for impactful social good — collaborative care of chronic diseases and depression in health sensing. Methodologically, it contributes to extant literature with a novel interpretable deep learning model for depression detection from sensor data.
Patients, doctors, and caregivers can deploy our model on mobile devices to monitor patients’ depression risks in real-time. Our model's interpretability also allows human experts to participate in the decision-making by reviewing the interpretation and making informed interventions. 

\hspace{0.1cm} 


\hspace{1cm}






%


\section{Introduction}
\label{sec:intro}

The Fourth Industrial Revolution is characterized by technological advancements in Artificial Intelligence (AI) and big data analytics \citep{shim_2022_transformative}. Along with its emergence and progress, numerous IT artifacts are implemented by organizations worldwide to modernize their services, scale up their businesses, or improve the efficiency of data exchange. Meanwhile, these IT artifacts record massive amounts of data, which describe the context and outcomes of users’ actions and form unique digital traces for each user \citep{hedman_2013_digital}. As part of this trend, digital traces of wearable sensor signals represent an immense and novel source of ecological data that reflect human behaviors and psychological characteristics \citep{chau_2020_finding}. The collection of such data opens up significant research opportunities for Information Systems (IS) researchers to develop innovative artifacts aimed at addressing critical societal and healthcare challenges \citep{zhang_2020_comprehensive}.

Chronic disease management is a significant societal challenge that can be addressed with novel IT artifacts. The attention to this area in the IS literature is growing rapidly due to its profound social and economic impacts. In the US alone, 133 million Americans suffer from one or more chronic diseases \citep{forbes_2022_our}, of which the most common ones are heart diseases, cancer, Alzheimer’s, and diabetes \citep{cdc_chronic_2022}. The population of chronic disease patients in the US is expected to reach over 143 million by 2050 \citep{ansah2023projecting}. Moreover, treatments of chronic diseases account for 90\% of the US's \$3.8 trillion annual healthcare cost \citep{cdc_us_2021}. 

While treatments of chronic diseases attract the primary research efforts in health sensing, care for depression associated with chronic diseases has not received due attention, although medical experts have repeatedly stressed the urgency of collaborative care for depression and chronic diseases \citep{katon_collaborative_2010}. Depression is epidemic among chronic disease patients, caused by frustrating long-haul symptoms and complex home regimens. Evidence shows that depression occurs in 17\% of cardiovascular cases, 23\% of cerebrovascular cases, 27\% of diabetes patients, and 40\% of cancer patients \citep{cdc_mental_2012}. According to the National Institutes of Health (NIH), chronic disease patients are twice as likely to suffer from depression as the general population, and depression could drastically elevate chronic disease severity \citep{nih_chronic_2022}. Therefore, mental care amid chronic diseases is essential. Accordingly, our first objective is to \textit{detect the occurrence of depression for chronic disease patients using sensing technologies}. When signs of depression are detected, our model could alert caregivers and doctors to actively intervene and treat chronic disease patients’ mental disorders.

A key challenge of using sensing technologies to detect depression is that motion sensors only capture patients’ physical movements, while depression is a disorder of mental activities. Nevertheless, medical studies have shown that depressed patients experience various physical symptoms, such as slow movement or speech, unexplained aches and pains, and lack of energy \citep{nhs_symptoms_2022, sloman_gait_1982,lemke_spatiotemporal_2000}. Such physical symptoms of depression are embodied in walking patterns that can be captured by motion sensors. The association between physical movements and depression supports our detection of depression from motion sensor data.

Due to the high stake and impact of health analytics outcomes, a black-box model is inadequate. Interpretability is essential to increase trust in a machine learning (ML) model, prevent failures, and justify its usage \citep{moss_demystifying_2022}. Moreover, an interpretable model allows medical experts to understand the reasoning behind the decision and make professional diagnoses and accurate interventions. Therefore, our second objective is to \textit{build an interpretable model for depression detection}. As mentioned above, the medical literature finds that depressed patients present typical walking symptoms \citep{nhs_symptoms_2022}. Therefore, these walking symptoms discovered from sensor data serve as a natural interpretation mechanism for the detection of depression.
Discovering prototypical patterns from the input to interpret ML predictions forms an active research forefront: prototype learning \citep{chen_this_2019,ming_interpretable_2019}. However, existing prototype learning methods are inadequate for our study because of the following challenges. 
First, existing methods predict and interpret from a single object (e.g., an image). In contrast, we need to design a method that can predict and interpret from a time series of multivariate time series sensor data, a data structure beyond the scope of existing time series prototype learning methods \citep{gee_2019_explaining}. Second, the severity of depression symptoms evolves over time, such as trending up, peaking, and fluctuating \citep{dattani_mental_2021}. Static prototypes defined by existing prototype learning methods cannot capture such dynamic and temporal patterns of depression symptom evolution.  
Accordingly, we need to define a novel temporal prototype that can capture such dynamic and temporal progressions of prototypes and design a new prototype learning method that can learn temporal prototypes from motion sensor data.

This study targets the interpretable detection of depression associated with chronic diseases and makes the following contributions. 
From a managerial perspective, our study is positioned in the health information technology (HIT) research area in IS. 
We extend sensing IS research with a novel AI-based solution and validate the potential of sensing technologies for depression management. 
Motivated by the unique challenges at the intersection of ML and its societal environments, our study develops a novel computational method to solve a critical societal problem (i.e. depression detection).
From a methodological perspective, we propose a novel interpretable deep learning method that interprets the detection of depression by learning temporal progressions of prototypes from sensor data.
Different from state-of-the-art prototype learning methods, our method takes a time series of multivariate time series sensor data as inputs and innovatively learns two types of prototypes from the data: prototypes of depression symptoms and prototypes of temporal symptom progression. It then detects the depression status of a chronic disease patient and interprets the decision based on these two types of prototypes. Extensive empirical analyses using real-world motion sensor data demonstrate the superior detection performance of our method over state-of-the-art benchmarks. Through a user study and a medical expert panel, we also show that our method outperforms the benchmark in interpretability. 
From a practical perspective, our proposed method can be implemented as a mobile app or be embedded as a function in existing mobile health apps. It is able to detect and monitor chronic disease patients’ depression risk in real-time based on motion sensor data collected by their mobile devices. Once our method detects early signs of depression, it can send alerts to caregivers and doctors to provide timely treatments, such as antidepressants and social support groups.

\section{Literature Review} \label{sec:lr}

\subsection{Health IS and Chronic Disease Management}\label{sec:lr:his}

To mitigate the consequences of chronic diseases, IS scholars, health organizations, and IT companies have been cultivating interdisciplinary intelligent chronic disease management \citep{dixon-woods_improving_2013}, or HIT. In the past decade, HIT for chronic disease management has evolved from hospital-based electronic health records (EHR) to patient-centric mobile devices and assistive technologies, while the analytics methods have expanded from statistical and econometric models to advanced computational models \citep{bardhan_2020_connecting}. These computational HIT studies in IS can be broadly categorized into three areas. First, the majority of IS studies use hospital-based information systems such as EHR, with application areas spanning adverse events prediction \citep{lin_2021_first}, hospital readmission \citep{xie_2021_readmission}, among others. The second area touches on analytics on online health communities and health social media \citep{zhang2024ketch}. These HIT studies aim to utilize online information technologies to connect patients, caregivers, and health professionals, with topics ranging from patient education \citep{liu_2020_go} and drug safety surveillance \citep{xie_2021_unveiling}, to medication nonadherence \citep{xie_understanding_2022}.

The last and emerging health IS research for chronic disease management is related to the increasingly popular mobile apps and wearable sensors \citep{yu_wearable_2022}. Despite the active research in the first two categories, mobile health systems design and analytics remain under-explored. Yet, it is an increasingly critical area. 
``\textit{Wearable sensors and home devices can play a pivotal role not only in monitoring the status of chronic disease patients and predicting adverse events before they occur but also in preventing the onset of chronic diseases}'' \citep{bardhan_2020_connecting}. Such resulting predictive models can generate fruitful managerial implications such as motivating patients to adhere to treatment protocols or individuals to lead a healthy lifestyle \citep{liu2024value}. 

In recent years, a few wearable sensor-based HIT studies have begun to emerge in premier IS journals. For instance, 
\cite{yu_wearable_2022} design an attention network for chronic disease management using sensor signals collected via mobile apps. \cite{zhu_2020_human} and \cite{zhu_deep_2021} deploy sensing technologies to the senior care setting where deep learning models can recognize activities of daily living.
An MIS Quarterly editorial refers to this type of research as the Type I ML research in IS, where the \textit{``designed IT artifacts are models and algorithms''} and \textit{``the methodological contributions of Type I ML research tend to be novel ML models and algorithms developed to solve important business and societal problems''} \citep{padmanabhan_machine_2022}. The Information Systems department within Management Science also welcomes this type of research and states it \textit{``combine(s) a methodological advance with an important and novel managerial application''} \citep{simchi_2020_editor}.

\subsection{Wearable Sensor Technology and Depression}\label{sec:lr:hsd}

This study belongs to health IS research that collects high-fidelity, real-time sensing data for intelligent chronic disease management. Studies in this area typically recruit 12 to 683 patients to conduct walking tests \citep{coelln_quantitative_2019}, during which a participant is instructed to walk for a certain distance while his or her walking signals are recorded by motion sensors. 
Appendix \ref{apd:health_sensing_lit} summarizes recent health sensing studies related to our study.
While the detection of chronic diseases, such as Parkinson's disease (PD) and diabetes, deserves research attention in the health sensing discipline, mental care is equally critical to chronic care. This is because irritating physical symptoms and tedious long-term disease management cause serious mental complications, leading to depression among 40-50\% of chronic disease patients \citep{reijnders_systematic_2008}. 
If left untreated, depression's impact could lead to greater functional disability, faster physical and cognitive deterioration, increased mortality, poorer quality of life, and increased caregiver distress \citep{marsh_depression_2013}. In this study, \textit{we aim to detect the occurrence of depression for chronic disease patients using motion sensor data.}

Medical studies have provided abundant evidence that motion symptoms are an essential manifestation of depression. 
Compared to healthy people, depressed patients exhibit the following patters: slower walk, slower lifting motion of the leg \citep{sloman_gait_1982}, shorter strides, slower gait velocity \citep{lemke_spatiotemporal_2000}, reduced vertical head movements, more slumped posture, and lower gait velocity \citep{michalak_embodiment_2009}.
Clinical studies have also proven that these physical depression symptoms can be reliably captured via self-conducted walking tests \citep{vancampfort_testretest_2020}. This is because, rather than temporary mood swings, major depressive disorder is persistent and causes significant impairment in daily life \citep{mayo_clinic_depression_2022}. For such a persistent disorder, the above-mentioned walking symptoms will be present in walking tests.

Related to this study, several prior works also recognize the opportunity of detecting depression based on mobility data. \cite{canzian_trajectories_2015} use GPS mobility traces to detect a patient's depressive state using SVM. \cite{jacobson_passive_2020} conduct a study of 31 patients and employ XGBoost to detect their depressive status based on sensor data collected from these patients. \cite{farhan_behavior_2016} recruit 79 participants for sensor data collection and detect their occurrence of depression using regression. Our study differs from these studies in the following aspects. First, these studies employ conventional ML methods, such as SVM, XGBoost, and regression, which have been shown to be less effective in various applications compared to deep learning models \citep{xie_understanding_2022,yu_wearable_2022,zhu_deep_2021}. Due to the high impact of this area and the rapid advances in deep learning, the computational methods for sensor-based depression detection are due a much-needed, timely revisit and upgrade. Second, because of their use of conventional ML methods, these studies have to craft features from sensor data. Such manual design and selection of features are labor-intensive, require significant domain knowledge, and could lead to inconclusive results \citep{hubble_wearable_2015,yu_wearable_2022}. Third, the intention of these models is to make predictions, not interpretations. Therefore, they are either black-box models or models that explain predictions based on simple, hand-crafted features from sensor data, such as mean and variance. These features are primitive and not relevant to depression, thus falling short of offering a meaningful and practical interpretation. In order for end users (e.g., doctors, patients, and caregivers) to trust and adopt a prediction model, meaningful interpretation is the key. 
As our society is progressing toward integration with ML, new regulations have been imposed to require verifiability, accountability, and more importantly, full transparency of algorithm decisions. A key example is the European General Data Protection Regulation (GDPR), which requires companies to provide data subjects the right to an explanation of algorithm decisions. Consequently, \textit{we aim to propose a novel interpretable deep learning model for depression detection}.

\subsection{Interpretable Machine Learning and Prototype Learning}\label{sec:lr:iml}

The majority of interpretable ML methods offer feature-based interpretation, where hand-crafted features are required, and their interpretation reveals the attribution of each feature to the prediction. These methods can be post-hoc, such as SHAP and LIME \citep{lundberg_unified_2017, kim_2023_rolex}, or model-based, such as generative additive model (GAM) \citep{caruana_intelligible_2015} and wide and deep learning (W\&D) \citep{xie_unbox_2020}. 
A related IS study is \cite{kim_2023_rolex}, who develops an interpretable method based on LIME for the predictions of heart disease, cancer, and fracture. 
While their study and ours fall into the same field and application domain, our contributions are significantly different from theirs. The method proposed by \cite{kim_2023_rolex} is a post-hoc interpretable method that predicts diseases with a predictive model and then interprets its prediction with a separate interpretation model. Our method, on the other hand, is not a post-hoc method. It builds the interpretable module into the predictive model and makes predictions and interpretations at the same time. In addition, \cite{kim_2023_rolex} use structured clinical data and offer feature-based interpretation. Yet, feature-based interpretations are not desirable in sensor-based depression detection studies. This is because, first,
although a few studies crafted simple sensor-based features — such as mean, variance, and standard deviation of a segment of signals \citep{oung_wearable_2015} — they do not characterize the symptoms of depression. Second, manually engineering meaningful features from sensor signals requires intensive labor and could even result in inconclusive results \citep{hubble_wearable_2015,yu_wearable_2022}. Such a practice also demands rich domain knowledge, which is not easily accessible \citep{yu_wearable_2022}. Fortunately, medical literature reveals that depressed patients present typical walking symptoms \citep{sloman_gait_1982,lemke_spatiotemporal_2000,nhs_symptoms_2022}. Without feature engineering, these walking symptoms can be fruitfully leveraged to interpret the detection of depression. For that purpose, an emergent line of interpretable ML research, prototype learning, has been proposed to utilize the prototypical part of a class (prototype) to interpret the prediction. For instance, a prototypical depression walking symptom learned from sensor signals can interpret why a patient is classified as depressed.

Prototype learning was originally proposed to interpret image recognition \citep{chen_this_2019}. When interpreting how to classify an image, one focuses on parts of the image and compares them with prototypical images from a given class. For instance, radiologists compare suspected tumors in X-rays with prototypical tumor images for the diagnosis of cancer \citep{chen_this_2019}. The resemblance of an input image to a prototypical class is the mechanism to interpret the classification of the image. 
Another example is that when humans describe why a bird picture is classified as a clay-colored sparrow, we might reason that the bird’s head and wing bars look like those of a prototypical clay-colored sparrow. Leveraging such a reasoning mechanism, \cite{chen_this_2019} devise ProtoPNet. 
To classify a bird picture into a particular bird species with interpretation, this model introduces a prototype layer where $K$ prototypes are assigned to each known species. Each of these prototypes is intended to capture the prototypical parts, or the most salient and typical representation, of the corresponding bird species, such as the head of a clay-colored sparrow or the wing of a cardinal. ProtoPNet embeds each prototype $j$ as a vector $p_j$ in the latent space, and defines \textit{prototype similarity} $s_j$ to measure how strongly prototype $j$ exists in the input bird picture by comparing $p_j$ and the feature maps (extracted via convolutional layers) of the picture in the latent space. The model subsequently classifies the input bird picture based on the weighted sum of the prototype similarities computed between this picture and each prototype. 
ProtoPNet visualizes each prototype as the most relevant region of the bird picture where the prototype most strongly exists, and interprets why it thinks the input bird picture should be classified as a particular species by identifying several parts from the picture that look like the prototypical parts (prototypes) of the species.

Other prototype learning models have also emerged, such as text-based prototype learning \citep{ming_interpretable_2019}, tree prototype learning \citep{nauta_neural_2021}, and more. Appendix \ref{apd:prototype_learning_lit} contrasts major prototype learning studies with our method.
Existing prototype learning methods interpret classifications of static objects, such as images and texts. Accordingly, these methods learn static prototypes (e.g., segments of images) and employ the learned static prototypes to interpret classifications. Therefore, they fail to model temporal progressions of events, which are critical for designing interpretable depression detection methods with wearable sensor data \citep{bockting_lifetime_2015,dattani_mental_2021}.

\subsection{MTSC, Prototype Learning for MTSC, and Time Series Deep Learning}\label{sec:lr:mtsc}

To account for time series data, the most closely related problem to ours is multivariate time series classification (MTSC), which aims to leverage the measurement of multiple variables over a period of time to assign data points to classes. MTSC has been used to solve classification problems in human activity recognition, diagnosis using electrocardiogram (ECG), electroencephalogram (EEG), magnetoencephalography (MEG), and systems monitoring \citep{ruiz_2021_great}. The existing MTSC models fall into five categories: distance measures, shapelets, histograms over a dictionary, interval summarising, and deep learning \citep{ruiz_2021_great}, which are summarized in Appendix \ref{apd:mtsc_lit}.

\begin{figure}[h]
    \centering
    \includegraphics[width=1\textwidth]{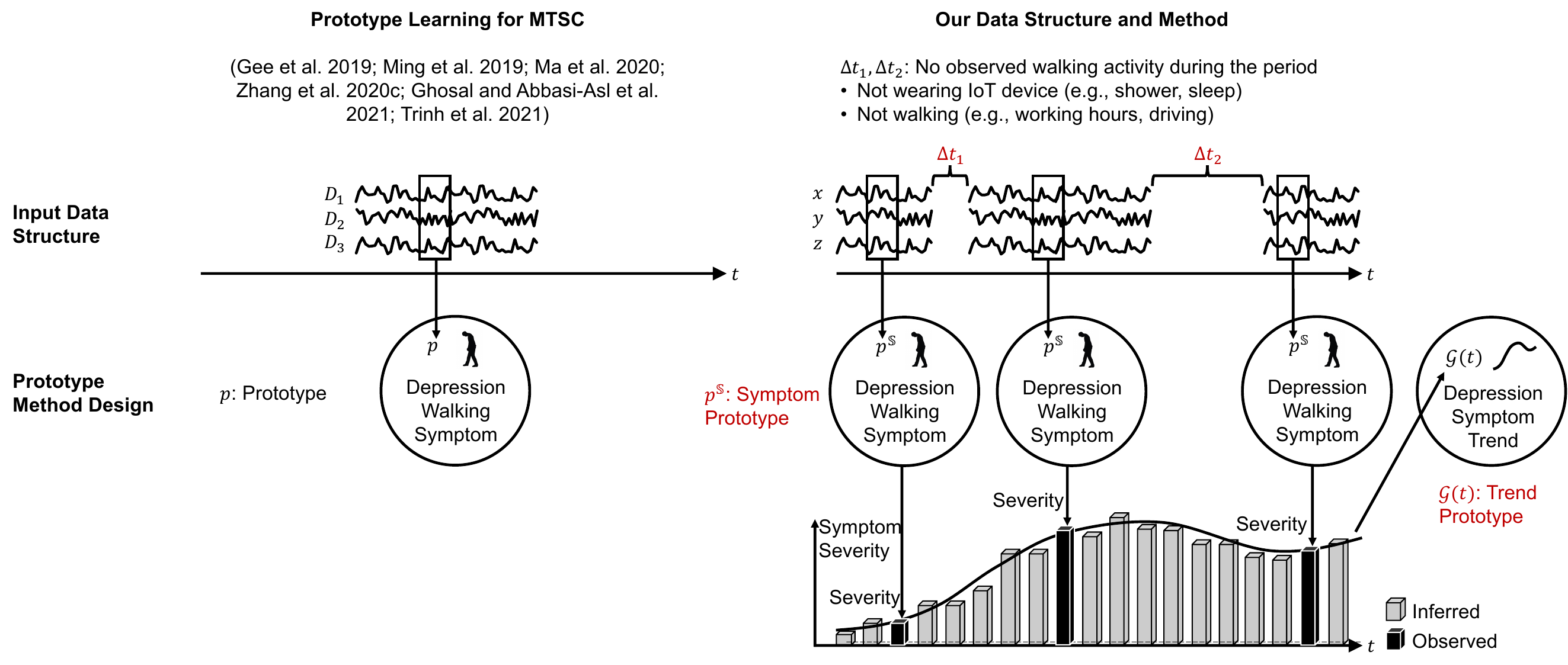}
    \caption{Our Method v.s. Prototype Learning for MTSC}
    \label{fig:our_vs_mtsc}
\end{figure}

Our method differs from existing MTSC models in terms of input data structure, model design, and model interpretability. From the perspective of input data structure, MTSC models classify multivariate time series data. Each input instance is a multivariate time series with $t$ observations: $X_k=\langle X_{1_k},X_{2_k},\dots,X_{t_k} \rangle$, where $X_{i_k}$ is a vector of $d$ dimensions for $i=1,2,\dots,t$ \citep{ruiz_2021_great}. Our study, on the other hand, aims to classify a time series of multivariate time series. That is, input data in our study is a time series of walking segments, each of which is a multivariate time series. Moreover, these walking segments are irregularly spaced in time. Figure \ref{fig:our_vs_mtsc} depicts a visualization of such differences.
Our input data structure, yet challenging and beyond the capability of existing MTSC methods, is fairly common in wearable sensor-based physical activity monitoring for two reasons. First, most studies conduct minutes-long walking tests for participants, and these walking tests are performed at arbitrary time points. Second, for home monitoring settings, participants do not wear the mobile device from time to time, e.g., not wearing it during showers and sleep. Also, they would not walk constantly. For instance, the time segments of sitting in front of the computer during working hours are not candidate data points. Time segments when driving are not useful either. Arbitrarily connecting the walking segments to a single multivariate time series is not feasible either, because this will distort the actual progression of health status over time. For example, depression progresses gradually even though some time segments are not observed. Directly connecting the observed segments will create an illusion of the depression status jump or shift instantly. 

To accommodate our data structure, we propose the following methodological novelties, in comparison to existing MTSC methods.
Our method first detects a series of discrete symptom severities from an irregularly spaced time series of multivariate time series walking segments. We then learn the underlying continuous temporal progressions of symptom severities by treating the detected symptom severities as discrete observations sampled from a carefully designed continuous temporal distribution.
The deep learning-based MTSC studies also suffer from setbacks in our context — i.e. from the interpretability perspective, they are black-box models, which cannot offer interpretable insights for health providers.
To promote interpretability in MTSC, recent studies resort to prototype learning. For instance, \cite{gee_2019_explaining} use ECG data to detect clinical bradycardia. ECG data are multivariate time series, which are treated as 2-D images, where the horizontal dimension represents the time. Then an autoencoder layer and a prototype layer are added for the classification and interpretation. The learned prototypes are snippets of the constructed image that shows which segment of the ECG data indicates typical bradycardia.
There are multiple similar studies in recent years.
Appendix \ref{apd:proto_mtsc_lit} compares our method with these MTSC studies in prototype learning.

From the perspective of input data structure, prototype learning methods for MTSC follow the studies in Appendix \ref{apd:mtsc_lit} that can only process a multivariate time series, such as ECG \citep{gee_2019_explaining, zhang_2020_tapnet}, vital signs \citep{ma_2020_interpretable}, and videos \citep{trinh_2021_interpretable}. However, our data is a time series of multivariate time series. These multivariate time series are irregularly spaced in time. As discussed in the MTSC section above, to accommodate this data structure, we propose novel method designs. 

\begin{figure}[h]
    \centering
    \includegraphics[width=0.6\textwidth]{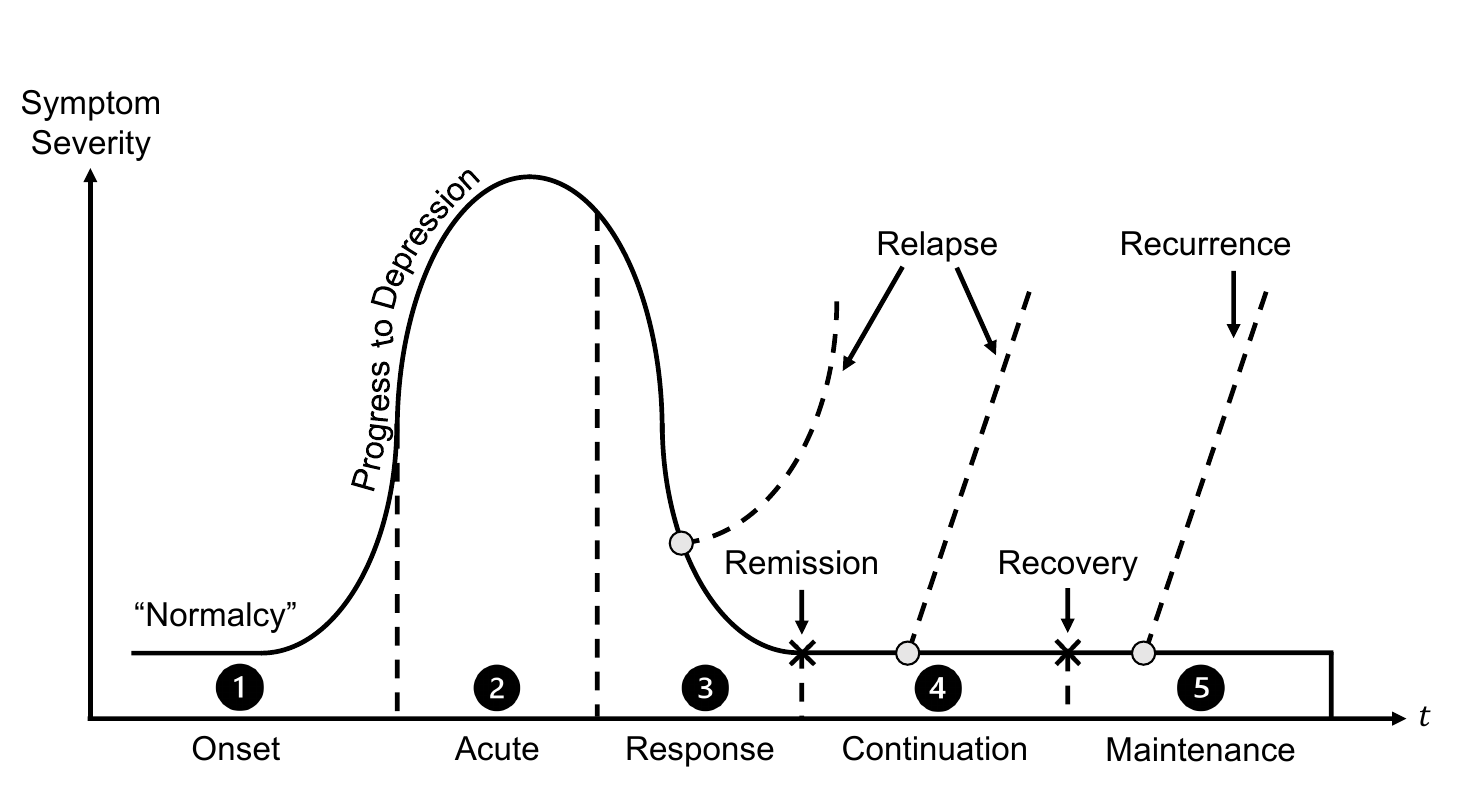}
    \caption{Temporal Progression of Depression \citep{bockting_lifetime_2015}}
    \label{fig:depression_trend}
\end{figure}

From the perspective of prototype method design, prototype learning methods for MTSC share a common pitfall of employing static prototypes to interpret classifications \citep{gee_2019_explaining,ming_interpretable_2019,ma_2020_interpretable,zhang_2020_tapnet,ghosal_2021_multi,trinh_2021_interpretable}, while our method utilizes our proposed dynamic prototypes to interpret depression detection. More concretely, in Section \ref{sec:method:trend}, we propose dynamic prototypes, namely trend prototypes, to capture temporal progressions of symptoms. For example, the method proposed by \cite{gee_2019_explaining} is a representative prototype learning method for MTSC. Specifically,  
\cite{gee_2019_explaining} adopt the prototype learning method for image classification \citep{chen_this_2019} by treating ECG data as static images. As discussed at the end of Section \ref{sec:lr:iml}, the prototype learning method proposed by \cite{chen_this_2019} employs static prototypes. In particular, the prototype in \cite{gee_2019_explaining} is a segment of ECG signals.
In depression detection, it is necessary to model depression symptoms and temporal progressions of depression symptoms \citep{dattani_mental_2021}. While depression symptoms can be modeled as static prototypes, temporal progressions of these symptoms need to be captured with dynamic prototypes.
This temporal dimension of our method's learned prototypes has practical implications because
medical literature suggests that depression symptoms are progressive and have multiple phases \citep{ bockting_lifetime_2015,dattani_mental_2021}. As depicted in Figure \ref{fig:depression_trend}, from the onset phase, a patient's symptom severity may trend up and progress to acute depression. After the acute phase, some patients' symptoms may peak, while others could exacerbate. A portion of the patients may recover from depression, and their symptom severity trends down. At any timepoint, a recovered patient is likely to relapse, and symptoms will subsequently recur and their severities may fluctuate. 
When applying existing prototype learning methods to our study, we are able to determine to what extent a single walking segment is ``normal.'' 
However, no depression symptom in a single walking segment does not imply that the patient has no depression because this segment might be during the ``normalcy phase,'' which could just be transitory and would quickly progress to severe depression. 
As a result, it is essential to detect and interpret the mental state of a patient based on a time series of walking segments that characterize the full course of symptom progression. 
Accordingly, we design trend prototypes that depict temporal progressions of symptom prototypes, which necessitate sophisticated function design, distributional inference, and generative process, as we articulate in Section \ref{sec:method:trend}.

Beyond the common limitation of static prototypes, each prototype learning for MTSC method has its own downsides.
When adapted to our study, \cite{ming_interpretable_2019}'s method only recognizes the entire walking segment as a prototype. This granularity is not desired, as this prototype cannot distinguish multiple symptoms from a single walking segment. 
\cite{zhang_2020_tapnet} use multiple data sources and include blackbox encoders in the feature extractor. Consequently, the prototype is only hidden embedding and cannot be traced back to a segment of the input. Because of this, \cite{zhang_2020_tapnet} only apply the prototype idea to prediction instead of interpretation. 
The prototypes learned in \cite{ghosal_2021_multi} are defined for each individual variable. It is more appropriate for MTSC problems where each dimension is a hand-crafted feature. In our context, hand-crafted features are not as effective as representation learning \citep{yu_wearable_2022}. 
The interpretation in \cite{trinh_2021_interpretable} is done in a post-hoc manner. However, many studies in interpretable ML have shown model-based interpretation, such as ours, is more faithful than post-hoc interpretation \citep{rudin_2019_stop,xie_unbox_2020}. Figure \ref{fig:our_vs_mtsc} visualizes the differences between our method and existing prototype learning for MTSC methods.

To solve the problem in the right part of Figure \ref{fig:our_vs_mtsc}, it is critical to learn two types of prototypes: symptom prototypes (e.g., short strides and slow gait velocity) and trend prototypes (e.g., symptom severity trending up, trending down, and fluctuating). To learn these prototypes, we need to tackle three methodological challenges. First, while rich sensor data can be collected during a walking segment, it is non-trivial to define prototypes representing depression symptoms, such that their existing strength (how strong a symptom presents in a walking segment) in a walking segment can be interpreted as the snapshot of symptom severity at the time of the walking segment. 
Second, the elapsed time between two consecutive walking segments is irregular. 
On the one hand, it is essential to explicitly consider these irregular inter-segment time intervals when inspecting the time series of symptom severities, because they reveal the progression of symptom severities and enable the definition of prototypes representing temporal depression symptom progression.
On the other hand, existing prototype learning for MTSC cannot deal with a time series of multivariate time series that are irregularly spaced in time, as indicated by the TP column in Table \ref{tb:prototypelearning} and the last column in Table \ref{tb:mtscpl}. This gap motivates us to develop a novel and effective method to analyze this new type of input for prototype learning.
Lastly, different patients often have varying numbers of walking segments within an observation time window. It is also unknown which phase of depression progression each walking segment is performed at. 
Consequently, given a prototypical temporal symptom progression, it is difficult to measure its existing strength in a walking segment sequence because the two objects are not aligned in time. That is, for each walking segment, we do not know against which part of the prototypical progression the walking segment should be compared. 

Another uniqueness of the wearable sensor data is that the walking segments are irregularly spaced in time. To design a novel approach to address this challenge, we review deep learning studies that incorporate the temporal dimension in prediction models. They have been applied to a variety of time-series predictions, such as Parkinson's prediction, cancer prediction, and temperature prediction \citep{li_predicting_2020,gao_distanced_2019,liu_td-lstm_2018}. Appendix \ref{apd:time_model_lit} summarizes these studies.
Although these ``time-aware'' models provide guidance to encode temporal information, their purposes are to address the data challenges to improve the \textit{prediction} performance of black-box models, such as LSTM. Distinct from these studies, our study aims to model a time series of irregularly spaced multivariate time series sensor data to improve the \textit{interpretation} and define a new dynamic prototype based on temporal progressions of symptoms, as an extension to the prototype learning models.

\section{The TempPNet Approach} \label{sec:method}

\subsection{Problem Setup} \label{sec:problem_setup}

For a patient $u$, let $y^{(u)}$ be the patient's depression status, where $y^{(u)}=1$ denotes depression, and $y^{(u)}=0$ represents non-depression. 
We also observe this patient's $N_u$ walking segments denoted by $X^{(u)}=\langle X^{(u)}_1,X^{(u)}_2,\dots,X^{(u)}_{N_u} \rangle$ that are recorded at timepoints $\langle t^{(u)}_1, t^{(u)}_2, \dots, t^{(u)}_{N_u} \rangle$. These walking segments can be measured by wearable sensors, such as the accelerometers in smartphones and smartwatches.
The $i$th walking segment $X_i^{(u)}$ observed at timepoint $t^{(u)}_i$ is comprised of a sequence of regularly sampled sensor data: $X^{(u)}_i= \langle a_1, a_2,\dots,a_L \rangle $, where vector $a_l$ is the sensor feature sampled at timepoint $l$.
Each sensor feature $a_l$ is derived from the accelerometer readings recorded in the walking segments. Please see Appendix \ref{apd:sen_features} for the details of $a_l$.

Let $\mathcal{U}$ denote the set of patients. We observe the dataset $\mathcal{D}=\{(X^{(u)},y^{(u)} )|u=1,2,\dots,|\mathcal{U}| \}$, where $X^{(u)}$ and $y^{(u)}$ represent the sequence of walking segments and the depression status of patient $u$, respectively, and $|\mathcal{U}|$ denotes the number of patients. Our objective is to learn a model from $\mathcal{D}$ that can detect the depression status of a new patient, based on sensor data collected from the patient's walking activities, and interpret the decision. Table \ref{tb:notation} summarizes the important notations.

\begin{table}[h]
\caption{Notation}
\label{tb:notation}
\footnotesize
\centering
\begin{threeparttable}
\begin{tabular}{L{0.07\textwidth} L{0.88\textwidth}}
\toprule
Notation &  \multicolumn{1}{c}{Description} \\ \midrule
$X_i$ & The sensor signal sequence of the $i$th walking segment of a focal patient. \\
$t_i$ & The timepoint when the walking segment $X_i$ is performed. \\
$H_i^{\mathbb{S}}$ & The feature matrix extracted from walking segment $X_i$. \\
$p_m^{\mathbb{S}}$ & The embedding vector for symptom prototype $m$. \\
$s_{m,i}^{\mathbb{S}}$ & The existence strength of symptom prototype $m$ in walking segment $X_i$, defined by Equation \ref{eq:s_m_i}. \\
$S$ & The symptom progression matrix detected for the focal patient, defined by Equation \ref{eq:sym_prog_mat}. \\
$\mathcal{G}^{(k)}(t)$ & The definition of trend prototype $k$, defined by Equation \ref{eq:g_k_t}. \\
$s_k^{\mathbb{T}}$ & The existence strength of trend prototype $k$ in symptom progression matrix $S$, defined by Equation \ref{eq:s_k_trend_final}. \\
\bottomrule
\end{tabular} 
\end{threeparttable}
\end{table}

\begin{figure}[h]
    \centering
    \includegraphics[width=0.7\textwidth]{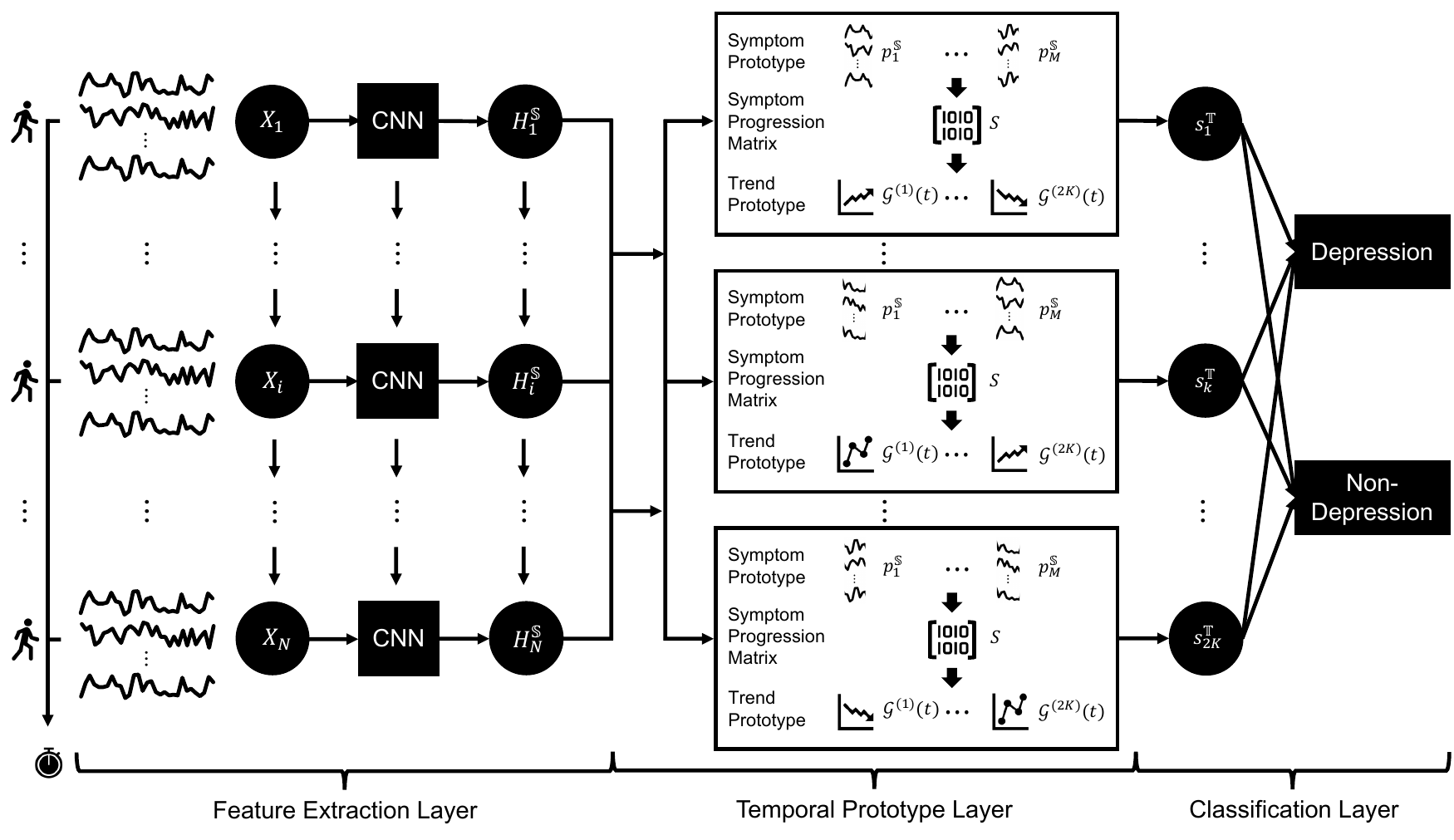}
    \caption{TempPNet Architecture}
    \label{fig:main_model}
\end{figure}

We present the \underline{Temp}oral \underline{P}rototype \underline{Net}work (TempPNet), an interpretable classifier equipped with a novel temporal prototype layer that detects depression status based on the prototypical temporal progression of walking symptoms. 
Figure \ref{fig:main_model} shows the architecture of our model, which features three building blocks: a feature extraction layer that represents a sequence of walking segments as a sequence of features, a temporal prototype layer that defines symptom prototypes and trend prototypes for depression detection and then computes the existing strength of these prototypes in a walking segment, and lastly a classification layer that classifies a patient into depression or non-depression based on the existing strength of trend prototypes.

\subsection{The Symptom Prototype and Symptom Progression Matrix } \label{sec:method:mat}

We consider a focal patient $u$, and thus drop the superscripts and subscripts related to the patient to simplify the notation. Recall that we can observe a sequence of $N$ walking segments for this patient, whose $i$th walking segment is represented by the sensor signal sequence $X_i$. To learn an effective representation of $X_i$, we employ a deep Convolutional Neural Network (CNN) layer \citep{lecun_1998_gradient,zhang_deep_2020}. Let $H_i^{\mathbb{S}}\in R^{n_o \times n_e}$ denote the learned embedding matrix for the sensor signal sequence $X_i$, where $n_o$ is the number of patches generated by the deep CNN layer and $n_e$ is the embedding dimension of each patch \citep{chen_this_2019}. We define $H_i^{\mathbb{S}}$ as
\begin{equation}
\label{eq:h_i_sym}
H_{i}^{\mathbb{S}} = \text{CNN}(X_{i})
\end{equation}
where the specifications of the deep CNN layer are articulated in Table \ref{tb:cnn_layer}. Let $H_{o|i}^{\mathbb{S}}$ denote the $o$th column of $H_{i}^{\mathbb{S}}$. 
The key benefit of using CNN for learning the embedding of $X_i$ is that each patch $H_{o|i}^{\mathbb{S}}$ has its own receptive field,
which can be identified and visualized as a local segment in $X_i$. Following \cite{chen_this_2019}, we identify the symptoms at the receptive field level.
We articulate how to leverage this property to inspect the learned prototypes in Section \ref{sec:method:proto_viz}.

Similar to prototype learning in image recognition, we aim to detect what prototypical walking symptoms that $H_{i}^{\mathbb{S}}$ (the representation of $X_i$) resembles. For this purpose, we need to learn $M$ symptom prototypes that represent typical walking symptoms, such as short strides and slow gait velocity \citep{lemke_spatiotemporal_2000, michalak_embodiment_2009, czech_gaitpy_2019}. These symptom prototypes are defined as the latent representation of prototypical waking patterns in the input signal. We embed each symptom prototype $m$ as a vector $p_m^{\mathbb{S}} \in R^{n_e}$ for $m=1,2,\dots,M$, which is in the same latent space as $H_{i}^{\mathbb{S}}$. Following the common strategy of prototype learning \citep{chen_this_2019}, symptom prototype vector $p_m^{\mathbb{S}}$ will be learned as model parameters, and then identified as and visualized by the walking segment where it presents most strongly. The interpretation mechanism is to find the prototypical patterns within the input sensor signal that are indicative of depression walking symptoms and thereby informative for depression detection. The middle part of Figure \ref{fig:main_model} provides an exemplar visualization of the symptom prototypes.
Let $s_{m,i}^{\mathbb{S}}$ denote the existing strength of symptom prototype $m$ in walking segment $X_i$, which can also be understood as the severity of symptom $m$ detected in the walking segment. We define $s_{m,i}^{\mathbb{S}}$ as
\begin{subequations}
\label{eq:s_m_i}
\begin{align}
s_{o|m,i}^{\mathbb{S}} &= \exp ( \gamma - {|| H_{o|i}^{\mathbb{S}} - p_m^{\mathbb{S}} ||}_{2}^2 ) \\
s_{m,i}^{\mathbb{S}} &= \max_{o=1,2,\dots ,n_o} s_{o|m,i}^{\mathbb{S}}
\end{align}
\end{subequations}
where $\gamma < 0$ is an infinitesimal constant to ensure $0 < s_{o|m,i}^{\mathbb{S}} < 1$ for $o=1,2,\dots,n_o$.
A high value of $s_{o|m,i}^{\mathbb{S}}$ suggests that strong existence of symptom prototype $m$, or high severity of walking symptom $m$, is detected in the $o$th region in walking segment $X_i$. As a result, $s_{m,i}^{\mathbb{S}}$ measures the overall severity of symptom prototype $m$ in walking segment $X_i$. 
Unlike existing prototype learning studies that predict and interpret based on static prototypes solely \citep{chen_this_2019,ming_interpretable_2019}, the severity of symptom prototype $s_{m,i}^{\mathbb{S}}$ is dynamic and forms a temporal progression pattern, as we observe patients' sensor data over time.
By collecting the severity scores of all symptom prototypes across the sequence of walking segments, we can construct the symptom progression matrix $ S \in R^{M \times N}$ as 
\begin{equation}
\label{eq:sym_prog_mat}
S = \begin{blockarray}{ccccc}
 t_1 & t_2 & \dots & t_N & \\
\begin{block}{[cccc]c}
s_{1,1}^{\mathbb{S}} & s_{1,2}^{\mathbb{S}} &\cdots & s_{1,N}^{\mathbb{S}} & \text{sym}_1 \bigstrut[t] \\
s_{2,1}^{\mathbb{S}} & s_{2,2}^{\mathbb{S}} & \cdots & s_{2,N}^{\mathbb{S}} & \text{sym}_2  \\
  \vdots & \vdots & \ddots & \vdots & \\
 s_{M,1}^{\mathbb{S}} & s_{M,2}^{\mathbb{S}} &\cdots & s_{M,N}^{\mathbb{S}} & \text{sym}_M \bigstrut[b]\\
\end{block}
\end{blockarray}
\end{equation}
where we label each column by the timepoint when the corresponding walking segment is performed, and each row by the corresponding symptom prototype ($\text{sym}_m$ denotes symptom $m$).
In matrix $S$, the $m$th row corresponds to the temporal progression of the severities of symptom $m$, while the $i$th column corresponds to the contemporaneous distribution of symptom severities observed at timepoint $t_i$. In other words, if we regard each symptom severity as a random variable that randomly evolves over time, then the column vector $S_i$ is a realization of these $M$ random variables drawn from their joint distribution specific to timepoint $t_i$.
From this perspective, the symptom progression matrix describes how the severities of the $M$ symptoms co-evolve over time. 
To make an interpretable detection, we need to detect whether such a matrix resembles typical depression or non-depression symptom progression trends such as in Figure \ref{fig:depression_trend}, which we define as trend prototypes.

\subsection{Detecting Prototypical Depression Symptom Progression Trends} \label{sec:method:trend}

A trend prototype is a prototypical depression or non-depression trend.
We aim to learn $2K$ trend prototypes: $K$ for the depression category and the remaining $K$ for the non-depression category. 
We view each trend prototype as a prototypical continuous co-evolution trajectory of the $M$ symptom severities over time, such as trending up, trending down, and fluctuating. Therefore, a trend prototype can be defined as a continuous vector-valued function of time.
Let $\mathcal{G}^{(k)}(t)$ denote the $k$th trend prototype, which maps a time scalar to a vector of symptom severities of length $M$. 
The mathematical details of $\mathcal{G}^{(k)}(t)$ will be given in Section \ref{sec:method:proto_layer:g_k_trend}.
We offer a particular interpretation of trend prototypes: if the focal patient's symptom progression matrix $S$ strongly resembles trend prototype $k$, then the symptom severities $S_i$ detected at timepoint $t_i$ should be close to $\mathcal{G}^{(k)}(t_i)$.
Based on this interpretation, the existing strength of trend prototype $k$ in symptom progression matrix $S$ can be measured as the overall closeness between $S_i$ and $\mathcal{G}^{(k)}(t_i)$ over observation timepoints $\langle t_1, t_2, \dots, t_{N} \rangle$. The implementation of this is articulated in Section \ref{sec:method:proto_layer:s_k_trend}.

\subsubsection{The Definition of Trend Prototypes:} \label{sec:method:proto_layer:g_k_trend}

A trend prototype is a continuous function of time that characterizes the evolution of symptom severities. This goal necessitates a continuous function that can trace any freely-valued trajectory over time. Studies have shown that any trajectory over time can be decomposed into a summation of sine and cosine curves in the frequency domain \citep{xu_inductive_2020, wang_word2fun_2021}. Inspired by those works, we design the trend prototype as follows. Let $\mathcal{G}_{m}^{(k)}(t)$ denote the severity of symptom $m$ at timepoint $t$ given by trend prototype $k$. Leveraging the time encoding method in \cite{xu_inductive_2020}, we compute $\mathcal{G}_{m}^{(k)}(t)$ as 
\begin{equation}
\label{eq:g_m_k}
\begin{aligned}
\mathcal{G}_{m}^{(k)}(t) &= \sigma \Big( \sqrt{\frac{1}{n_d}} \sum_{j=1}^{n_d} p_{k,m,j}^{\mathbb{T}} \cos(\omega_j t + \theta_j) + \sqrt{\frac{1}{n_d}} \sum_{j=1}^{n_d} p_{k,m,n_d+j}^{\mathbb{T}} \sin(\omega_j t + \theta_j) \Big) \\
  &= \sigma \big( \Phi ( t )^T p_{k,m}^{\mathbb{T}} \big)
\end{aligned}    
\end{equation}
where $p_{k,m}^{\mathbb{T}} = [p_{k,m,1}^{\mathbb{T}}, p_{k,m,2}^{\mathbb{T}}, \dots, p_{k,m,2 n_d}^{\mathbb{T}} ]^T \in R^{2 n_d}$ is the coefficient vector specific to trend prototype $k$ and symptom prototype $m$, $\Phi(t)$ is the time encoding function \citep{xu_inductive_2020} that transforms timepoint $t$ from a scalar to a numeric vector, defined as
\begin{equation}
\label{eq:time_encoding}
\begin{aligned}
\Phi(t) = \sqrt{\frac{1}{n_d}} \big[ &\cos(\omega_1 t+\theta_1), \cos(\omega_2 t+\theta_2), \dots, \cos(\omega_{n_d} t+\theta_{n_d}), \\
& \sin(\omega_1 t+\theta_1), \sin(\omega_2 t+\theta_2), \dots, \sin(\omega_{n_d} t+\theta_{n_d}) \big]^T
\end{aligned}
\end{equation}  
which is parameterized by frequency vector $\omega=[\omega_1,\omega_2,\dots,\omega_{n_d}]$ as well as phase vector $\theta=[\theta_1,\theta_2,\dots,\theta_{n_d}]$, and lastly $\sigma(.)$ is the sigmoid function defined as $\sigma(x)=1/\big(1+\exp(-x)\big)$ to ensure $0 < \mathcal{G}_{m}^{(k)}(t) < 1$ such that $\mathcal{G}_{m}^{(k)}(t)$ and $s_{m,i}^{\mathbb{S}}$ are in the same range and therefore comparable to each other.
Let $\sigma^{-1}(.)$ denote the inverse of the sigmoid function. Namely, $\sigma^{-1}(x)=\log \frac{x}{1-x}$. 
Our design of $\Phi(t)$ ensures that $\sigma^{-1}\big( \mathcal{G}_{m}^{(k)}(t) \big)$ is able to trace a freely-valued trajectory over time so that the temporal progressions of symptom severities can be seamlessly modeled by a continuous function. 
By learning the three vectors $p_{k,m}^{\mathbb{T}}$, $\omega$ and $\theta $ as model parameters, while treating $n_d$ as a hyperparameter, Equation \ref{eq:g_m_k} amounts to learning a prototypical temporal progression of symptom $m$ in the frequency domain. In a compact form, trend prototype $k$ can be written as Equation \ref{eq:g_k_t},
where $p_k^{\mathbb{T}} \in R^{M \times 2 n_d}$ is the coefficient matrix that specifies the function representation of trend prototype $k$, and is obtained by row stacking the transpose of vectors $p_{k,m}^{\mathbb{T}}$ for $m=1,2,\dots,M$. We can view $p_k^{\mathbb{T}}$ as the embedding form of trend prototype $k$ that is analogous to $p_m^{\mathbb{S}}$, the embedding form of symptom prototype $m$. 
The depression trend prototypes could be characterized by a rising symptom severity. Such a severity may have deviations from time to time while keeping the upward trend. These trend prototypes correspond to the onset and acute phases in Figure \ref{fig:depression_trend}. The non-depression trend prototype may be characterized by a downward severity trend, suggesting a previously depressed patient is recovering. The non-depression trend prototype could also be a curve fluctuating around a fixed level, indicating the patient never had depression symptoms or has fully recovered from prior depression. 
When a non-depressed patient relapses, depression trend prototypes could reappear in his or her walking sensor data.
In the subsequent discussion, we will frequently use $\sigma^{-1}\big(\mathcal{G}^{(k)}(t)\big)$, which means applying the inverse of the sigmoid function element-wisely on $\mathcal{G}^{(k)}(t)$. Therefore, we formally define Equation \ref{eq:g_k_t_tilde} to simplify the notation.
\begin{equation}
\label{eq:g_k_t}
\mathcal{G}^{(k)}(t) = \sigma \big( p_k^{\mathbb{T}} \Phi(t) \big)
\end{equation}
\begin{equation}
\label{eq:g_k_t_tilde}
{\tilde{\mathcal{G}}}^{(k)}(t) = \sigma^{-1}\big(\mathcal{G}^{(k)}(t)\big) = p_k^{\mathbb{T}} \Phi(t)
\end{equation}

\subsubsection{Detecting the Existing Strength of a Trend Prototype: }\label{sec:method:proto_layer:s_k_trend} 

Let $s_k^{\mathbb{T}}$ denote the existing strength of trend prototype $k$ in symptom progression matrix $S$.
Intuitively, $s_k^{\mathbb{T}}$ should be derived by comparing $S_i$ and $\mathcal{G}^{(k)}(t_i)$ for each timepoint $t_i$, where the former is the ``observed'' symptom severities detected at $t_i$, the latter is its corresponding ``ideal'' values in the prototypical progression patterns captured by trend prototype $k$. The larger the former deviates from the latter across time, the less likely trend prototype $k$ exists in the symptom progression matrix.
However, we argue that for interpretability concerns, $S_i$ should be compared to $\mathcal{G}^{(k)}(t_i-t_0^{(k)})$ rather than $\mathcal{G}^{(k)}(t_i)$, where $t_0^{(k)}$ is a latent variable indicating when the trend starts in the patient's timeline. The computation and rationale of $t_0^{(k)}$ are articulated below.

\begin{figure}[h]
    \centering
    \includegraphics[width=0.55\textwidth]{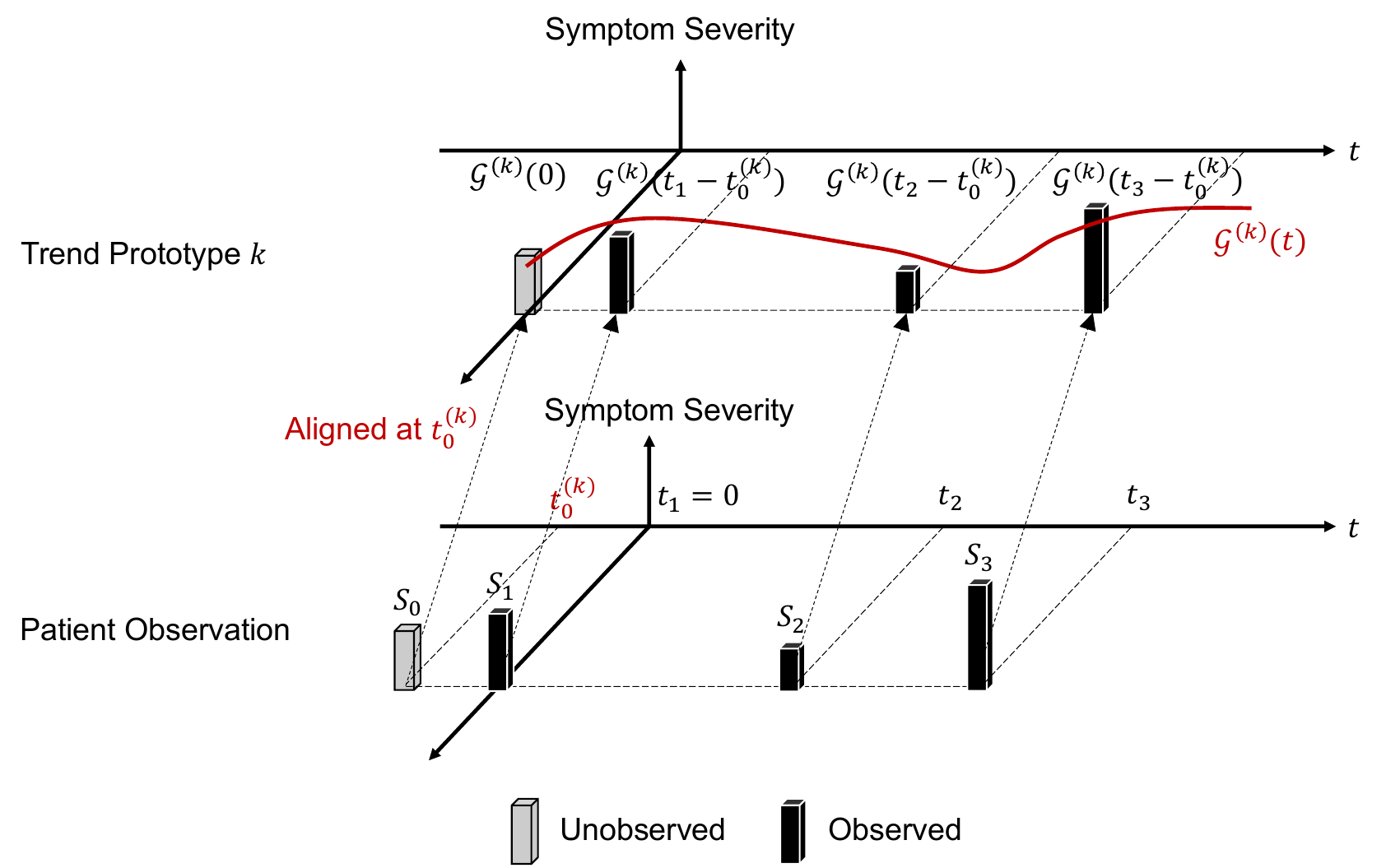}
    \caption{Latent Trend Starting Time}
    \label{fig:trend_patient}
\end{figure}

Recall that Equation \ref{eq:g_m_k} defines $\mathcal{G}^{(k)}(t)$ as a function on the domain $t \in (-\infty, +\infty)$. However, to understand what temporal progression that trend prototype $k$ captures, it is necessary to ensure that only a finite segment of $\mathcal{G}^{(k)}(t)$ carries useful information, so that only this segment needs to be visualized to inspect trend prototype $k$. For this purpose, we need to ``bound'' the trend by a starting time and an ending time.
We treat $t=0$ as the starting time of trend prototype $k$, and only work with the right part of $\mathcal{G}^{(k)}(t)$ defined on the non-negative domain $t \in [0, +\infty)$, as shown in Figure \ref{fig:trend_patient}. This design has two implications. Continuing with the example in Figure \ref{fig:trend_patient}, on one hand, $\mathcal{G}^{(k)}(0)$ should be treated as the initial state of trend prototype $k$, and it is the relative timepoint measuring how much time has elapsed since $t=0$ that truly matters. On the other hand, we can only observe the absolute timepoints when the focal patient performs walking segments in reality (e.g., $t_1$, $t_2$, and $t_3$). In general, we cannot assume that $t_1$, the absolute timepoint of the first walking segment performed by the patient, is the starting time of trend prototype $k$ in the patient's timeline, because $t_1$ is determined by when the first walking segment is performed as well as the range of the observation window. However, the depression progression trend of a patient $\mathcal{G}^{(k)}(t)$ would start regardless of whether a walking segment is performed or not. That unobserved trend starting time is noted as $t_0^{(k)}$.
If we were able to observe the zero-th walking segment $X_0$ at absolute timepoint $t_0^{(k)}$, and detect symptom severities $S_0$ from it by computing $s_{m,0}^{\mathbb{S}}$ in accordance with Equation \ref{eq:s_m_i} for $m=1,2,\dots,M$, then $S_0$ should be compared to $\mathcal{G}^{(k)}(0)$. 
In this sense, $S_i$, the symptom severity vector detected at $t_i$ in the patient's timeline, should be compared to $\mathcal{G}^{(k)}(t_i-t_0^{(k)})$, the corresponding prototypical values in the trend's timeline.
In Figure \ref{fig:trend_patient}, the detected symptom severity is compared with trend prototype $k$ by aligning the two timelines at a latent personalized trend starting time $t_0^{(k)}$. 

With $t_0^{(k)}$ at hand, only relative timepoints given by $t_i-t_0^{(k)}$ for $i=1,2,\dots,N$ carry information. As a result, we set $t_1=0$ and measure other timepoints including $t_0^{(k)}$ relative to $t_1$ without loss of generality.
To facilitate the learning of interpretable trend prototypes, we additionally impose the constraint $t_0^{(k)} < t_1=0$. If $t_0^{(k)} > 0$, the symptom severities detected before $t_0^{(k)}$ need to be compared to the left part of $\mathcal{G}^{(k)}(t)$ defined on domain $t \in (-\infty, 0)$, which we intend to avoid for interpretability concerns. 
For example, it is undesired to compare $S_1$ with $\mathcal{G}^{(k)}(t_1-t_0^{(k)})=\mathcal{G}^{(k)}(-t_0^{(k)})$.
Allowing such a comparison is in conflict with our interpretation of $t=0$ as the starting time of trend prototype $k$. Moreover, it also renders the learned function $\mathcal{G}^{(k)}(t)$ hard to inspect, because the progression patterns within the interval $t\in(-t_0^{(k)}, 0)$ also encode some prototypical patterns summarized from data. Given that $t_0^{(k)}$ varies across patients, the meaningful part of $\mathcal{G}^{(k)}(t)$ is different for different patients. 
In contrast, if we enforce $t_0^{(k)}<0$, all detected symptom severities will be compared to the right part of $\mathcal{G}^{(k)}(t)$ defined on the domain $t \in (0, +\infty)$, making $t=0$ a natural patient-independent starting point to inspect $\mathcal{G}^{(k)}(t)$.
Armed with the notion of $t_0^{(k)}$, we proceed to the definition of $s_k^{\mathbb{T}}$ by assuming a pre-computed $t_0^{(k)}$, and then discuss how to infer $t_0^{(k)}$ by the end of this section, where we introduce another design to impose an effective trend ending time. 

As pointed out by \cite{chen_this_2019}, a prototype can be viewed as a cluster, and its existing strength in an instance is essentially a closeness measurement between the instance and the cluster. 
To develop a principled definition of $s_k^{\mathbb{T}}$, we draw inspiration from the Gaussian Mixture Model \citep{murphy_probabilistic_2022}, a classic clustering algorithm, which measures the closeness between an instance and a cluster in terms of how likely the instance can be generated by the Gaussian component characterized by the cluster. By making an analogy in our context, we view the columns of $S$ as being generated from a time-varying distribution characterized by trend prototype $k$ at aligned timepoints $t_i-t_0^{(k)}$ for $i=1,2,\dots,N$. Different from the Gaussian Mixture Model which deals with instances described by freely-valued vectors, the entries of $S$ are bounded in the interval $(0,1)$. To properly specify the generation of the columns of $S$, we leverage the logistic-normal distribution, whose samples fall between $(0,1)$. We introduce this distribution first.

Let $x \sim \mathcal{N}(\mu, \Sigma)$ denote a random vector drawn from the $M$-dimensional normal distribution of mean $\mu \in R^{M}$ and covariance $\Sigma \in R^{M \times M}$.
Let $z=\sigma(x)$, which is obtained by transforming $x$ element-wisely with the sigmoid function such that $0 < z_m=\sigma(x_m) < 1$ for $m=1,2,\dots,M$. Then, $z$ follows $\mathcal{L}\mathcal{N}(\mu, \Sigma)$, the $M$-dimensional logistic-normal distribution with mean $\mu$ and covariance $\Sigma$, and its density can be computed as Equation \ref{eq:sigmoid_gauss_density}, where $\sigma^{-1}(z)$ means applying the inverse of the sigmoid function element-wisely on $z$, and $\det[\Sigma]$ is the determinant of matrix $\Sigma$. The derivation of Equation \ref{eq:sigmoid_gauss_density} can be found in Appendix \ref{apd:der_logistic_normal}.
\begin{equation}
\label{eq:sigmoid_gauss_density}
\mathcal{L}\mathcal{N}(z|\mu, \Sigma) =
\frac{1}{\prod_{m=1}^{M} z_{m}(1-z_{m})}
\frac{\exp \big( -\frac{1}{2} ( \sigma^{-1}(z) - \mu )^{T}{\Sigma }^{-1} (\sigma^{-1}(z) - \mu) \big) }{\sqrt{(2\pi)^{M} \det [ \Sigma] }}
\end{equation}

Now, consider a focal patient whose symptom progression matrix exhibits the presence of trend prototype $k$ to some extent. 
In this case, we regard $S_i$ as a realization drawn from the contemporaneous distribution of symptom severities given by $\mathcal{L}\mathcal{N}({\tilde{\mathcal{G}}}^{(k)}(t_i-t_0^{(k)}), I)$, the $M$-dimensional logistic-normal distribution characterized by mean ${\tilde{\mathcal{G}}}^{(k)}(t_i-t_0^{(k)})$ and covariance $I$ (the identity matrix), where ${\tilde{\mathcal{G}}}^{(k)}(t)$ is defined by Equation \ref{eq:g_k_t_tilde}, and $t_0^{(k)}$ is the personalized trend starting time explained previously. 
In our setting, $S_i$ is compared to $\mathcal{G}^{(k)}(t_i-t_0^{(k)})$ at the aligned timepoint $t_i-t_0^{(k)}$, while ${\tilde{\mathcal{G}}}^{(k)}(t)$ depicts the mean series of a time-varying logistic-normal distribution which characterizes some prototypical symptom progression patterns subject to a sigmoid transformation. 
We assume that the covariance of this time-varying distribution is the constant identity matrix for parsimonious concerns. 

The generative process of the symptom progression matrix is summarized in Appendix \ref{apd:algorithm}.
Then, we define $s_k^{\mathbb{T}}$ (Equation \ref{eq:s_k_trend_final}) as a scalar proportional to the log-likelihood of generating the column vectors of $S$ in accordance with Appendix \ref{apd:algorithm}. $S_{m,i}$ is the entry at row $m$ and column $i$ in matrix $S$. The second step of Equation \ref{eq:s_k_trend_final} is derived by expanding $\mathcal{L}\mathcal{N} \big( S_{i}| \tilde{\mathcal{G}}^{(k)}(t_i^{(k)}), I \big)$ using Equation \ref{eq:sigmoid_gauss_density} and then simplifying the resulting terms. The intuition is that the more likely the columns of $S$ can be observed from the generative process characterized by $\mathcal{G}^{(k)}(t)$, the stronger the evidence is that trend prototype $k$ exists in the symptom progression matrix of the focal patient. 
Equation \ref{eq:s_k_trend_final} ensures $0< s_k^{\mathbb{T}} < 1$, which means that the existence strength of trend prototypes has the same value range with symptom prototypes as defined by Equation \ref{eq:s_m_i}. Indeed, by comparing Equations \ref{eq:s_k_trend_final} and \ref{eq:s_m_i}, an obvious analogy can be established between the definition of $s_k^{\mathbb{T}}$ and $s_{o|m,i}^{\mathbb{S}}$.
\begin{equation}
\label{eq:s_k_trend_final}
\begin{aligned}
s_k^{\mathbb{T}} &= \sigma \Big( \log \prod_{i=1}^{N} \mathcal{L}\mathcal{N} \big( S_{i}| \tilde{\mathcal{G}}^{(k)}(t_i^{(k)}), I \big) \Big) \\
&= \sigma \Big( \sum_{i=1}^{N} \sum_{m=1}^{M} \log \frac{1}{S_{m,i}(1-S_{m,i})} - \frac{NM}{2} \log 2\pi  - \frac{1}{2} \sum_{i=1}^{N} || \sigma^{-1}(S_{i})-\tilde{\mathcal{G}}^{(k)}(t_i^{(k)}) ||_2^2   \Big)
\end{aligned}
\end{equation}

Lastly, we complete the definition of $s_k^{\mathbb{T}}$ by specifying the inference procedure of $t_0^{(k)}$. Because $t_0^{(k)}$ is used to generate the symptom progression matrix $S$, given $S$, we should be able to inversely infer $t_0^{(k)}$. 
Based on this intuition, we introduce an inference network, that takes $S$ and the observation timepoints $\langle t_1, t_2, \dots, t_{N} \rangle$ as input, and outputs $t_0^{(k)}$ as a negative scalar. Specifically, the inference network is defined in Equation \ref{eq:infer_t0}. $\oplus$ is the concatenation operator, $\Phi(.)$ is the time encoding function defined by Equation \ref{eq:time_encoding} but parameterized separately, GRU is a Gated Recurrent Unit layer \citep{cho_properties_2014} used to capture the temporal information in the symptom progression matrix augmented by time features, and lastly $h_i^{\mathbb{T}} \in R^{n_e}$ is the hidden state of the GRU layer at step $i$. In Equation \ref{eq:infer_t0_b}, $h_N^{\mathbb{T}}$ is the last hidden state of the GRU layer, which serves as a feature vector summarizing the information contained in $S$ and $\langle t_1, t_2, \dots, t_{N} \rangle$, $w_{k} \in R^{n_e}$ is a learnable parameter specific to trend prototype $k$, and $n_w>0$ is a hyperparameter. 
Equation \ref{eq:infer_t0} enforces that $-n_w < t_0^{(k)} < 0$. We impose this constraint to make the learned trend prototype easier to interpret. Consider the case where the observation window is 1 week, which means that the time difference between the first and the last walking segments is at most 1 week. If we set $n_w=1$ (time is measured in weeks), then $t_0^{(k)}$ is at most 1 week before the first walking segment, which implies that the symptom severities detected from the last walking segment will be compared to $\mathcal{G}^{(k)}(2)$ in the most extreme scenario. Consequently, we only need to inspect the segment of $\mathcal{G}^{k}(t)$ defined on the interval $t \in [0, 2]$, since only this segment has been compared to some real data. In this sense, $n_w$ reflects our belief on which segment of $\mathcal{G}^{k}(t)$ can be learned from data with reasonable qualities. 
\begin{subequations}
\label{eq:infer_t0}
\begin{align}
h_{i}^{\mathbb{T}} &= \text{GRU}\big(h_{i-1}^{\mathbb{T}}, S_{i} \oplus \Phi(t_i) \big) \label{eq:infer_t0_a} \\
t_0^{(k)} &= - n_w \sigma( w_{k}^T h_N^{\mathbb{T}} ) \label{eq:infer_t0_b}
\end{align}
\end{subequations}

\subsubsection{Superior Properties of the Trend Prototype:}\label{sec:method:proto_layer:summary} 
First, no matter how irregularly the focal patient has performed walking segments over time, the symptom progression matrix detected from these walking segments can always be compared to each trend prototype in a consistent manner. The irregularity of inter-segment time intervals is explicitly considered, because each trend prototype is a continuous function of time, and therefore the differences in inter-segment time intervals are reflected in the evaluation differences of function values.
Second, while trend prototype $k$ is learned from irregularly spaced point observations of symptom severities, it can be inspected as a complete temporal symptom progression by evaluating $\mathcal{G}^{(k)}(t)$ at regularly and densely spaced timepoints. Moreover, a trend prototype does not need to be attached to a single patient. Instead, walking segments gathered for different patients could be used to learn different segments of a trend prototype, which together pinpoint a complete temporal symptom progression spanning the observation window.

\subsection{Learning Objective} \label{sec:method:lr_obj}

The classification layer of TempPNet computes the probability of depression given the input sensor signal $X$ of a focal patient, which is defined as
\begin{equation}
\label{eq:class_layer}
    P(y=1|X) = \sigma \Big( \sum_{k \in \mathcal{K}^{+} } s_k^{\mathbb{T}} - \sum_{k \in \mathcal{K}^{-} } s_k^{\mathbb{T}} \Big)
\end{equation}
where $\mathcal{K}^{+}=\{1,2,\dots,K\}$, $\mathcal{K}^{-}=\{K+1,K+2,\dots,2K\}$, and $s_k^{\mathbb{T}}$ is defined by Equation \ref{eq:s_k_trend_final}. The above definition imposes the relationship that a high probability of depression is due to the detection of strong overall existing strength of depression trend prototypes, while weak overall existing strength of non-depression trend prototypes. 
The objective function of TempPNet to be minimized is defined based on the binary cross-entropy loss with two additional regularization terms. Specifically, 
\begin{equation}
  \label{eq:objective}
   \mathcal{L}_{\text{TempPNet}} = - \frac{1}{|\mathcal{U}|} \sum_{u \in \mathcal{U} } \log P(y^{(u)}|X^{(u)}) + \lambda_{\mathbb{S}} R_{\mathbb{S}} + \lambda_{\mathbb{T}} R_{\mathbb{T}}
\end{equation}
where $P(y^{(u)}|X^{(u)})$ is computed in accordance with Equation \ref{eq:class_layer} for patient $u$, $R_{\mathbb{S}}$ and $R_{\mathbb{T}}$ respectively denote the regularization term for the symptom prototypes and the trend prototypes, and lastly, $\lambda_{\mathbb{S}}$ and $\lambda_{\mathbb{T}}$ are the hyperparameters balancing the influence of the regularization terms. 
Let $\mathcal{U}^{+}$ denote the set of depressed patients, and $\mathcal{U}^{-}$ the set of non-depressed patients.
For symptom prototypes, we define $R_{\mathbb{S}}$ as
\begin{equation}
    \label{eq:R_S}
    R_{\mathbb{S}} = \frac{1}{M} \sum_{m=1}^{M} \Big( \frac{1}{|\mathcal{U}^{-}|} \sum_{u \in \mathcal{U}^{-}} 
       \frac{1}{N_u} \sum_{i=1}^{N_u} s_{m,i|u}^{\mathbb{S}}
       - 
       \frac{1}{|\mathcal{U}^{+}|} \sum_{u \in \mathcal{U}^{+}} 
       \frac{1}{N_u} \sum_{i=1}^{N_u} s_{m,i|u}^{\mathbb{S}} \Big)
\end{equation}
where $s_{m,i|u}^{\mathbb{S}}$ is the symptom severity computed in accordance with Equation \ref{eq:s_m_i} for patient $u$, and $\frac{1}{N_u}\sum_{i=1}^{N_u} s_{m,i|u}^{\mathbb{S}}$ is the average severity level of symptom $m$ detected in the walking test sequence of patient $u$. Minimizing $R_{\mathbb{S}}$ enforces that the average symptom severity level should be on average higher in the depressed group than it is in the non-depressed group for all symptoms, and thereby imposes our expected interpretation of symptom prototypes.
For trend prototypes, we define $R_{\mathbb{T}}$ as 
\begin{equation}
\label{eq:R_T}
\begin{aligned}
R_{\mathbb{T}} =& \frac{1}{|\mathcal{U}^{+}|} \sum_{u \in \mathcal{U}^{+}} 
   \big( \max_{k \in \mathcal{K}^{-} } s_{k|u}^{\mathbb{T}}  - \min_{k \in \mathcal{K}^{+} } s_{k|u}^{\mathbb{T}} \big)
   + \\
   & \qquad \frac{1}{|\mathcal{U}^{-}|} \sum_{u \in \mathcal{U}^{-}} 
   \big( \max_{k \in \mathcal{K}^{+} } s_{k|u}^{\mathbb{T}} - \min_{k \in \mathcal{K}^{-} } s_{k|u}^{\mathbb{T}} \big)
\end{aligned}
\end{equation}
where $s_{k|u}^{\mathbb{T}}$ is the trend existing strength computed in accordance with Equation \ref{eq:s_k_trend_final} for patient $u$. Minimizing the first (second) summation term in Equation \ref{eq:R_T} encourages the model to detect strong existing evidence for at least one depression (non-depression) trend prototype from the walking segment sequence of each depressed (non-depressed) patient, while weak existing evidence for all non-depression (depression) trend prototypes from the same walking segment sequence.

\subsection{Prototype Visualization} \label{sec:method:proto_viz}

TempPNet interprets its decision according to the following mechanism. If the focal patient is classified into the depression class, 
we must have $P(y=1|X) > 0.5$ under the decision threshold $0.5$, which implies that $\sum_{k \in \mathcal{K}^{+}} s_k^{\mathbb{T}} > \sum_{k \in \mathcal{K}^{-}} s_k^{\mathbb{T}}$, i.e., 
the overall existing strength of depression trend prototypes must be stronger than non-depression trend prototypes. 
In this case, we can interpret the decision made by TempPNet by inspecting the depression trend prototypes that strongly present in the walking segment sequence of the focal patient, i.e., trend prototypes with values of $s_k^{\mathbb{T}}$.
Different from existing prototype learning methods, where prototypes are learned in a latent space that is not directly interpretable \citep{chen_this_2019}, the trend prototypes learned by TempPNet can be directly inspected by visualizing them as trajectories of symptom severities evolving over time. Moreover, as explained in Section \ref{sec:method:proto_layer:s_k_trend}, to inspect one trajectory of a trend prototype, it is enough to visualize its segment for the time interval $t\in [0, n_w+n_T]$, where $n_w$ is the hyperparameter defined in Equation \ref{eq:infer_t0_b}, and $n_T$ is the observation window defined as the maximum span from the timepoint of the first walking segment to the timepoint of the depression label.

However, trend prototype $p_{k,m}^{\mathbb{T}}$ does not stand alone. It depicts the typical temporal progression pattern of symptom $m$. Therefore, visualizing the underlying symptom prototype $m$ is critical for a thorough understanding of model decisions. Given that the learned embedding vector of symptom prototype $m$ is $p_{m}^{\mathbb{S}}$, 
we leverage the sensor data to offer an interpretation for $p_{m}^{\mathbb{S}}$ in the following approach. First, we search for the particular walking segment where symptom prototype $m$ presents most strongly across all patients. 
Mathematically, this means solving for the combination $(u^*,i^*) = \argmax_{u,i} s_{m,i|u}^{\mathbb{S}}$, where $u \in \mathcal{U}$ and $i\in \{1,2,\dots,N_u \}$. 
Second, based on the identified sensor signal sequence $X_{i^*}^{(u^*)}$, we use Equation \ref{eq:s_m_i} to further search for the particular patch $o^*$ where symptom prototype $m$ presents most strongly. Mathematically, $o^* = \argmax_{o} s_{o|m,i^{*}}^{\mathbb{S}}$, where $o \in \{1,2,\dots,n_o\}$ and the patient index $u^*$ is omitted for simplicity.
Lastly, we inspect symptom prototype $m$ by visualizing the receptive field of this patch, which corresponds to a local segment in $X_{i^*}^{(u^*)}$ that contributes mostly to the computation of $H_{o^*|i^*}^{\mathbb{S}}$, the embedding vector of this patch defined by Equation \ref{eq:h_i_sym}. 
To this end, we adopt the gradient-based approach \citep{luo_understanding_2016}. For each timepoint $l$ in $X_{i^*}^{(u^*)}$, we compute $\partial H_{o^*|i^*}^{\mathbb{S}} / \partial a_l$, which is the Jacobian matrix of the patch embedding vector w.r.t. the sensor feature recorded at timepoint $l$. 
By measuring the overall importance score of $a_l$ relative to $H_{o^*|i^*}^{\mathbb{S}}$ as $|| \partial H_{o^*|i^*}^{\mathbb{S}} / \partial a_l ||$, the Frobenius norm of the Jacobian matrix, we can pinpoint the receptive field of patch $o^*$ as a local segment in $X_{i^*}^{(u^*)}$, within which the importance scores of sensor signal inputs are above a pre-specified threshold such as zero.

\vspace{-5pt} 
\section{Empirical Analyses} \label{sec:em_ana}

\subsection{Research Context and Data Collection}  \label{sec:em_ana:nhanes}
We have obtained the National Health and Nutrition Examination Survey (NHANES) dataset. The NHANES dataset includes depression labels, measured by PHQ-9, for a variety of chronic disease patients. It also includes time series wearable sensor data collected by smartwatches. The background, data types, data collection process and preprocessing of the NHANES dataset is articulated in Appendix \ref{apd:nhanes_dataset}.
Our final dataset for analysis includes participants with at least one chronic disease and participated in the wearable sensor measurement, encompassing 1,069 patients (42\% are depressed). We split this dataset into 60\% for training, 20\% for validation, and 20\% for test.
To show our method's generalizability, we obtained a second dataset: mPower, a smartphone-based study that collects daily motion sensor signals for 916 chronic disease patients \citep{bot_mpower_2016}. We replicate all the empirical analyses for the mPower dataset (Appendix \ref{apd:mpower}). The results are consistent with the NHANES dataset.

\subsection{Depression Detection Evaluation} \label{sec:em_ana:eval}

The benchmark selection and hyperparameter settings are shown in Appendix \ref{apd:benchmark}. We adopt F1-score, precision, and recall as the evaluation metrics. The best model should have the highest F1-score. The detection performance, input, and interpretability of our model and the benchmarks are reported in Tables \ref{tb:pred_eval_ml_nhanes} and \ref{tb:pred_eval_mtsc_nhanes}. These performances are the mean of 10 experimental runs. We also report the standard deviations of the performances.
We first compare with the commonly used machine and deep learning models in sensing studies.
Compared to the best deep learning model (RNN), our model increases F1-score by 0.084. This increase is attributed to our model's capability of capturing temporal symptom progression and depression symptoms. Compared to the leading feature-based ML model (SVM), TempPNet boosts F1-score by 0.128. This performance enhancement is due to our model's ability to learn effective features from the raw sensor signal.

\begin{table}[h]
\centering
\caption{Detection Performance Comparison with Machine and Deep Learning Methods (NHANES)}
\label{tb:pred_eval_ml_nhanes}
\footnotesize
\begin{threeparttable}
\begin{tabular}{L{80pt}C{50pt}C{50pt}C{57pt}C{57pt}C{57pt}C{57pt}}
\toprule
 Model & Input & Interpretable & F1-score & Precision & Recall \\ \midrule
 TempPNet (Ours) & Raw sensor & Yes & $0.778 \pm 0.009$ & $0.764 \pm 0.011$ & $0.792 \pm 0.016$ \\
  CNN & Raw sensor & No & $0.686 \pm 0.014$ & $0.664 \pm 0.146$ & $0.710 \pm 0.163$ \\
 RNN & Raw sensor & No & $0.694 \pm 0.013$ & $0.684 \pm 0.174$ & $0.704 \pm 0.176$ \\
 KNN & Features & No & $0.492 \pm 0.000$ & $0.978 \pm 0.000$ & $0.328 \pm 0.000$ \\
 SVM & Features & No & $0.650 \pm 0.000$ & $0.985 \pm 0.000$ & $0.485 \pm 0.000$ \\
 Random forest & Features & No & $0.562 \pm 0.056$ & $0.611 \pm 0.228$ & $0.602 \pm 0.157$ \\
 AdaBoost & Features & No & $0.508 \pm 0.000$ & $0.979 \pm 0.000$ & $0.343 \pm 0.000$ \\
 XGBoost & Features & No & $0.629 \pm 0.000$ & $0.984 \pm 0.000$ & $0.463 \pm 0.000$ \\
 \bottomrule
\end{tabular}
\end{threeparttable}
\end{table}

As described in the literature review, our study is related to MTSC and prototype learning for MTSC. Although our model significantly differs from these models in data structure and model design, we still apply those studies to our problem in order to show the successful design choice of our model and validate our methodological novelty. Compared to regular MTSC models without temporal progressions of prototypes \citep{ismail_2020_inceptiontime}, our model increases F1-score by 0.102. This result proves that capturing the prototypes and their temporal progressions assists in detection performance. Compared to the best-performing prototype learning for MTSC model \citep{ming_interpretable_2019}, TempPNet improves F1-score by 0.073. Such a significant performance gain indicates that capturing the temporal progressions of prototypes greatly contributes to depression detection.

\begin{table}[h]
\centering
\caption{Detection Performance Comparison with MTSC and Prototype Learning for MTSC (NHANES)}
\label{tb:pred_eval_mtsc_nhanes}
\footnotesize
\begin{threeparttable}
\begin{tabular}{L{78pt}C{55pt}C{100pt}C{55pt}C{55pt}C{55pt}}
\toprule
 Model  & Interpretable & Progression of Prototype & F1-score & Precision & Recall \\ \midrule
 TempPNet (Ours) & Yes & Yes & $0.778 \pm 0.009$ & $0.764 \pm 0.011$ & $0.792 \pm 0.016$  \\
 \cite{chen_this_2019} & Yes & No & $0.698 \pm 0.013$ & $0.686 \pm 0.144$ & $0.710 \pm 0.165$ \\
 \cite{ming_interpretable_2019} & Yes & No & $0.705 \pm 0.012$ & $0.707 \pm 0.179$ & $0.704 \pm 0.176$ \\
 \cite{gee_2019_explaining} & Yes & No & $0.672 \pm 0.017$ & $0.646 \pm 0.135$ & $0.701 \pm 0.160$ \\
 \cite{ma_2020_interpretable} & Yes & No & $0.695 \pm 0.014$ & $0.681 \pm 0.177$ & $0.710 \pm 0.180$ \\
 \cite{ismail_2020_inceptiontime} & No & No & $0.676 \pm 0.016$ & $0.646 \pm 0.127$ & $0.755 \pm 0.142$ \\
 \bottomrule
\end{tabular}
\end{threeparttable}
\end{table}

Since our model consists of multiple critical design components, we further perform ablation studies to show their effectiveness, as reported in Table \ref{tb:ablation_nhanes}. We first remove the latent trend starting time design ($t_0^{(k)}$). We also remove the trend prototype design. After removing the trend prototype, the model loses the capability of detecting temporal symptom progression. Consequently, we test two options: using the last symptom severity to detect depression and using the average symptom severity over time to detect depression. Table \ref{tb:ablation_nhanes} suggests that removing any design component will significantly hamper the detection accuracy, proving that our design choice is optimal.

\begin{table}[h]
\centering
\caption{Ablation Studies (NHANES)}
\label{tb:ablation_nhanes}
\footnotesize
\begin{threeparttable}
\begin{tabular}{L{250pt}C{57pt}C{57pt}C{57pt}}
\toprule
 Model & F1-score & Precision & Recall \\ \midrule
 TempPNet (Ours) & $0.778 \pm 0.009$ & $0.764 \pm 0.011$ & $0.792 \pm 0.016$   \\
 TempPNet removing offset $t_0^{(k)}$ & $0.694 \pm 0.024$ & $0.731 \pm 0.020$ & $0.661 \pm 0.036$ \\
 Remove trend prototype using last symptom severity & $0.711 \pm 0.022$ & $0.719 \pm 0.021$  & $0.703 \pm 0.025$  \\
 Remove trend prototype using average symptom severity & $0.724 \pm 0.012$ & $0.708 \pm 0.013$ & $0.740 \pm 0.017$ \\
 \bottomrule
\end{tabular}
\end{threeparttable}
\end{table}

To reduce noise in the sensor data and avoid overfitting, sensor-based prediction studies usually downsample the sensor signals \citep{sigcha_deep_2020}. We test the effect of different sample rates in Table \ref{tb:analysis_frequency_nhanes}: 10 Hz, 20 Hz, and 30 Hz. The results suggest that 10 Hz signal frequency achieves the best performance. Therefore, we use the 10 Hz signal frequency for all the other analyses.

\begin{table}[h]
\centering
\caption{Analysis of Signal Frequency (NHANES)}
\label{tb:analysis_frequency_nhanes}
\footnotesize
\begin{threeparttable}
\begin{tabular}{L{100pt}C{57pt}C{57pt}C{57pt}C{57pt}}
\toprule
 Signal Frequency  & F1-score & Precision & Recall \\ \midrule
 10 Hz  & $0.778 \pm 0.009$ & $0.764 \pm 0.011$ & $0.792 \pm 0.016$   \\
 20 Hz  & $0.754 \pm 0.011$ & $0.742 \pm 0.015$ & $0.766 \pm 0.018$ \\
 30 Hz  & $0.747 \pm 0.013$ & $0.735 \pm 0.014$ & $0.761 \pm  0.017$ \\
 \bottomrule
\end{tabular}
\end{threeparttable}
\end{table}

Certain pre-existing chronic diseases and depression may have a correlation because they share similar walking symptoms. To make sure that our model is actually detecting depression instead of other pre-existing chronic diseases, we perform evaluations that are conditioned on whether a patient has a chronic disease or not. Since our dataset contains numerous chronic diseases, we report two conditions (have or not have) for each disease to be concise in the main manuscript. In Appendix \ref{apd:eval_disease_severity_nhanes}, we further select two diseases (diabetes and kidney disease) to showcase our model's performances when conditioned on more nuanced disease severity levels. If our model indeed detects depression, we expect that, conditioned on each disease, our model's performance on depression detection should remain consistently high. If our model only detects pre-existing chronic diseases, conditioned on having (or not having) the disease, the performance should be very low because, in this group, the model has not seen different values of the outcome, thus unable to update parameters. Table \ref{tb:eval_condition_disease_severity}'s results prove that given having (or not having) any chronic disease, our model is able to accurately detect depression consistently. The results of evaluations on more nuanced disease severity levels in Appendix \ref{apd:eval_disease_severity_nhanes} show the same conclusion, suggesting our depression detection is robust in any pre-existing disease severity levels. Therefore, our model indeed detects depression rather than pre-existing chronic diseases.

\begin{table}[h]
\centering
\caption{Evaluations Conditioned on Chronic Diseases (NHANES)}
\label{tb:eval_condition_disease_severity}
\scriptsize
\begin{threeparttable}
\begin{tabular}{C{30pt}C{15pt}C{48pt}C{48pt}C{48pt}C{40pt}C{15pt}C{48pt}C{48pt}C{48pt}}
\toprule
 Disease & Status  & F1-score & Precision & Recall & Disease & Status  & F1-score & Precision & Recall \\ \midrule
 HBP & Yes  & $0.806 \pm 0.021$ & $0.806 \pm 0.032$ & $0.807 \pm 0.021$ & Gout & Yes  &  $0.741 \pm 0.036$ & $0.716 \pm 0.059$ & $0.769 \pm 0.036$ \\
 HBP & No  &  $0.770 \pm 0.024$ & $0.752 \pm 0.039$ & $0.789 \pm 0.025$  & Gout & No  &  $0.779 \pm 0.021$ & $0.767 \pm 0.038$ & $0.793 \pm 0.022$  \\
 CVD & Yes  & $0.807 \pm 0.023$ & $0.767 \pm 0.038$ & $0.853 \pm 0.026$  & Heart & Yes  &  $0.759 \pm 0.045$ & $0.743 \pm 0.042$ & $0.778 \pm 0.065$  \\
 CVD & No  &  $0.769 \pm 0.016$ & $0.768 \pm 0.029$ & $0.771 \pm 0.013$  & Heart & No  &  $0.779 \pm 0.016$ & $0.763 \pm 0.030$ & $0.797 \pm 0.013$ \\
 Diabetes & Yes  & $0.777 \pm 0.028$ & $0.769 \pm 0.056$ & $0.787 \pm 0.014$  & Stroke & Yes  & $0.839 \pm 0.055$ & $0.810 \pm 0.080$ & $0.873 \pm 0.043$ \\
 Diabetes & No  &  $0.775 \pm 0.035$ & $0.758 \pm 0.064$ & $0.796 \pm 0.024$  & Stroke & No  & $0.772 \pm 0.022$ & $0.764 \pm 0.038$ & $0.782 \pm 0.022$  \\
 Kidney & Yes  & $0.809 \pm 0.037$ & $0.807 \pm 0.061$ & $0.813 \pm 0.018$  & Emphysema & Yes  & $0.825 \pm 0.053$ & $0.778 \pm 0.087$ & $0.882 \pm 0.027$ \\
 Kidney & No  & $0.763 \pm 0.040$ & $0.757 \pm 0.081$ & $0.773 \pm 0.023$  & Emphysema & No  & $0.769 \pm 0.020$ & $0.747 \pm 0.032$ & $0.794 \pm 0.019$  \\
 Asthma & Yes  & $0.795 \pm 0.015$ & $0.735 \pm 0.046$ & $0.869 \pm 0.039$  & Bronchitis & Yes  & $0.805 \pm 0.035$ & $0.742 \pm 0.055$ & $0.881 \pm 0.026$  \\
 Asthma & No  &  $0.767 \pm 0.017$ & $0.782 \pm 0.033$ & $0.754 \pm 0.016$  & Bronchitis & No  & $0.752 \pm 0.022$ & $0.746 \pm 0.038$ & $0.759 \pm 0.021$  \\
 Celiac & Yes  & $0.755 \pm 0.093$ & $0.892 \pm 0.102$ & $0.689 \pm 0.175$  & COPD & Yes  &  $0.811 \pm 0.025$ & $0.756 \pm 0.043$ & $0.876 \pm 0.018$  \\
 Celiac & No  &  $0.783 \pm 0.018$ & $0.772 \pm 0.033$ & $0.795 \pm 0.016$  & COPD & No  & $0.767 \pm 0.020$ & $0.762 \pm 0.036$ & $0.773 \pm 0.021$  \\
 Arthritis & Yes  & $0.808 \pm 0.022$ & $0.738 \pm 0.031$ & $0.894 \pm 0.022$  & Cancer & Yes  & $0.766 \pm 0.024$ & $0.777 \pm 0.059$ & $0.764 \pm 0.080$  \\
 Arthritis & No  & $0.758 \pm 0.031$ & $0.793 \pm 0.029$ & $0.728 \pm 0.049$  & Cancer & No  &  $0.782 \pm 0.023$ & $0.765 \pm 0.047$ & $0.801 \pm 0.017$  \\
 \bottomrule
\end{tabular}
\end{threeparttable}
\end{table}

\subsection{Interpretation of Depression Detection}

\begin{figure}[h]
    \centering
    \includegraphics[width=1\textwidth]{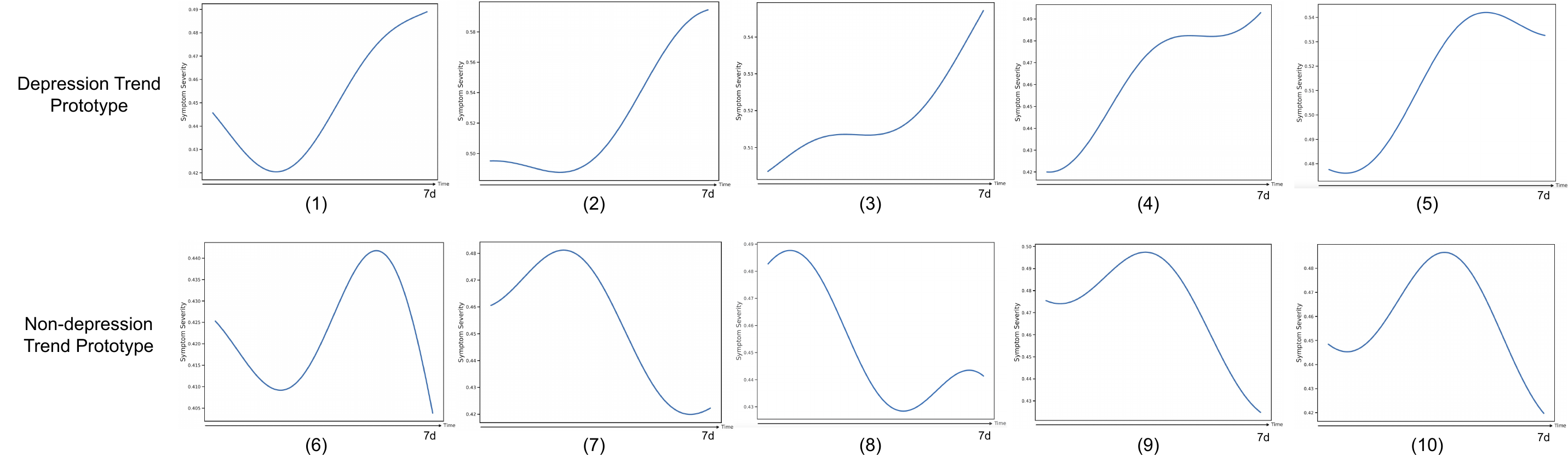}
    \caption{Trend Prototypes}
    \label{fig:trend-results-nhanes}
\end{figure}

Beyond depression detection, TempPNet is capable of interpreting why a patient is classified as depressed by presenting the contributing temporal symptom progression (trend prototype) and the corresponding walking symptom (symptom prototype). Figure \ref{fig:trend-results-nhanes} shows the trend prototypes that our model learned from the NHANES dataset. These trend prototypes are the prototypical depression or non-depression trends. They are learned in a data-driven manner and show representative trends. For each picture, the x-axis is time, and the y-axis is symptom severity.

\begin{figure}[h]
    \centering
    \includegraphics[width=0.9\textwidth]{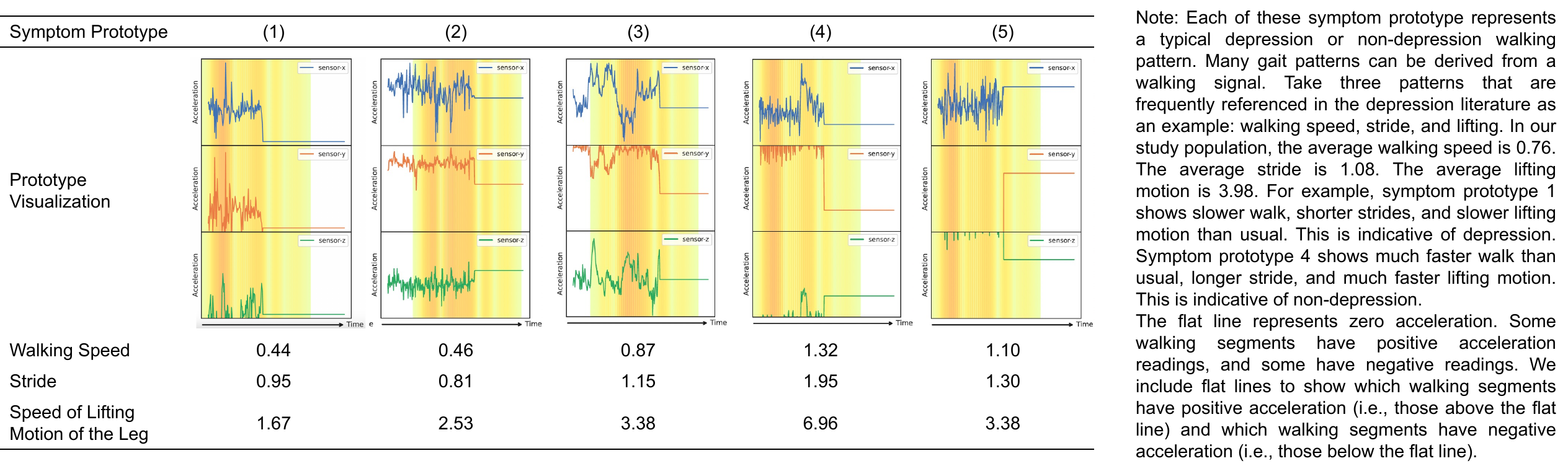}
    \caption{Symptom Prototypes}
    \label{fig:symptom-results-nhanes}
\end{figure}

Trend prototypes (1)-(5) are depression trend prototypes. They represent the severities of depression symptoms trending up. Some of them have deviations from time to time in the upward trend, representing temporary symptom relief and deterioration of depression. This conforms with the typical depression trend \citep{bockting_lifetime_2015}. Trend prototypes (6)-(10) are non-depression trend prototypes, where (7)-(10) represent trending down and (6) represents fluctuating at a certain level. These are not typical depression trends. The trend prototypes are learned using all the patients' data rather than relying on a single patient's data. Each patient's observed walking data could be at different stages of a trend — some at the rising stage, some at the stable stage, among others. Together they depict a complete trend. Multiple patients could also share the same trend prototype if their symptom severity levels are at the same stage (e.g., all on the rise). Each trend prototype is coupled with underlying symptoms. Figure \ref{fig:symptom-results-nhanes} shows the symptom prototypes that our model learned.

The prototype visualization in Figure \ref{fig:symptom-results-nhanes} shows the sensor signal of the symptoms. These symptom prototypes are learned by our method across all patients. Each patient may present none, one, or more of these symptoms in their walking patterns. Prior literature suggests that depression walking symptoms can be reflected in gait patterns, such as walking speed, stride, and speed of the lifting motion of the leg \citep{sloman_gait_1982,lemke_spatiotemporal_2000,nhs_symptoms_2022}. To accompany the interpretation of the detection of a patient, these gait patterns can be computed for the symptom prototypes.

\begin{figure}[h]
    \centering
    \includegraphics[width=\textwidth]{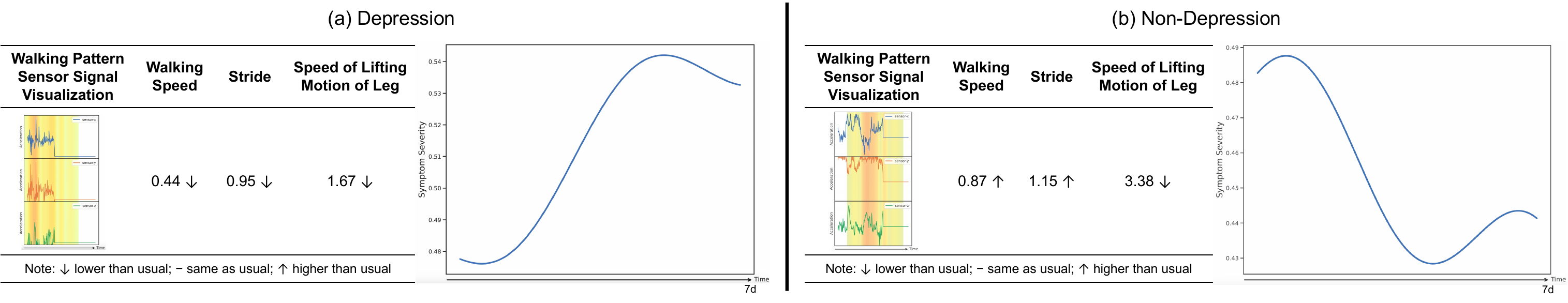}
    \caption{Interpretation of A Depressed and A Non-depressed Patient}
    \label{fig:interpret-nhanes}
\end{figure}

Leveraging the above-learned trend prototypes and symptom prototypes, our model can interpret the detection of depression for every patient. We randomly select two patients (one depressed and one non-depressed) and showcase TempPNet's interpretation for them. Figure \ref{fig:interpret-nhanes} shows the interpretations of the depressed and non-depressed patients. For simplicity, we only show the trend prototype with the highest existing strength and the corresponding symptom prototype with the highest existing strength in these examples. For the symptom prototype, we also compute a few exemplar gait patterns using GaitPy \citep{pypi_2024_Gaitpy} to explain the encoded information from the visualization. The arrows after the gait patterns denote whether a pattern is higher or lower than an average human. They do not imply any trend information (neither go up nor go down).

TempPNet detects the patient in Figure \ref{fig:interpret-nhanes}(a) as depressed for two reasons. First, this patient's walking patterns strongly present a walking symptom like in the left part of Figure \ref{fig:interpret-nhanes}(a). This walking symptom is manifested as slower-than-usual walking speed, shorter stride, and slower lifting motion of the leg. This symptom conforms with the depression physical symptoms in the literature \citep{sloman_gait_1982,lemke_spatiotemporal_2000,nhs_symptoms_2022}. Second, the severity of this symptom presents a temporal progression pattern like the right part of Figure \ref{fig:interpret-nhanes}(a). TempPNet believes this temporal symptom progression pattern resembles a typical depression progression pattern. According to the depression progression literature \citep{bockting_lifetime_2015,dattani_mental_2021}, this judgment makes sense — this patient's depression walking symptom first worsens rapidly and then peaks, similar to the onset and acute phases in Figure \ref{fig:depression_trend}.
TempPNet detects the patient in Figure \ref{fig:interpret-nhanes}(b) as non-depressed for two reasons. First, this patient's walking patterns strongly present a walking symptom like in the left part of Figure \ref{fig:interpret-nhanes}(b). This walking symptom is manifested as faster-than-usual walking speed, longer stride, and slower lifting motion of the leg. This symptom does not resemble the typical depression walking symptoms. Second, the severity of this symptom presents a temporal progression pattern like the right part of Figure \ref{fig:interpret-nhanes}(b). The symptom severity trends down and has fluctuations in the middle. This trend does not resemble a typical depression trend.

As medical experts are aware of what a typical depression walking symptoms and temporal progression look like, our interpretation helps doctors understand why a patient is detected as depressed. When our interpretation matches the medical understanding, as is the case in Figure \ref{fig:interpret-nhanes}, doctors would trust our model’s decision more. Based on our interpretation, doctors can also assess what symptoms the patient presents and in what phase of the depression progression the patient is at. This interpretation information coupled with our detection result could inform doctors to take corresponding interventions to treat depression with increased confidence and precision.

\vspace{-5pt} 
\subsection{Human Evaluations}
\vspace{-5pt} 
\subsubsection{Large Scale User Study:}

The major contribution of TempPNet in interpretation lies in its capability of interpreting temporal symptom progression. To validate that capturing temporal symptom progression improves interpretability of our model, we conduct a user study. We informed the participants that they would be assigned an interpretable ML model to detect depression using sensor data. After filtering out those who failed the attention check and manipulation check (articulated below), there are 66 participants in the user study. We randomly select two patient samples that the model detects as depressed and non-depressed and show the participants how the model interpreted those decisions. We design two randomized groups: one presenting TempPNet's interpretation, and the other presenting the baseline's interpretation (without temporal symptom progression).

The first part of the user study collects five control variables: age, education, gender, trust in AI, and health literacy. The health literacy instrument is adopted from \cite{osborne_2013_grounded}. This user study passed randomization checks. The summary statistics and randomization \textit{p}-values are reported in Appendix \ref{apd:user_study_designs}.
Since the interpretation is based on depression walking symptoms, we design a depression knowledge education session for all participants, showing the depression walking symptoms (Appendix \ref{apd:user_study_designs}). 
After reading such information, they are asked to answer four questions (Appendix \ref{apd:user_study_designs}) to test their understanding. If they answer these questions incorrectly, an error message will prompt and direct them to read the information and choose again. After they answer the test questions correctly, they have sufficient knowledge to understand the model interpretations.

Next, we inform the participants the context, input, and output. Then, we show the model interpretation to each group respectively, as shown in Appendix \ref{apd:user_study_designs}.
A manipulation check question is asked to verify whether the participants read the interpretation carefully (``How does the walking speed of the above walking symptom compare to the usual case?''), whose answer can be understood from the arrow sign after the value of the walking speed in the interpretation figure. 

Subsequently, we ask the participants to rate the interpretability of the given model.
The definition of interpretability varies across domains \citep{lee_2018_focused}. In business analytics and ML \citep{lee_2018_focused,miller_2019_explanation,molnar_interpretable_2020}, studies define interpretability as the degree to which a user can trust the cause of a decision \citep{miller_2019_explanation, molnar_interpretable_2020, xie_unbox_2020}. 
Trust in ML models also holds paramount importance in healthcare analytics for practitioners' and patients' adoption considerations. Following the literature, we use trust in automated systems as a interpretability metric. We also use other interpretability metrics in the expert evaluation in the next subsection. The measurement scale of trust is adopted from \cite{jian_2000_foundations}. The Chronbach's Alpha is 0.884, suggesting excellent reliability. An attention check question is added (``Please just select neither agree nor disagree.''). 
The participants' trust in our model (mean = 2.26) is significantly higher than the baseline model (mean = 1.71, \textit{p} < 0.001). This result indicates that interpreting the temporal symptom progression is a critical booster for users' trust in depression prediction models, endorsing our model design.

Afterwards, we show the participants the interpretation of the other model as a comparison. We ask them to choose a model that they would finally adopt. 58 participants (88\%) chose TempPNet over the baseline. When asked why they adopt TempPNet rather than the baseline, 43 participants mentioned, in similar wording, that ``because this model visualized the walking symptom progression.'' The above user study results prove that by interpreting the temporal symptom progression, TempPNet improves interpretability, which offers empirical evidence for the contribution of our model innovation.

\subsubsection{Medical Expert Panel:} \label{sec:expert_panel}

The user study employs non-experts who may be unable to judge the clinical meaningfulness of the interpretations. Interpretability also has other objectives. To address these limitations, we employ a medical expert panel. 
One representative study that employs a medical expert panel is a recent MISQ article by \cite{kim_2023_rolex}, which also develops an interpretable ML method for healthcare applications. 
\cite{kim_2023_rolex}'s method is a post-hoc method and provides feature-based interpretations, which are undesired in our context, as articulated in Section \ref{sec:lr:iml}.
Since both our study and \cite{kim_2023_rolex} are IS studies that develop interpretable ML methods for healthcare applications, we follow the evaluation precedence used by \cite{kim_2023_rolex} to conduct our expert panel evaluation.
Specifically, we employed an expert panel of 8 medical professionals, including 2 clinical psychologists (both having PhDs) and 6 psychiatrists (all having MDs), which is larger than the expert panel utilized by \cite{kim_2023_rolex} (4 experts). 
The size of our expert panel is also larger than or comparable to that of expert panels used in clinical diagnostic research — ``\textit{Most panels used two members (29 of 63 papers, 46\%), followed by three members (18 of 63 papers, 29\%). The maximum reported number of members was nine.}'' \citep{bertens_2013_use}.
Our medical experts have an average of 9 years of clinical experience. We conducted a 20-minute interview with each of the medical experts independently; each interview consisted of a survey and a structured interview that lets the interviewee evaluate the interpretations and prototypes produced by our method.
The front part of the survey (introduction of the model and interpretation figures) is the same as the one used in the previous user study, as this part proved to be easy to understand by both expert and non-expert readers. We randomly split the medical experts into two equally-sized groups. Group 1 read our model's interpretation (Figure \ref{fig:user-study-temppnet-nhanes}), and group 2 read the baseline model's interpretation (Figure \ref{fig:user-study-protopnet-nhanes}). Following that, we asked 4 questions to evaluate various aspects of interpretability (The question wordings are adopted from \cite{kaur_2020_interpreting} and \cite{wang_2021_partially}): 1) How reasonable do you think the model's interpretation is? 2) How trustworthy do you think the model's interpretation is? 3) How clinically meaningful do you think the model's interpretation is? 4) How useful do you think the model's interpretation is for depression screening practices? All these questions used a scale of 0-4 (0 = ``not at all'' and 4 = ``extremely'').

After the survey, we told each expert there was a model that had learned typical symptom prototypes and trend prototypes from a dataset of wearable sensor signals. We then presented to each expert symptom prototypes produced by our model (Figure \ref{fig:symptom-results-nhanes}) and trend prototypes generated by our model (Figure \ref{fig:trend-results-nhanes}). Finally, we asked each expert three questions: 1) Are these prototypes clinically meaningful when interpreting the depression of patients? 2) If using these prototypes to explain the depression case of a patient, are these prototypes useful in assisting your depression screening process? 3) Do you have any further comments on these prototypes and the model?

\begin{figure}[h]
    \centering
    \includegraphics[width=0.6\textwidth]{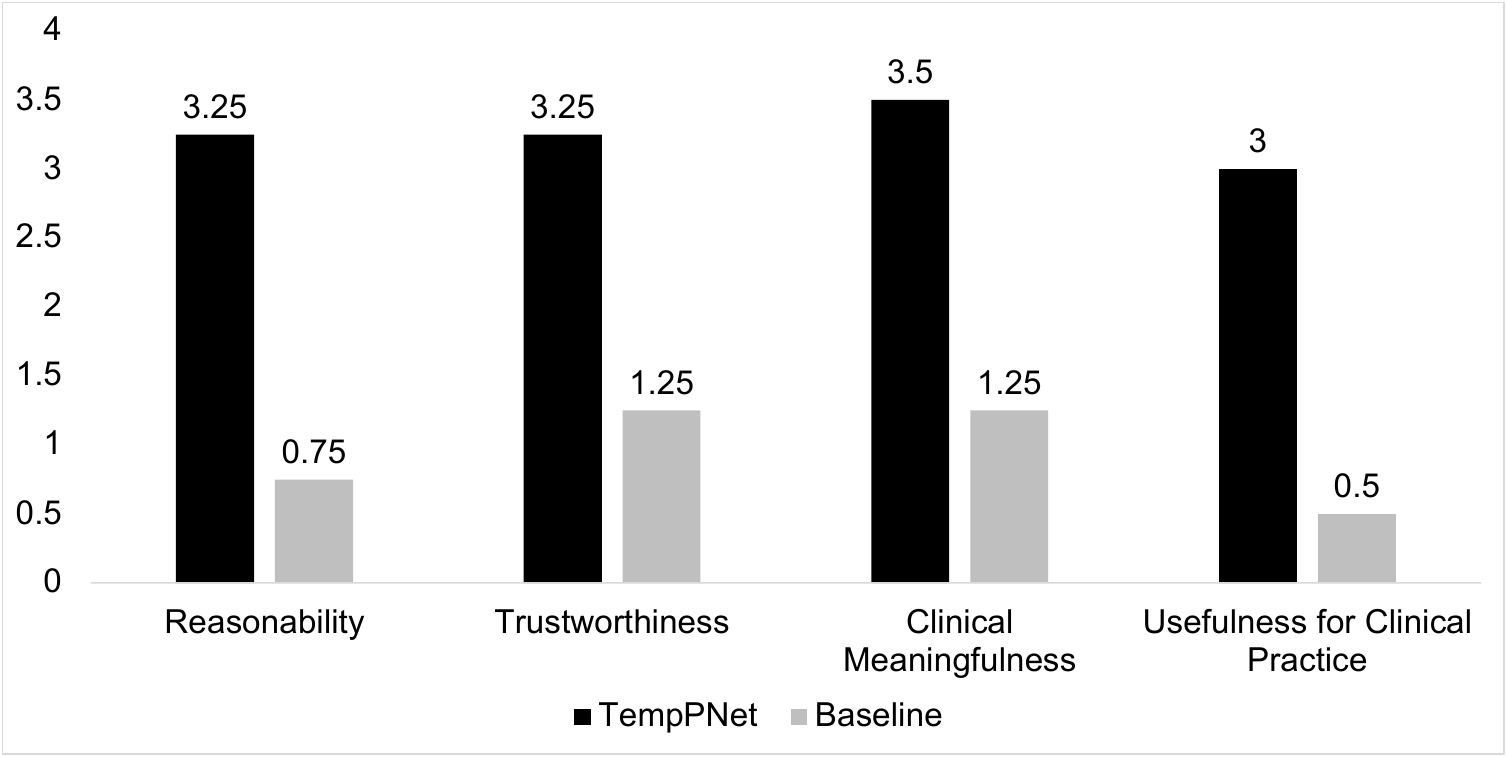}
    \caption{Expert Panel Survey Results (Mean in Each Group)}
    \label{fig:expert_panel}
\end{figure}

The survey and interview results reaffirm the better interpretability of our model. In terms of the survey (Figure \ref{fig:expert_panel}), 
for reasonability, trustworthiness, clinical meaningfulness, and usefulness for depression screening practices, the answers within the same group are pretty consistent — group 1 experts (presented with interpretations by our method) all rated 3 or 4 (on a scale of 0-4), and group 2 experts (presented with interpretations by the baseline method) all rated 0-2. This proves that our new trend prototypes indeed are meaningful and useful clinically. 
Experts who read the baseline interpretation in the survey commented unanimously that showing a snapshot of a symptom without knowing its temporal progression is unreliable in understanding the depression status of a patient, which further validates the importance of our introduced trend prototypes.
For the structured interview, all 8 experts agreed that prototypes produced our model (Figures \ref{fig:trend-results-nhanes} and \ref{fig:symptom-results-nhanes}) are clinically meaningful and useful in assisting their depression screening process. The experts further commented that our model, if implemented as an app, will be highly useful for their clinical practices. They noted that current depression diagnoses often involve subjectivity, and that patients usually do not describe their feelings well — e.g., whether they are feeling better or worse is difficult to be understood objectively by clinicians. Symptom and trend prototypes learned by our method using objective sensor data, on the contrary, provide an objective assessment of a patient's depression status and its trend, thereby assisting clinicians' understanding of the status and progress of a patient's depression and alleviating subjectivity. The experts also mentioned that using our model's interpretations in practice would significantly enhance their understanding of patients' depression conditions.

\vspace{-5pt} 
\section{Conclusion} 
\label{sec:discuss}

\subsection{Summary and Contributions}

Depression detection is fundamental to comprehensive chronic care in the health sensing area. Although a few studies tapped this domain, they do not offer a meaningful interpretation of their decisions. Prototype learning methods fall short in modeling the temporal progression of depression. To address these limitations, we propose a novel interpretable deep learning method to detect depression using sensor data while interpreting its decisions based on the temporal progression of depression. We conduct extensive evaluations to demonstrate superior detection performance of our method over state-of-the-art benchmarks and showcase its interpretation of decisions. Our human evaluations show that our method outperforms the benchmark in terms of interpretability.

Our work belongs to the computational genre of design science research in IS, which develops computational methods to solve business and societal problems and aims to make methodological contributions \citep{rai_editors_2017,simchi_2020_editor,padmanabhan_machine_2022}. In this regard, our proposed TempPNet, the designed IT artifact, is a novel prototype learning method that processes a time series of multivariate time series and interprets decisions based on prototypes of temporal symptom progression. To design TempPNet, we innovatively overcome three methodological challenges: learning prototypes of depression symptoms, learning prototypes of temporal symptom progression from walking segments performed at irregular time intervals, and modeling walking segments of varying lengths. This IT artifact design process reveals three general design principles: 1) Capturing the temporal evolution of physical patterns can improve the performance of predictive systems; 2) Identifying typical patterns of an existing class for prediction not only promotes model interpretability but also enhances predictive accuracy; 3) Learning representation from raw physical activities is more useful than feature engineering. These design principles offer computational IS scholars valuable references when searching through a solution space to design novel IT artifacts.

To the HIT literature, we validate the potential of leveraging sensing information technology for depression management amid regular treatments of chronic diseases. Our proposed novel interpretable deep learning method can detect depression associated with chronic diseases using motion sensor data and interpret its decisions. Applying our model, new business models can be developed to assist depression treatment for chronic disease patients in real-time.

\subsection{Managerial Implications and Future Work} \label{sec:implications}

Our study offers managerial implications for multiple key stakeholders, including doctors, caregivers, and healthcare systems. 
Doctors can encourage their chronic disease patients to install wellness trackers to actively monitor patients' walking data. The wellness tracker data can be analyzed by our method to detect depression risk on a daily basis. When our method signals a high depression risk, the doctors could review the interpretation given by our model and understand what depression walking symptoms the patient presents and how the patient's depression severity has progressed over time. Timely and personalized interventions can be taken to prevent the deterioration of depression and the negative spill-over effects for physical chronic symptom treatments. 
Physicians are also concerned about their patient's satisfaction. If the model detects depression correctly, patients' satisfaction will be higher. 
Our empirical results show that our model significantly outperforms the baselines. Therefore, the satisfaction of our model is likely to be higher than the baseline models.

Caregivers' home care is one of the key determinants of successful chronic disease management. Our method offers an opportunity for caregivers to keep track of the daily mental health updates of their loved ones. Caregivers can use our sensor-based depression detection mobile system for patients as well as themselves. When patients' mental health deteriorates, the caregivers can actively seek medical help, provide emotional support, and look for social support groups for the patients. Taking care of chronic disease patients can be stressful. This depression tracker on their own mobile devices serves as a reminder for self-care.
Patients themselves can be aware of their mental health status and be proactive in their health management.

Since depression poses a significant economic burden, our method enables active monitoring of depression and timely intervention, thus allowing for better resource allocation and potentially saving untreated depression costs. There are an estimated 307 million smartphone users in the US \citep{statista_topic_2022}, among which 62\% (190.34 million) use m-health apps that track users' motion data to monitor health conditions \citep{vicert_2021_HealthApps}. These are consenting users for collecting and using their physical data captured by motion sensors. Nearly 60\% of US adults have at least one chronic disease \citep{Hoffman_2022_chronicdisease}. Therefore, there are an estimated $190.34\times60\%=114.20$ million consenting m-health app users who suffer from chronic diseases. They represent 34.71\% of the US's 329 million population. The annual direct healthcare cost of untreated mental illnesses in the state of Indiana is \$708.5 million \citep{taylor_2023_economic}. Depression healthcare costs account for around 24.82\% of mental illness healthcare costs \citep{apa_2021_economic,ahqr_2022_healthcare}. Therefore, the annual healthcare cost of untreated depression in the state of Indiana is estimated to be \$708.5 million $\times 24.82\% = \$175.85$ million. Based on the population ratio of Indiana to the US \citep{britannica_2023_USStates}, the annual healthcare cost of untreated depression in the US is estimated to be \$8.58 billion, among which the chronically-ill consenting m-health app users account for $8.58\times34.71\%=\$2.98$ billion. 
Given such a sizable cost of untreated depression, using our model wisely could have a considerable positive impact on healthcare systems. Specifically, the probability of depression predicted by our model can serve as input for downstream optimization models, which allocate limited medical resources among healthcare systems to maximize the number of depression cases treated.

Our study has limitations and can be extended in various ways.
Our proposed method focuses on walking activities, while other activities, such as work productivity and sleep quality, can also serve as indicators of depression. Future research could enhance our method by incorporating data on these other activities, although obtaining such data would incur additional cost and require further user consent.
In addition, depression detection is one problem that can be effectively solved by our method. It can be adapted to tackle challenges in many other areas, such as mobile analytics, health information technology, investment portfolio choice, and social media analytics. A detailed discussion of our method's implications for these areas is provided in Appendix \ref{apd:managerial_for_other_area}. 
Future research could collect time series data and adapt our method to address important challenges in these areas.

\section{Acknowledgments}
Jiaheng Xie and Xiao Fang are supported by the University of Delaware Research Foundation Strategic Initiatives (UDRF-SI) Grant and Alfred Lerner College of Business and Economics Research Grant. Jiaheng Xie and Xiao Fang are not supported by any other funds. 
Xiaohang Zhao is supported by the Fundamental Research Funds for the Central Universities: ``High-Quality Development of Digital Economy: An Investigation of Characteristics and Driving Strategies (Grant Number 2023110139)''
and ``Intelligent Decision-Making Theories and Methods for Online Digital Platforms (Grant Number 2023110318).''

\clearpage

\bibliographystyle{informs2014} 
\bibliography{Project-IDDvSD.bib} 
\clearpage

\renewcommand{\theequation}{\arabic{equation}} 
\renewcommand{\thetable}{\arabic{table}}

\ECSwitch

\ECHead{\center Supplementary Materials}

\vspace{0.4cm}

\begin{center}
\textbf{Care for the Mind Amid Chronic Diseases: An Interpretable AI Approach Using IoT}    
\end{center}

\hspace{0.2cm}
\begin{center}

Jiaheng Xie$^{1,**}$, Xiaohang Zhao$^{2,*,**}$, Xiang Liu$^{1}$, Xiao Fang$^{1}$ \\
\vspace{0.2cm}
$^1$ Lerner College of Business and Economics, \\
University of Delaware, Newark, DE, USA \\

$^2$ School of Information Management \& Engineering, \\ Shanghai University of Finance and Economics, Shanghai, China \\

\vspace{0.1cm}
$*$ Corresponding Author: Xiaohang Zhao, \href{mailto:xiaohangzhao@mail.shufe.edu.cn}{xiaohangzhao@mail.shufe.edu.cn} \\
$**$ Equal Contribution \\
\end{center}
\vspace{0.3cm}

\begin{appendices}

\section{Recent Health Sensing Studies}\label{apd:health_sensing_lit}

Table \ref{tb:healthsensing} shows the recent health sensing studies related to our study. Using walking sensor data as the input, existing studies have applied machine and deep learning methods, such as convolutional neural networks (CNNs), support vector machines (SVMs), and random forests (RFs), to detect the occurrence and severity of chronic diseases.

\begin{table}[h]
\centering
\caption{Summary of Recent Health Sensing Studies}
\label{tb:healthsensing}
\small
\begin{threeparttable}
\begin{tabular}{L{110pt}L{70pt}L{50pt}L{100pt}L{90pt}}
\toprule
 Study & Data & Number of Subjects & Task & Model \\ \midrule
 \citesec{anand_listener_2015_a} & Walking tests & 25 & PD detection & Regression, NB, RF\\
 \citesec{um_data_2017_a} & Walking tests & 30 & PD motor detection & CNN \\
 \citesec{millor_gait_2017_a} & Walking tests & 431 & Frailty prediction & Decision tree \\
 \citesec{watanabe_development_2017_a} & Walking tests & 12 & Diabetes forefoot load detection & Statistical analysis \\
 \citesec{polat_hybrid_2019_a} & Walking tests & 16 & PD prediction & Regression \\
\citesec{coelln_quantitative_2019_a} & Walking tests & 683 & PD prediction & Cox \\
\citesec{rastegari_machine_2019_a} & Walking tests & 43 & PD diagnosis & SVM, RF, NB, AdaBoost \\
 \citesec{nemati_estimation_2020_a} & Cough and speech & 21 & Lung disease prediction & Regression, SVM, RF, MLP  \\
 \citesec{moon_classification_2020_a} & Walking tests & 524 & PD prediction & NN, SVM, KNN, decision tree, RF \\
\citesec{piau_when_2020_a} & Walking tests & 125 & Fall detection & Regression \\

 \bottomrule
\end{tabular}
\end{threeparttable}
\end{table}

\section{Recent Prototype Learning Studies}\label{apd:prototype_learning_lit}

Table \ref{tb:prototypelearning} shows the recent prototype learning studies. Prototype learning has been extended to interpret text classification. In this vein, \cite{ming_interpretable_2019} propose ProSeNet, which adds the prototype layer after the sequence encoder (e.g., RNN). This model is able to predict the class of a sentence (e.g., positive or negative) and explain which part of the sentence (prototype) leads to such a prediction result. 
A number of prototype learning models have also been proposed for various tasks. For example, \cite{rymarczyk_protopshare_2021} develop ProtoPShare that captures sharing property between each pair of prototypes. \cite{nauta_neural_2021} combine decision trees and prototype learning so that the prototype reasoning process can be streamlined as a tree structure. \cite{singh_these_2021} design two groups of prototypes: one group that the input looks like and the other group that the input does not look like. 

\begin{table}[h]
\centering
\caption{Existing Prototype Learning Methods v.s. Our Method}
\label{tb:prototypelearning}
\small
\begin{threeparttable}
\begin{tabular}{L{105pt}L{185pt}L{140pt}L{15pt}}
\toprule
  Study & Novelty & Input &  TP\tnote{*}\\ \midrule
 \citesec{chen_this_2019_a}& Prototype for image classification & An image &  No\\
 \citesec{hase_interpretable_2019_a}&  Hierarchical prototype & An image &  No\\
 \citesec{ming_interpretable_2019_a}&  Prototype for text classification & A piece of text & No \\
 \citesec{xu_attribute_2020_a}& Represent attribute for zero-shot learning & An image &  No \\
 \citesec{shitole_one_2021_a}&  Attention maps to explain a classifier & An image & No \\
 \citesec{rymarczyk_protopshare_2021_a}& Prototype parts share & An image & No \\
 \citesec{nauta_neural_2021_a}&  Prototype and decision tree & An image & No \\
 \citesec{wang_interpretable_2021_a}&  Add embedding space using manifold & An image & No \\
 \citesec{singh_these_2021_a}&  Positive reasoning and negative reasoning & An image & No \\
 \citesec{nauta_this_2021_a}&  Generate textual info about prototypes & An image & No \\
 Our Method&  Capture temporal progression of the input & A sequence of walking segments & 
 Yes \\ \bottomrule
\end{tabular}
\begin{tablenotes}\footnotesize
    \item[*] TP stands for ``Temporal Progression'', indicating whether a model is capable of detecting interpretable temporal progression patterns from its input and then leveraging these patterns for prediction and interpretation. Depression symptoms exhibit a temporal progression, such as in Figure \ref{fig:depression_trend}.
\end{tablenotes}
\end{threeparttable}
\end{table}

\section{MTSC Studies}\label{apd:mtsc_lit}

Table \ref{tb:mtsc} shows the MTSC studies. The distance-based models usually use 1-nearest neighbor coupled with a bespoke distance function. Different from cross-sectional data, the distance between multivariate time series can be computed using dynamic time warping (DTW) \citepsec{shokoohi_2017_generalizing_a}. Shapelets are discriminatory sub-series that have practical meaning. The shapelets-based models use selected shapelets using random forests \citepsec{karlsson_2016_generalized_a}. For the histogram-based models, words in the form of unigrams and bigrams are extracted for all time series and dimensions using a sliding window for a range of window lengths. The words for each dimension and window length are concatenated into a single bag of words histogram for a series \citepsec{schafer_2017_multivariate_a}. Interval summarizing-based models, such as the Canonical Interval Forest is an ensemble of time series trees built using the Canonical Time-Series Characteristics features and simple summary statistics extracted from phase dependant intervals \citepsec{middlehurst_2020_canonical_a}. For deep learning-based models, various time series encoders, such as ResNet \citepsec{wang_2017_time_a} and InceptionTime \citep{ismail_2020_inceptiontime}, are adopted to represent the multivariate time series data. Then, multiple layers of deep learning models can be deployed for classification.

\begin{table}
\centering
\caption{Existing Multivariate Time Series Classification Methods v.s. Our Method}
\label{tb:mtsc}
\small
\begin{threeparttable}
\begin{tabular}{L{115pt}L{70pt}L{195pt}L{55pt}}
\toprule
  Study &  Category & Input & Interpretable\\ \midrule
 \citesec{karlsson_2016_generalized_a}&  Shapelets & A multivariate time series of UCR data &  No\\
 \citesec{shokoohi_2017_generalizing_a}&  Distance measures & A multivariate time series of cricket umpire signals &  No\\
 \citesec{schafer_2017_fast_a}&  Histogram & A multivariate time series of UCR data &  No\\
 \citesec{schafer_2017_multivariate_a}&  Histogram & A multivariate time series &  No\\
 \citesec{bagnall_2020_usage_a}& Distance measures & A multivariate time series of UCR data &  No\\
 \citesec{middlehurst_2020_canonical_a}& Interval summarising & A multivariate time series of UCR data  &  No\\
 \citesec{dempster_2020_rocket_a}&  Interval summarising & A multivariate time series of UCR data &  No\\
 \citesec{ismail_2020_inceptiontime_a}&  Deep learning & A multivariate time series of synthetic data &  No\\
 Our Method &  Deep learning & A time series of irregularly spaced walking segments. Each is a multivariate time series &  Yes \\ \bottomrule
\end{tabular}
\end{threeparttable}
\end{table}

\section{Prototype Learning for MTSC Methods}\label{apd:proto_mtsc_lit}

Table \ref{tb:mtscpl} shows the prototype learning for MTSC studies. Although \citesec{ming_interpretable_2019_a} does not directly tackle the MTSC problem, its text sequence model can be adapted to process MTSC-based prototype learning problems. The learned prototype is a sentence.
\citesec{ma_2020_interpretable_a} apply prototype learning to interpret the MTSC classification of vital signs. The time series vital signs are processed as an image. Multiple CNN layers and a prototype layer are used to predict Myocardial infarction. The learned prototypes are a segment of vital signs.
\citesec{zhang_2020_tapnet_a} devise TapNet for MTSC problems with high dimensionality and limited training data issues. TapNet leverages a low-dimensional feature extractor to reduce the dimension from the multivariate time series. To address the limited training data issue, the authors propose a Random Dimension Permutation as a data augmentation mechanism. As multiple data sources are concatenated and a black-box LSTM layer is deployed in the feature extractor, the prototype cannot be traced back to a local region of the input, thus hindering the model's interpretability. Consequently, \citesec{zhang_2020_tapnet_a} only focus on prediction. 
\citesec{ghosal_2021_multi_a} develop a framework for interpretable MTSC. The multivariate time series is first separated into multiple univariate time series. For each variable, an LSTM encoder extracts a representation for its time series. A prototype layer is stacked next to learn typical patterns from each variable independently. In the end, the prototype similarities from each variable are concatenated to make the classification. In the interpretation phase, prototypes of each variable are shown independently, which are a segment of the univariate time series. 
\citesec{trinh_2021_interpretable_a} propose DPNet to detect deep fake videos. Videos are a special format of multivariate time series where the dimensions are the image channels. An encoder represents each video as a single tensor, which is compared with the prototypes. These prototypes are embeddings of typical deep fake videos. Unlike most other prototype learning studies (model-based interpretation), DPNet's interpretation is independent of its prediction process, conducted in the post-training phase (post-hoc interpretation). The authors use Timed Quality Temporal Logic (TQTL), which is an induction method. Using the TQTL, the authors pick a clip of a video that resembles most of a fake video.

\begin{table}
\centering
\caption{Existing Prototype Learning for MTSC Methods v.s. Our Method}
\label{tb:mtscpl}
\small
\begin{threeparttable}
\begin{tabular}{L{90pt}L{150pt}L{125pt}L{55pt}}
\toprule
  Study &  Input & Prototype & Temporal Progression of Prototype \\ \midrule
 \citesec{gee_2019_explaining_a}&  A multivariate time series of ECG & A segment of ECG signal &  No\\
 \citesec{ming_interpretable_2019_a}  & A sentence & A sentence & No \\
 \citesec{ma_2020_interpretable_a}  & A multivariate time series of vital signs  & A segment of vital sign signal & No \\
 \citesec{zhang_2020_tapnet_a}&  A multivariate time series of ECG & A hidden embedding &  No\\
 \citesec{ghosal_2021_multi_a}&  A four-dimensional time series of simulated data & A segment of a univariate series &  No\\
 \citesec{trinh_2021_interpretable_a}&  Videos & A clip of a video &  No\\
 Our Method &  A time series of irregularly spaced multivariate time series (i.e., walking segments) & 1) A region of sensor signal (symptom); 2) progression of symptom (trend) & Yes \\ \bottomrule
\end{tabular}
\end{threeparttable}
\end{table}

\section{Deep Learning Models Considering Time Irregularity}\label{apd:time_model_lit}

Table \ref{tb:time_models} shows the deep learning models considering time irregularity. Their mechanisms of incorporating time fall into three approaches. The first approach utilizes a continuous time function to model the time series data, so that the irregularly spaced temporal input can be implicitly considered \citepsec{li_predicting_2020_a}. An example of this category is \citesec{li_predicting_2020_a} who design an ordinary differential equation to model the temporal walking physical symptoms of Parkinson's disease. The second approach modifies the sequence feature extraction model (e.g., LSTM) and adds the time interval between consecutive states in the cell state \citepsec{baytas_patient_2017_a,zhang_time-aware_2020_a,gao_distanced_2019_a}. \citesec{baytas_patient_2017_a} use this approach to predict patient subtyping using EHR. The third approach proposes new temporal embeddings and adds them as additional features to the model \citepsec{li_ea-lstm_2019_a,liu_td-lstm_2018_a,mei_transformer_2022_a}. For instance, \citesec{mei_transformer_2022_a} devise a time-varying embedding that is sensitive to time changes.

\begin{table}[h]
\centering
\caption{Deep Learning Models Considering Time Irregularity}
\label{tb:time_models}
\small
\begin{threeparttable}
\begin{tabular}{L{55pt}L{65pt}L{140pt}L{120pt}L{55pt}}
\toprule
 Study & Context & Mechanism of Incorporating Time & Purpose & Interpretable \\ \midrule
 \citesec{li_ea-lstm_2019_a} & Times series prediction & Transfer different attention weights to different timesteps & Diversify the attention for time series prediction &  No \\
 \citesec{gao_distanced_2019_a} & Cancer detection & Consider time intervals in LSTM & Encode irregular timepoint input to prediction & No  \\
 \citesec{li_predicting_2020_a} & Parkinson's detection & Ordinary differential equations to model time series & Encode time dimension to prediction &  No \\
 \citesec{baytas_patient_2017_a} & Patient subtyping & Add time interval in LSTM cell state & Encode irregular timepoint input to prediction &  No \\
 \citesec{liu_td-lstm_2018_a} & Temperature prediction & Design closeness, period, and trend as additional features & Encode temporal input to prediction &  No \\
 \citesec{mei_transformer_2022_a} & Time series prediction & Design time-varying embedding as additional features & Encode time dimension to prediction & No  \\
 \citesec{zhang_time-aware_2020_a} & Health state detection & Add time interval in LSTM cell state & Encode irregular timepoint input to prediction &  No \\
 Ours & Depression detection & Design continuous temporal prototypes & Model a time series of irregularly spaced multivariate time series to improve interpretation &  Yes \\
 \bottomrule
\end{tabular}
\end{threeparttable}
\end{table}

\section{The Generative Process of $S$ Characterized by Trend Prototype $k$}\label{apd:algorithm}

Algorithm \ref{alg:gen_enrich} shows the generative process of $S$ characterized by trend prototype $k$.

\begin{algorithm}[h]
\caption{The Generative Process of $S$ Characterized by Trend Prototype $k$}\label{alg:gen_enrich}
\begin{algorithmic}[1]
\State Compute $t_0^{(k)}$ via Equation \ref{eq:infer_t0}.
\State Compute $t_i^{(k)} = t_i - t_0^{(k)}$ for $i=1,2,\dots,N$.
\State Compute ${\tilde{\mathcal{G}}}^{(k)}(t_i^{(k)})$ for $i=1,2,\dots,N$ using Equation \ref{eq:g_k_t_tilde}.
\State Draw $S_{i} \sim \mathcal{L} \mathcal{N}({\tilde{\mathcal{G}}}^{(k)}(t_i^{(k)}), I)$ for $i=1,2,\dots,N$.
\end{algorithmic}
\end{algorithm}

\section{The Definition of Sensor Features}\label{apd:sen_features}

\begin{figure}[h]
    \centering
    \includegraphics[width=0.7\textwidth]{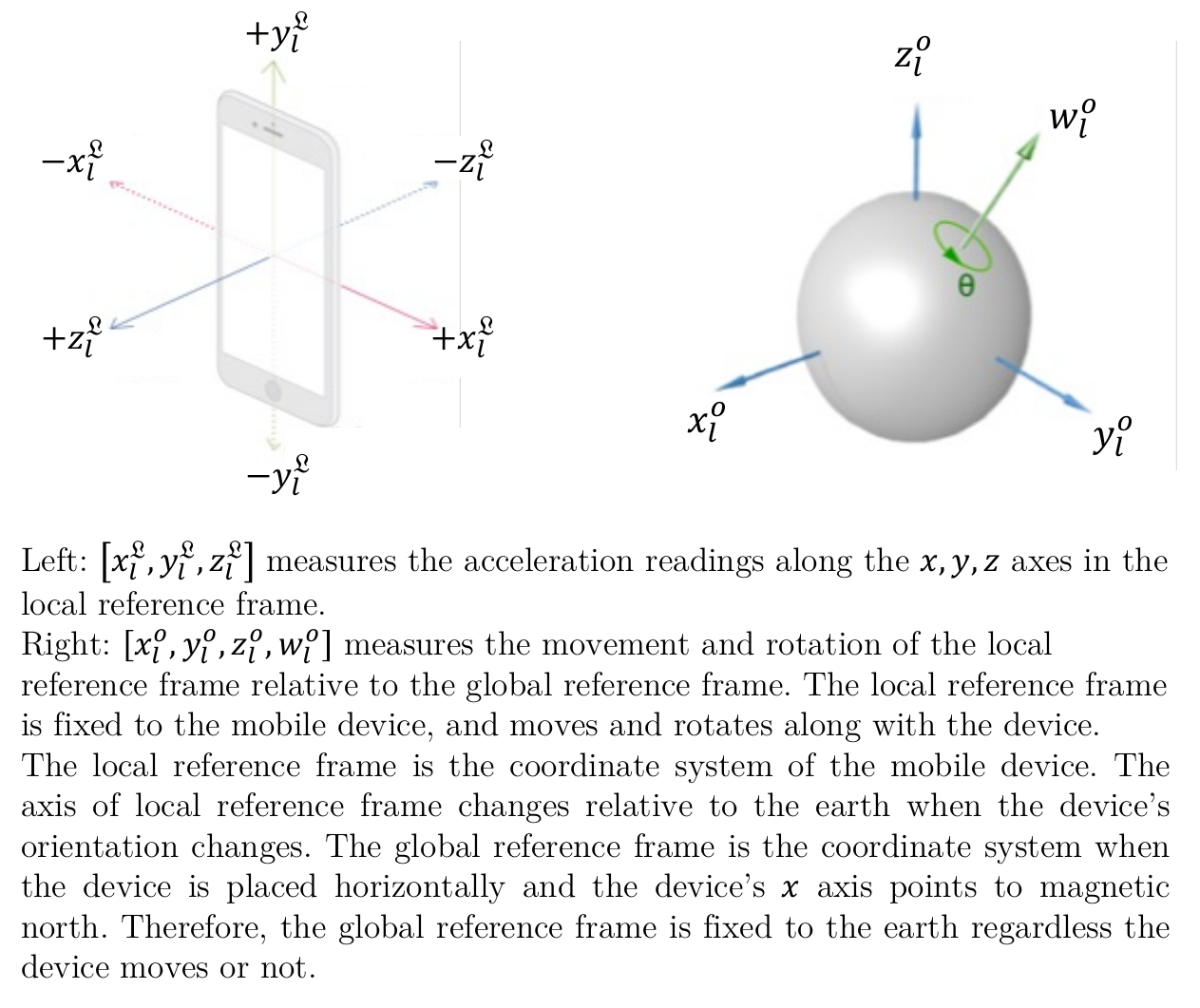}
    \caption[]{Sensor Readings \citepsec{apple_inc_understanding_2022_a} }
    \label{fig:frame_change}
\end{figure}

We explain the content of a sensor feature as mentioned in Section \ref{sec:problem_setup}.
At each timepoint $l$, the mobile sensor collects accelerometer readings $[x_l^{\mathfrak{L}},y_l^{\mathfrak{L}},z_l^{\mathfrak{L}}]$ and orientation readings $[x_l^{o},y_l^{o},z_l^{o},w_l^{o}]$. A graphic illustration of these sensor readings is shown in Figure \ref{fig:frame_change}.

The accelerometer readings are in the local reference frame, which most existing studies rely on \citepsec{piau_when_2020_a,yu_wearable_2022_a,zhu_deep_2021_a}. However, these local reference readings do not reflect the precise walking pattern in the geographic coordinate system, because it moves and rotates along with the mobile device. To address this issue and make the interpretation more meaningful in the geographic sense, we decide to work with the readings in the global reference frame. We transform the local reference frame accelerometer vector $v_l^{\mathfrak{L}} =[x_l^{\mathfrak{L}},y_l^{\mathfrak{L}},z_l^{\mathfrak{L}}]^T$ to the global reference frame as $v_l^\mathfrak{G} =[x_l^{\mathfrak{G}},y_l^{\mathfrak{G}},z_l^{\mathfrak{G}}]^T$ via the quaternion rotation 
$$v_l^{\mathfrak{G}} = \mathcal{R}_l v_l^{\mathfrak{L}}$$ 
where $\mathcal{R}_l$ is the rotation matrix derived from quaternion $[x_l^{o},y_l^{o},z_l^{o},w_l^{o}]$ as follows. In general, given a quaternion $[x,y,z,w]$, the corresponding rotation matrix $\mathcal{R}$ is defined as\footnote{\url{https://en.wikipedia.org/wiki/Quaternions_and_spatial_rotation}}
\begin{equation}
\label{eq:rotation_matrix}
   \mathcal{R} = \begin{bmatrix}
       w^2+x^2-y^2-z^2 & 2 x y-2w z & 2 x z+2 w y \\  
        2 x y+2 w z & w^2-x^2+y^2-z^2 & 2 y z-2w x \\
       2 x z-2w y & 2 y z + 2 w x & w^2 - x^2- y^2 + z^2
   \end{bmatrix}.
\end{equation}
Then, we define the sensor feature $a_l$ as
\begin{equation}
a_{l}=[x_l^{\mathfrak{G}},y_l^{\mathfrak{G}},z_l^{\mathfrak{G}} ]^T.
\end{equation}

Following the best practice of data augmentation for mobile sensor data \citepsec{um_data_2017_a,zhang_deep_2020_a}, during the training stage, we further transform the input sensor features with random rotations to improve the generalization ability of our model. To do this, we first sample a quaternion using Algorithm \ref{alg:sample_quaternion} where $\text{Uniform}(b_1,b_2)$ means the uniform distribution on the interval $[b_1,b_2]$, and then plug the obtained quaternion into Equation \ref{eq:rotation_matrix} to construct a random rotation matrix. Within each training epoch and for each walking test, we construct a random rotation matrix $\tilde{\mathcal{R}}$ as described, and then use it to transform all sensor features of the walking test as $\langle \tilde{\mathcal{R}} a_1, \tilde{\mathcal{R}} a_2, \dots, \tilde{\mathcal{R}} a_L \rangle$. Once the training stage is finished, we use the original sensor features $\langle a_1, a_2, \dots, a_L \rangle$ to do inference.

\begin{algorithm}[h]
\caption{Sample a Quaternion}\label{alg:sample_quaternion}
\begin{algorithmic}[1]
\State Draw $x \sim \text{Uniform}(0,1)$, $y \sim \text{Uniform}(0,1)$, $z \sim \text{Uniform}(0,1)$.
\State Let $\text{norm}=\sqrt{x^2+y^2+z^2}$, and set $x=x/\text{norm}$, $y=y/\text{norm}$, $z=z/\text{norm}$.
\State Draw $\theta \sim \text{Uniform}(0,2\pi)$.
\State Set $w=\cos(\theta/2)$, $x=x \sin(\theta/2)$, $y=y \sin(\theta/2)$, $z=z \sin(\theta/2)$
\State Return $[x,y,z,w]$
\end{algorithmic}
\end{algorithm}

\section{The Density Function of the Logistic-Normal Distribution}\label{apd:der_logistic_normal}

We employ the change of variables formula \citepsec{murphy_probabilistic_2022_a} to derive Equation \ref{eq:sigmoid_gauss_density}, which in general states that if a random vector $x \in R^{M}$ is mapped to another random vector $z \in R^{M}$ by an invertible
function $f$, i.e., $z=f(x)$, then the density function of $z$, denoted by $p_z(z)$, is related to the density function of $x$, denoted by $p_x(x)$, by the following relationship:
\begin{equation}
\label{eq:change_of_var}
p_z(z) = p_x\big( g(z) \big) \big| \det[J_g(z)] \big|
\end{equation}
where $g$ is the inverse function of $f$, $J_{g}(z) \in R^{M \times M}$ is the Jacobian matrix of $g$ evaluated at $z$, $\det [.]$ is the matrix determinant operator, and $|.|$ is the absolute value operator. 

In our case, $z=\sigma(x)$, which means that $f=\sigma$ and $g=\sigma^{-1}$. Using the fact that $\partial x_m/\partial z_m = 1 / (z_m(1-z_m))$, the corresponding Jacobian matrix can be computed as
\begin{equation}
J_{g}(z) =  \begin{pmatrix}
    \frac{1}{z_1 (1-z_1)} & 0 & \dots & 0 \\
    0 & \frac{1}{z_2 (1-z_2)} & \dots & 0 \\
    \vdots & \vdots & \ddots & \vdots \\
    0 & 0 & \dots & \frac{1}{z_M (1-z_M)}
  \end{pmatrix},
\end{equation}
which is a diagonal matrix because $\sigma^{-1}$ has been element-wisely applied on $z$. Given that $0< z_m < 1$ for $m=1,2,\dots,M$, all the diagonal elements are positive, and thus we have 
\begin{equation}
\label{eq:jacobain_det}
   \big| \det[J_g(z)] \big|  = \frac{1}{\prod_{m=1}^{M} z_{m}(1-z_{m})}.
\end{equation}
Recall that $x \sim \mathcal{N}(\mu, \Sigma)$, then the corresponding density function is given by 
\begin{equation}
\label{eq:gauss_density}
p_x(x) = \frac {\exp \big(-\frac{1}{2}(x-\mu)^{T}{\Sigma}^{-1}(x-\mu) \big)}{ \sqrt{(2 \pi)^{M} \det[\Sigma] }}.
\end{equation}
Plugging Equation \ref{eq:jacobain_det} and \ref{eq:gauss_density} into Equation \ref{eq:change_of_var}, and setting $x=\sigma^{-1}(z)$, we obtain Equation \ref{eq:sigmoid_gauss_density}.

\section{NHANES Dataset} \label{apd:nhanes_dataset}

NHANES is designed to assess the health and nutritional status of adults and children in the United States. The survey is unique in that it combines interviews and physical examinations for a nationally representative sample of all ages. To produce reliable statistics, NHANES over-samples persons 60 and older, African Americans, and Hispanics. The NHANES interview includes demographic, socioeconomic, dietary, and health-related questions. The examination component consists of medical, dental, and physiological measurements, as well as laboratory tests administered by highly trained medical personnel.\footnote{https://www.cdc.gov/nchs/nhanes/index.htm} 
These medical measurements offer accurate labels for a variety of chronic diseases, such as high blood pressure, heart disease, diabetes, chronic kidney disease, asthma, arthritis, stroke, and cancer. Among the health measurements, depression is one of the prioritized diseases of NHANES. NHANES deploys PHQ-9 to collect patients' depression diagnoses, making it a clinically accurate and relevant dataset for depression research \citepsec{yu_2020_trends_a,vallance_2011_associations_a}. Because of such a wide range of chronic disease coverage as well as precise measurement of depression, this dataset is an ideal testbed for our study.

In order to invite research to examine the impact of physical activities on chronic diseases, NHANES added wearable sensor data collection in conjunction with depression and other chronic disease diagnoses in the 2013 cohort. For the wearable sensor data collection, participants were first mailed a sensor device and wore it for continuous seven days. Timestamps were captured to ensure the sensor data were indeed recorded during this study window. This is sufficiently long to depict the severity change of patients' depression, as the depression severity assessment window recommended by the American Psychiatric Association is seven days\footnote{\href{https://www.psychiatry.org/File\%20Library/Psychiatrists/Practice/DSM/APA_DSM5_Severity-Measure-For-Depression-Adult.pdf}{Severity Measure for Depression (American Psychiatric Association)}}. The device was the ActiGraph model GT3X+, which measured acceleration every 1/80th of a second (80 Hz). 
Because of such fine-grained wearable sensor data, accurate clinical depression diagnoses from PHQ-9, and the large and representative participant base, we choose the NHANES dataset for all the following empirical analyses.

To construct our input data, we utilize the accelerometer data. The data were collected continuously. However, not all segments are usable. Patients may not wear the device from time to time, take a shower, sleep, sit while working, and so on. These time segments are not walking data and are not applicable to our study. Therefore, we use the state-of-the-art walking activity detector to filter the walking segments \citepsec{czech2019gaitpy_a}. We acknowledge that other activities other than walking, such as sleep quality and work productivity, may also indicate depression. Measuring these activities requires different sensors other than accelerometers. From the practical point of view, imposing such hardware and additional data collection requirements results in additional consensus from users and excludes a large portion of low-income users whose devices are not equipped with those types of sensors. Those device restrictions, hardware deficiency, increased cost, and consensus issues will eventually impede the successful implementation and equitability of the resulting detection model.
Our walking-based detection, on the other hand, is the easiest to implement as such functions can be deployed to existing m-health apps where location consensus is already obtained and accelerometers are readily installed in most mobile devices.

\section{Benchmark Methods and Hyperparameter Settings} \label{apd:benchmark}

According to our literature review, we select three groups of benchmarks. The first group is black-box deep learning models, CNN and RNN. They have been commonly used in prior motion sensor-based predictions \citepsec{yu_wearable_2022_a,zhu_deep_2021_a}. The second group uses manually crafted features, such as mean, variance, and standard deviation of sensor signals, as the input \citepsec{oung_wearable_2015_a,yu_wearable_2022_a}. These features are shown in Table \ref{tb:baseline_features}. The benchmarks in the second group and their features are in line with \citesec{yu_wearable_2022_a}, which includes k-nearest neighbors (KNN), support vector machine (SVM), random forest, AdaBoost, and XGBoost. 
The third group includes the state-of-the-art and the most widely recognized MTSC and prototype learning for MTSC models. From the MTSC studies, we select the most state-of-the-art study \citepsec{ismail_2020_inceptiontime_a} as a benchmark, since it is also deep learning-based, thus being able to learn representation from raw sensor data. It also achieves better performance than other MTSC studies. The other MTSC studies are traditional machine learning-based which require feature engineering. However, \citesec{ismail_2020_inceptiontime_a} is not interpretable. From prototype learning for MTSC studies, we select \citesec{ma_2020_interpretable_a,gee_2019_explaining_a,ming_interpretable_2019_a,chen_this_2019_a} as benchmarks, because they are the most state-of-the-art, and their input data format is the closest to sensor data. These are the most related benchmarks to this study. Compared to our model, these benchmarks cannot model or interpret the temporal progressions of prototypes. 
The hyperparameters of the benchmarks are summarized in Table \ref{tb:benchmark}. These hyperparameters are fine-tuned for each benchmark after large-scale experiments. The following evaluation results report the fine-tuned performances for the benchmarks.

\begin{table}[h]
\centering
\caption{Features for Conventional Machine Learning Models}
\label{tb:baseline_features}
\small
\begin{threeparttable}
\begin{tabular}{L{160pt}L{290pt}}
\toprule
 Feature Name & Formula \\ \midrule
 Mean x-axis values &  $u_{x} = \frac{1}{L} \sum_{l=1}^{L} v_{x,l}$ \\
 Mean y-axis values &  $u_{y} = \frac{1}{L} \sum_{l=1}^{L} v_{y,l}$ \\
 Mean z-axis values &  $u_{z} = \frac{1}{L} \sum_{l=1}^{L} v_{z,l}$ \\
 St. D. of x-axis values &  $\sigma_{x} = \sqrt{\frac{1}{L-1} \sum_{l=1}^{L} (v_{x,l} - u_{x} )^2 } $ \\
 St. D. of y-axis values &  $\sigma_{y} = \sqrt{\frac{1}{L-1} \sum_{l=1}^{L} (v_{y,l} - u_{y} )^2 } $ \\
 St. D. of z-axis values &  $\sigma_{z} = \sqrt{\frac{1}{L-1} \sum_{l=1}^{L} (v_{z,l} - u_{z} )^2 } $ \\
 Mean magnitude &  $u_{v} = \frac{1}{L} \sum_{l=1}^{L} ||v_l||$, $||v_l||=\sqrt{v_{x,l}^2+v_{y,l}^2+v_{z,l}^2} $ \\
 St. D. of magnitude &  $\sigma_{v} = \sqrt{\frac{1}{L-1} \sum_{l=1}^{L} (||v_l|| - u_v)^2 } $ \\
 Mean x-axis jerk & $\alpha_{x} = \frac{1}{L-1} \sum_{l=1}^{L-1} d_{x,l}$, where $d_{x,l}=v_{x,l+1}-v_{x,l}$ \\
 Mean y-axis jerk & $\alpha_{y} = \frac{1}{L-1} \sum_{l=1}^{L-1} d_{y,l}$, where $d_{y,l}=v_{y,l+1}-v_{y,l}$  \\
 Mean z-axis jerk & $\alpha_{z} = \frac{1}{L-1} \sum_{l=1}^{L-1} d_{z,l}$, where $d_{z,l}=v_{z,l+1}-v_{z,l}$  \\
 St. D. of x-axis jerk &  $\beta_{x} = \sqrt{\frac{1}{L-2} \sum_{l=1}^{L-1} (d_{x,l} - \alpha_{x} )^2 } $ \\
 St. D. of y-axis jerk &  $\beta_{y} = \sqrt{\frac{1}{L-2} \sum_{l=1}^{L-1} (d_{y,l} - \alpha_{y} )^2 } $ \\
 St. D. of z-axis jerk &  $\beta_{z} = \sqrt{\frac{1}{L-2} \sum_{l=1}^{L-1} (d_{z,l} - \alpha_{z} )^2 } $ \\
 Mean jerk magnitude &  $\alpha_{d} = \frac{1}{L-1} \sum_{l=1}^{L-1} ||d_l|| $, where $||d_l||=\sqrt{d_{x,l}^2+d_{y,l}^2+d_{z,l}^2}$ \\
 St. D. of jerk magnitude & $\beta_{d} = \sqrt{\frac{1}{L-2} \sum_{l=1}^{L-1} (||d_l|| - \alpha_{d})^2 } $  \\
 Stride time variability on x-axis & \multicolumn{1}{p{290pt}}{
 (1) Identify signal peaks in x-axis, $[t_1,t_2,\dots,t_Q]$;\newline 
 (2) Identify stride times $[d t_1,d t_2,\dots,d t_{Q-1}]$, where $d t_i=t_{i+1}-t_i$; \newline 
 (3) Compute stride time variability $V_{x}=\sqrt{\frac{1}{Q-2} \sum_{i=1}^{Q-1} (d t_i - \overline{d t} )^2 }$ }  \\
 Stride time variability on y-axis & \multicolumn{1}{p{290pt}}{
 (1) Identify signal peaks in y-axis, $[t_1,t_2,\dots,t_Q]$;\newline 
 (2) Identify stride times $[d t_1,d t_2,\dots,d t_{Q-1}]$, where $d t_i=t_{i+1}-t_i$; \newline 
 (3) Compute stride time variability $V_{y}=\sqrt{\frac{1}{Q-2} \sum_{i=1}^{Q-1} (d t_i - \overline{d t} )^2 }$ }  \\
 Stride time variability on z-axis & \multicolumn{1}{p{290pt}}{
 (1) Identify signal peaks in z-axis, $[t_1,t_2,\dots,t_Q]$;\newline 
 (2) Identify stride times $[d t_1,d t_2,\dots,d t_{Q-1}]$, where $d t_i=t_{i+1}-t_i$; \newline 
 (3) Compute stride time variability $V_{z}=\sqrt{\frac{1}{Q-2} \sum_{i=1}^{Q-1} (d t_i - \overline{d t} )^2 }$ }  \\
 Stride time vairability on magnitude & \multicolumn{1}{p{290pt}}{
 (1) Identify signal peaks in magnitude, $[t_1,t_2,\dots,t_Q]$;\newline 
 (2) Identify stride times $[d t_1,d t_2,\dots,d t_{Q-1}]$, where $d t_i=t_{i+1}-t_i$; \newline 
 (3) Compute stride time variability $V_{v}=\sqrt{\frac{1}{Q-2} \sum_{i=1}^{Q-1} (d t_i - \overline{d t} )^2 }$ }  \\
 \bottomrule
\end{tabular}
\begin{tablenotes}\footnotesize
    \item Recall that given a walking segment, we observe a sequence of sensor signals $\langle v_1^{\mathfrak{L}}, v_2^{\mathfrak{L}}, \dots, v_L^{\mathfrak{L}} \rangle$, where $v_l^{\mathfrak{L}}$ is the accelerometer readings recorded in the local reference frame at timepoint $l$, as explained in Appendix \ref{apd:sen_features}. To simplify notation, we drop the superscript $\mathfrak{L}$, and write $v_l=[v_{x,l},v_{y,l},v_{z,l}]^T$.
Following \citesec{yu_wearable_2022_a}, we define the following features for each given walking segment, shown in Table \ref{tb:baseline_features}.
\end{tablenotes}
\end{threeparttable}
\end{table}

\begin{table}[h]
\centering
\caption{Benchmark Hyperparameter Settings}
\label{tb:benchmark}
\small
\begin{threeparttable}
\begin{tabular}{L{80pt}L{200pt}C{80pt}}
\toprule
 Model & Parameter & Values \\ 
 \midrule
 TempPNet   & CNN channels & $(256,512,256,128)$   \\
       &CNN kernel sizes &$8\times 1$    \\
       & Time encoding dimensions &$64$    \\
KNN & Number of neighbors & $5$\\
SVM & Regularization parameter &$1$\\
        &Kernel coefficient &$0.001$\\
Random Forest & Number of estimators &$100$\\
AdaBoost & Number of estimators &$50$ \\
XGBoost & Number of estimators &$100$ \\
    & Minimum loss reduction for further partition &$1.5$ \\
    & Subsample & $0.6$ \\
ProtoPNet & CNN channels &$(32,64,128,256)$ \\
       &CNN kernel sizes & $3\times 3$  \\
ProSeNet & GRU hidden units &$64$ \\
 \bottomrule
\end{tabular}
\end{threeparttable}
\end{table}

To implement our model, we adopt the following hyperparameter setting.
We set the embedding dimension of symptom prototypes as $n_e=128$, the time embedding dimension as $n_d=64$, the regularization weights as $\lambda_\mathbb{S}=0.1$ and $\lambda_\mathbb{T}=0.1$. We train our model with the Adam optimizer \citepsec{kingma_adam:_2015_a} with a learning rate of $0.001$ and a batch size of $32$. The specification of the CNN layer is as follows:

Recall that $X_i$ denotes the sequence of sensor features of the $i$th walking segment performed by a focal patient. In this section, we focus on extracting the feature matrix $H_i^\mathbb{S}$ from a single walking segment, and therefore drop the subscript $i$ to simplify the notation. 
In general, different walking segments have different lengths. In what follows, we assume that $X$ has been downsampled with the frequency of 10 Hz, while the treatment of other sampling frequencies should be adjusted proportionally.

To facilitate batch training, we reshape each segment into a matrix of size $3 \times 300$, by either padding it with zero columns if its length is smaller than $300$, or discarding the extra columns if its length is larger than $300$.
After the reshaping step, we treat each segment $X_j$ as an input of $3$ channels and length $300$, and then use the same CNN layer to extract a feature matrix of size $128 \times 5$ from $X_j$. The CNN layer is composed by a sequence of one-dimensional convolution (Conv1d) layers, each followed in order by a one-dimensional batch normalization layer (BatchNorm1d), a max pooling layer (MaxPool1d), and lastly a leaky ReLU layer (LeakyReLU) with slope $0.01$ for non-linear activation \citepsec{goodfellow_deep_2016_a}. Following the style of \citesec{zhu_deep_2021_a}, we report the detailed architecture of the CNN layer in Table \ref{tb:cnn_layer}. Since the BatchNorm1d layer and the LeakyReLU layer do not change the input shape, we do not list them in Table \ref{tb:cnn_layer}. After the last Conv1d layer in Table \ref{tb:cnn_layer}, we do not add the MaxPool1d layer nor the LeakyReLU layer.

\begin{table}[h]
\centering
\caption{The Specification of the CNN Layer}
\label{tb:cnn_layer}
\small
\begin{threeparttable}
\begin{tabular}{L{50pt}C{40pt}C{30pt}C{40pt}C{60pt}}
\toprule
 & Kernel Size & Stride & Output Channel & Output Shape \\ \midrule
 Conv1d & 8 & 1 & 256 & (256, 293) \\
 MaxPool1d & 2 & 2 & & (256, 146) \\
 
 Conv1d & 8 & 1 & 512 & (512, 139) \\
 MaxPool1d & 2 & 2 & & (512, 69) \\
 
 Conv1d & 8 & 1 & 256 & (256, 62) \\
 MaxPool1d & 2 & 2 & & (256, 31) \\
 
 Conv1d & 8 & 1 & 128 & (128, 24) \\
 MaxPool1d & 2 & 2 & & (128, 12) \\
 
 Conv1d & 8 & 1 & 128 & (128, 5) \\
 \bottomrule
\end{tabular}
\end{threeparttable}
\end{table}

\section{Evaluations Conditioned on Pre-existing Chronic Disease Severity (NHANES)} \label{apd:eval_disease_severity_nhanes}

Since our dataset contains numerous pre-existing chronic diseases, we select two of them (diabetes and kidney disease) to showcase our model's performances when conditioned on specific disease severities, reported in Tables \ref{tb:eval_condition_pd_severity_diabetes} and \ref{tb:eval_condition_pd_severity_kidney}. Diabetes severity is determined based on HgvA1c levels \citepsec{care_2018_standards_a,king_2019_dietary_a}. Kidney disease severity is based on KIQ scores\footnote{https://wwwn.cdc.gov/Nchs/Nhanes/2013-2014/KIQ\_U\_H.htm}.

\begin{table}[h]
\centering
\caption{Evaluations Conditioned on Diabetes Severity (NHANES)}
\label{tb:eval_condition_pd_severity_diabetes}
\small
\begin{threeparttable}
\begin{tabular}{C{57pt}C{90pt}C{57pt}C{57pt}C{57pt}C{57pt}}
\toprule
 HgbA1c & Diabetes Severity  & F1-score & Precision & Recall \\ \midrule
 $<5.7$ & No diabetes  & $0.770 \pm 0.022$ & $0.750 \pm 0.040$ & $0.793 \pm 0.016$ \\
 $5.7-6.4$ & Pre-diabetes  & $0.796 \pm 0.022$ & $0.792 \pm 0.046$ & $0.802 \pm 0.020$  \\
 $6.5-9$ & Diabetes  &  $0.762 \pm 0.047$ & $0.734 \pm 0.054$ & $0.795 \pm 0.068$ \\
 $>9$ & Severe diabetes  &  $0.789 \pm 0.091$ & $0.801 \pm 0.025$ & $0.797 \pm 0.181$  \\
 \bottomrule
\end{tabular}
\end{threeparttable}
\end{table}

\begin{table}[h]
\centering
\caption{Evaluations Conditioned on Kidney Disease Severity (NHANES)}
\label{tb:eval_condition_pd_severity_kidney}
\small
\begin{threeparttable}
\begin{tabular}{C{57pt}C{57pt}C{57pt}C{57pt}C{57pt}}
\toprule
 KIQ Score &  F1-score & Precision & Recall \\ \midrule
 $1-2$ &   $0.763 \pm 0.040$ & $0.757 \pm 0.081$ & $0.773 \pm 0.023$  \\
 $3-6$ &   $0.809 \pm 0.037$ & $0.807 \pm 0.061$ & $0.813 \pm 0.018$  \\
 \bottomrule
\end{tabular}
\end{threeparttable}
\end{table}

\section{Second Dataset (mPower) Results}\label{apd:mpower}

\subsection{Data Collection and Preprocessing}  \label{sec:em_ana:data}

For generalizability considerations, we have obtained the second dataset: mPower, a smartphone-based study that collects daily motion sensor signals for chronic disease patients \citepsec{bot_mpower_2016_a}. To acquire the depression label, we leverage the MDS-UPDRS survey from this dataset. The MDS-UPDRS survey is originally used to evaluate Parkinson's disease severity. Part of its questions overlaps with the PHQ-9 depression assessment questionnaire. We select those overlapped questions to measure depression status, including MDS-UPDRS 1.3-1.5 and 1.7-1.8, whose total score is 20. In clinical practice, patients with a PHQ-9 score over 4 (total score is 27) are diagnosed as depressed \citepsec{patient_phq-9_2022_a}. Similarly, we label patients whose MDS-UPDRS score is over $20 \times 4/27 \approx 3$ as depressed and the remaining as non-depressed. Since the MDS-UPDRS is a crude depression screening measure, this is to predict depressed mood rather than diagnose depression. Our data usage (depression detection using MDS-UPDRS and sensor) has been approved by Synapse and our institute's IRB. The walking tests in the mPower dataset were done individually and unsupervised at home. The participants downloaded an app on their mobile phones. The app gives participants instructions to walk. The app automatically records the walking data. Only the mobile phone was used. No other equipment was necessary. This test setting is the norm in sensor-based disease monitoring, as is widely adopted by the health sensing studies in Table \ref{tb:healthsensing}.

To construct our input data, we utilize the accelerometer data from the mPower dataset. These data are collected from walking tests – each test is composed of walking 20 steps in a straight line (outbound), turning around and standing for 30 seconds (rest), and walking 20 steps back (return). In the walking tests, the accelerometer records a tri-axial acceleration reading
sampled at a frequency of 100 Hz. To reduce noise and prevent overfitting, we follow the standard sensor data preprocessing technique \citepsec{sigcha_deep_2020_a} to resample the readings at a frequency of 10 Hz.
For each patient, we select a window of two weeks and utilize the accelerometer data in this time window as the sensor input for this patient. 
Unlike the 7-day time window in the NHANES analysis which is recommended by American Psychiatric Association as the depression severity assessment window, we choose the two-week time window in the mPower dataset, because the walking tests are conducted voluntarily by users and thus not as dense as the continuously recorded NHANES dataset. We use a relatively longer time window to include sufficient walking tests for model training. This time window is also not too long to be irrelevant to the current depression status.
We only select the patients with at least one chronic disease based on their answers to two questions in the demographic survey: ``Have you been diagnosed by a medical professional with Parkinson disease?'' and ``Has a doctor ever told you that you have any of the following conditions? Please check all that apply.'' Among them, we also remove the patients who did not participate in the walking experiments (no walking sensor data). In the end, we generated a dataset of 3,154 walking tests, encompassing 916 chronic disease patients (496 depressed and 420 non-depressed). Each walking test includes a sequence of motion sensor readings. Due to the complexity and high budget of sensor data collection, our data size is in line with or larger than most sensor-based disease prediction studies \citepsec{zhu_deep_2021_a,jacobson_passive_2020_a,farhan_behavior_2016_a,moon_classification_2020_a,coelln_quantitative_2019_a}. We split this dataset into 60\% for training, 20\% for validation, and 20\% for test.

\subsection{Depression Prediction Evaluation} \label{sec:ap:em_ana:eval}

We first compare with the commonly used machine and deep learning models in sensing studies.
Compared to the best deep learning model (RNN), our model increases F1-score by 0.074. This increase is attributed to our model's capability of capturing temporal symptom progression and depression symptoms. Compared to the leading feature-based ML model (XGBoost), TempPNet boosts F1-score by 0.113. This performance enhancement is due to our model's ability to learn effective features from the raw sensor signal.

\begin{table}[h]
\centering
\caption{Prediction Performance Comparison with Machine and Deep Learning Methods (mPower)}
\label{tb:pred_eval_ml}
\small
\begin{threeparttable}
\begin{tabular}{L{80pt}C{50pt}C{50pt}C{57pt}C{57pt}C{57pt}C{57pt}}
\toprule
 Model & Input & Interpretable & F1-score & Precision & Recall \\ \midrule
 TempPNet (Ours) & Raw sensor & Yes & $0.774 \pm 0.019$ & $0.805 \pm 0.035$ & $0.746 \pm 0.039$  \\
 CNN & Raw sensor & No & $0.689 \pm 0.015$ & $0.594 \pm 0.030$ & $0.821 \pm 0.047$  \\
 RNN & Raw sensor & No & $0.700 \pm 0.017$ & $0.630 \pm 0.041$ & $0.790 \pm 0.033$  \\
 KNN & Features & No & $0.500 \pm 0.000$ & $0.500 \pm 0.000$ & $0.500 \pm 0.000$  \\
 SVM & Features & No & $0.627 \pm 0.000$ & $0.512 \pm 0.000$ & $0.808 \pm 0.000$  \\
 Random forest & Features & No & $0.577 \pm 0.041$ & $0.572 \pm 0.097$ & $0.610 \pm 0.119$  \\
 AdaBoost & Features & No  & $0.615 \pm 0.000$ & $0.500 \pm 0.000$ & $0.800 \pm 0.000$  \\
 XGBoost & Features & No & $0.661 \pm 0.000$ & $0.580 \pm 0.000$ & $0.769 \pm 0.000$  \\
 \bottomrule
\end{tabular}
\end{threeparttable}
\end{table}

Compared to regular MTSC models without temporal progressions of prototypes \citepsec{ismail_2020_inceptiontime_a}, our model increases F1-score by 0.078. This result proves that capturing the prototypes and their temporal progressions assists in prediction performance. Compared to the best-performing prototype learning for MTSC model \citepsec{chen_this_2019_a}, TempPNet improves F1-score by 0.058. Such a significant performance gain indicates that capturing the temporal progressions of prototypes greatly contributes to depression prediction.

\begin{table}[h]
\centering
\caption{Prediction Performance Comparison with MTSC and Prototype Learning for MTSC (mPower)}
\label{tb:pred_eval_mtsc}
\small
\begin{threeparttable}
\begin{tabular}{L{90pt}C{57pt}C{57pt}C{57pt}C{57pt}C{57pt}C{57pt}}
\toprule
 Model  & Interpretable & Progression of Prototype & F1-score & Precision & Recall \\ \midrule
 TempPNet (Ours) & Yes & Yes & $0.774 \pm 0.019$ & $0.805 \pm 0.035$ & $0.746 \pm 0.039$  \\
 \citesec{chen_this_2019_a} & Yes & No & $0.716 \pm 0.015$ & $0.694 \pm 0.030$ & $0.741 \pm 0.047$ \\
 \citesec{ming_interpretable_2019_a} & Yes & No & $0.701 \pm 0.017$ & $0.630 \pm 0.041$ & $0.790 \pm 0.033$ \\
 \citesec{gee_2019_explaining_a} & Yes & No & $0.683 \pm 0.016$ & $0.577 \pm 0.030$ & $0.836 \pm 0.053$ \\
 \citesec{ma_2020_interpretable_a} & Yes & No & $0.704 \pm 0.023$ & $0.628 \pm 0.077$ & $0.801 \pm 0.091$ \\
 \citesec{ismail_2020_inceptiontime_a} & No & No & $0.696 \pm 0.011$ & $0.730 \pm 0.018$ & $0.737 \pm 0.018$ \\
 \bottomrule
\end{tabular}
\end{threeparttable}
\end{table}

Since our model consists of multiple critical design components, we further perform ablation studies to show their effectiveness, as reported in Table \ref{tb:ablation}. We first remove the latent trend starting time design ($t_0^{(k)}$). We also remove the trend prototype design. After removing the trend prototype, the model loses the capability of detecting temporal symptom progression. Consequently, we test two options: using the last symptom severity to predict depression and using the average symptom severity over time to predict depression. Table \ref{tb:ablation} suggests that removing any design component will significantly hamper the prediction accuracy, proving that our design choice is optimal.

\begin{table}[h]
\centering
\caption{Ablation Studies (mPower)}
\label{tb:ablation}
\small
\begin{threeparttable}
\begin{tabular}{L{250pt}C{57pt}C{57pt}C{57pt}}
\toprule
 Model & F1-score & Precision & Recall \\ \midrule
 TempPNet (Ours) & $0.774 \pm 0.019$ & $0.805 \pm 0.035$ & $0.746 \pm 0.039$  \\
 TempPNet removing offset $t_0^{(k)}$ & $0.726 \pm 0.031$ & $0.678 \pm 0.101$ & $0.816 \pm 0.106$ \\
 Remove trend prototype using last symptom severity & $0.741 \pm 0.012$ & $0.712 \pm 0.032$ & $0.775 \pm 0.041$ \\
 Remove trend prototype using average symptom severity & $0.744 \pm 0.015$ & $0.729 \pm 0.035$ & $0.763 \pm 0.033$ \\
 \bottomrule
\end{tabular}
\end{threeparttable}
\end{table}

As our model takes the sensor data from an observation window as the input, we analyze how the length of the observation window influences the prediction accuracy. We show the results of the 2-week, 4-week, 8-week, and 16-week observation windows in Table \ref{tb:analysis_observation}.
Beyond two weeks, patients may have depressive and non-depressive episodes from time to time. Thus, noisy observations arise. Therefore, we use the 2-week observation window for all the other analyses.

\begin{table}[h]
\centering
\caption{Analysis of Observation Window (Signal Frequency = 10 Hz; mPower)}
\label{tb:analysis_observation}
\small
\begin{threeparttable}
\begin{tabular}{L{100pt}C{57pt}C{57pt}C{57pt}C{57pt}}
\toprule
 Observation Window & F1-score & Precision & Recall \\ \midrule
 2 weeks & $0.774 \pm 0.019$ & $0.805 \pm 0.035$ & $0.746 \pm 0.039$ \\
 4 weeks & $0.738 \pm 0.021$ & $0.708 \pm 0.029$ & $0.772 \pm 0.039$ \\
 8 weeks & $0.730 \pm 0.027$ & $0.683 \pm 0.084$ & $0.810 \pm 0.103$ \\
 16 weeks & $0.721 \pm 0.031$ & $0.651 \pm 0.077$ & $0.828 \pm 0.087$ \\
 \bottomrule
\end{tabular}
\end{threeparttable}
\end{table}

To reduce noise in the sensor data and avoid overfitting, sensor-based prediction studies usually downsample the sensor signals \citepsec{sigcha_deep_2020_a}. We test the effect of different sample rates in Table \ref{tb:analysis_frequency}: 10 Hz, 20 Hz, and 30 Hz. The results suggest that 10 Hz signal frequency achieves the best performance. Therefore, we use the 10 Hz signal frequency for all the other analyses.

\begin{table}[h]
\centering
\caption{Analysis of Signal Frequency (mPower)}
\label{tb:analysis_frequency}
\small
\begin{threeparttable}
\begin{tabular}{L{100pt}C{57pt}C{57pt}C{57pt}C{57pt}}
\toprule
 Signal Frequency  & F1-score & Precision & Recall \\ \midrule
 10 Hz  & $0.774 \pm 0.019$ & $0.805 \pm 0.035$ & $0.746 \pm 0.039$ \\
 20 Hz  & $0.752 \pm 0.034$ & $0.674 \pm 0.075$ & $0.866 \pm 0.062$ \\
 30 Hz  & $0.725 \pm 0.034$ & $0.631 \pm 0.086$ & $0.877 \pm 0.080$ \\
 \bottomrule
\end{tabular}
\end{threeparttable}
\end{table}

PD and depression have a certain correlation because they share similar walking symptoms. To make sure that our model is actually predicting depression instead of PD severity, we perform evaluations that are conditioned on the PD severity. We divide the patients into groups based on their PD severity score (the summation of the MDS-UPDRS questions \citepsec{goetz_2008_movement_a}). Conditioned on each PD severity score, we report our model's depression prediction performance. To make sure there is sufficient data points in a group to train and test our model, we only select the groups where there are at least 20 patients. This is also in line with the health sensing studies in Table \ref{tb:healthsensing}, where most studies have more than 20 subjects. Groups smaller than that do not have enough statistical power, thus inappropriate to perform reliable evaluations. If our model indeed predicts depression, we expect that, conditioned on each PD severity score, our model's performance should remain consistently high. If our model only predicts PD severity, conditioned on a PD severity score, the performance should be very low because in this group the model has not seen different values of the outcome, thus unable to update parameters well. Table \ref{tb:eval_condition_pd_severity}'s results prove that given any PD severity score, our model is able to accurately predict depression consistently. Therefore, our model indeed predicts depression rather than PD severity.

\begin{table}[h]
\centering
\caption{Evaluations Conditioned on PD Severity (mPower)}
\label{tb:eval_condition_pd_severity}
\small
\begin{threeparttable}
\begin{tabular}{C{90pt}C{57pt}C{57pt}C{57pt}C{57pt}}
\toprule
 PD Severity Score  & F1-score & Precision & Recall \\ \midrule
 6  & $0.796 \pm 0.042$ & $0.944 \pm 0.027$ & $0.692 \pm 0.064$ \\
 7  & $0.768 \pm 0.095$ & $0.729 \pm 0.112$ & $0.817 \pm 0.084$ \\
 8  & $0.774 \pm 0.045$ & $0.772 \pm 0.032$ & $0.778 \pm 0.060$ \\
 9  & $0.837 \pm 0.044$ & $0.936 \pm 0.036$ & $0.757 \pm 0.050$ \\
 10  & $0.814 \pm 0.098$ & $0.780 \pm 0.116$ & $0.853 \pm 0.076$ \\
 11  & $0.752 \pm 0.068$ & $0.637 \pm 0.042$ & $0.917 \pm 0.126$ \\
 12  & $0.848 \pm 0.035$ & $0.779 \pm 0.012$ & $0.932 \pm 0.066$ \\
 13  & $0.844 \pm 0.038$ & $0.748 \pm 0.030$ & $0.969 \pm 0.056$ \\
 14  & $0.779 \pm 0.027$ & $0.858 \pm 0.037$ & $0.713 \pm 0.035$ \\
 15  & $0.780 \pm 0.060$ & $0.848 \pm 0.034$ & $0.722 \pm 0.094$ \\
 16  & $0.810 \pm 0.062$ & $0.931 \pm 0.108$ & $0.720 \pm 0.046$ \\
 17  & $0.845 \pm 0.137$ & $0.879 \pm 0.085$ & $0.821 \pm 0.180$ \\
 18  & $0.743 \pm 0.098$ & $0.686 \pm 0.092$ & $0.812 \pm 0.114$ \\
 \bottomrule
\end{tabular}
\end{threeparttable}
\end{table}

\subsection{Interpretation of Depression Prediction}

\begin{figure}[h]
    \centering
    \includegraphics[width=1\textwidth]{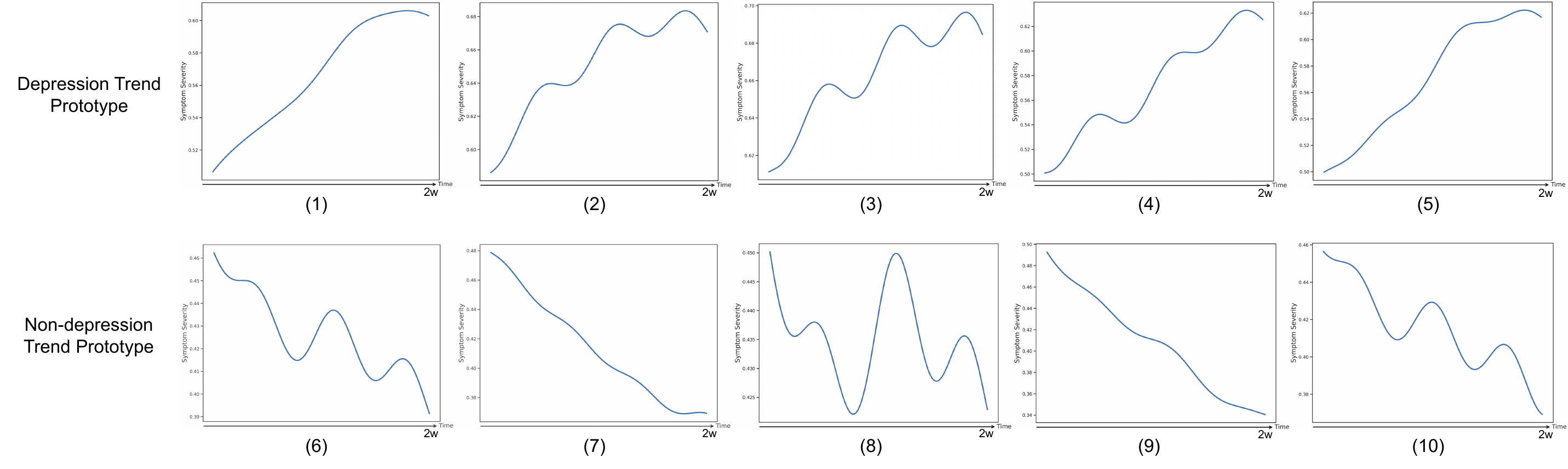}
    \caption{Trend Prototypes}
    \label{fig:trend-results}
\end{figure}

Beyond depression prediction, TempPNet is capable of interpreting why a patient is classified as depressed by presenting the contributing temporal symptom progression (trend prototype) and the corresponding walking symptom (symptom prototype). Figure \ref{fig:trend-results} shows the most salient trend prototypes that our model learned. These trend prototypes are the prototypical depression or non-depression trend. For each picture, the x-axis is time, and the y-axis is symptom severity.

\begin{figure}[h]
    \centering
    \includegraphics[width=1\textwidth]{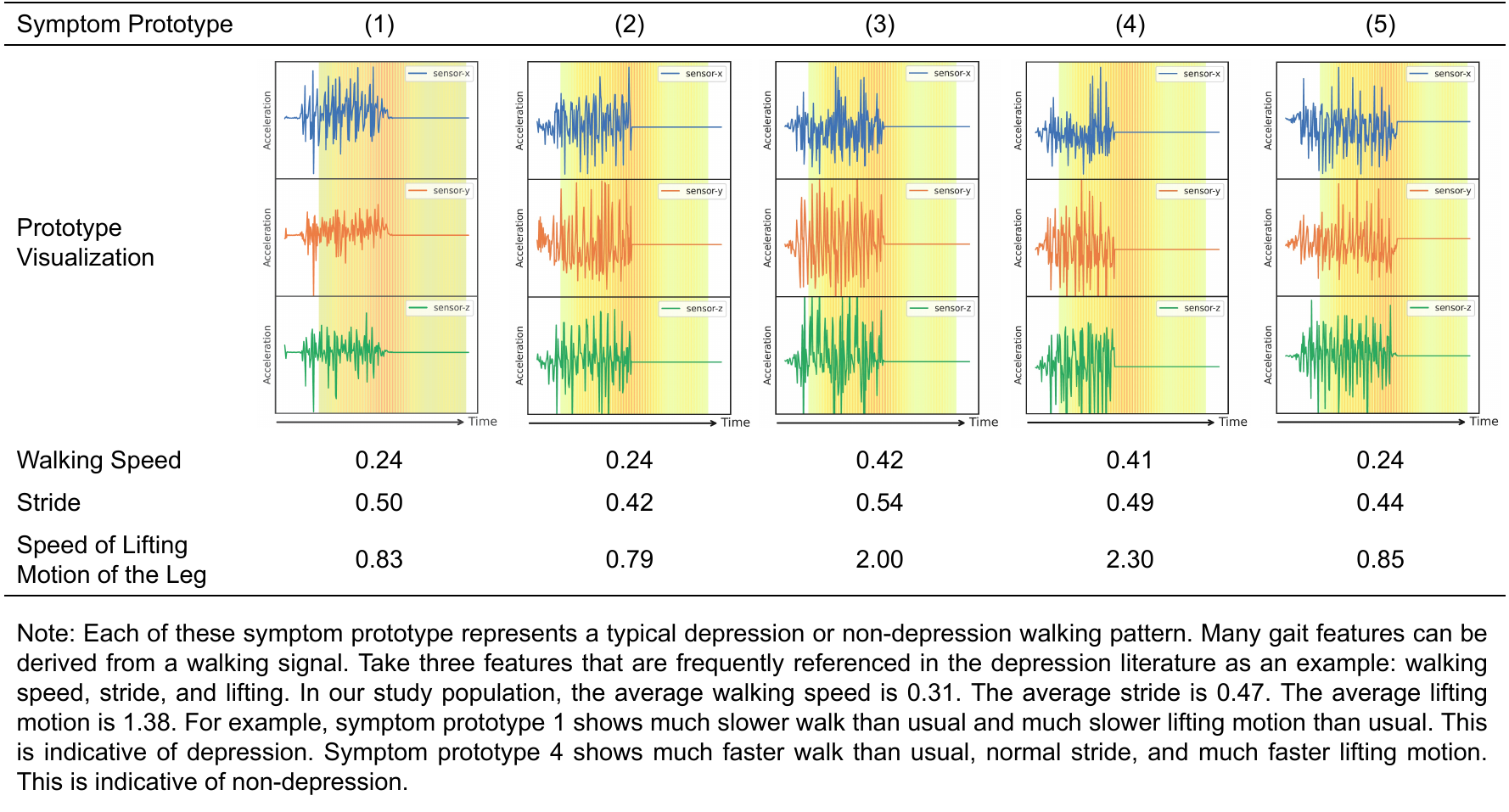}
    \caption{Symptom Prototypes}
    \label{fig:symptom-results}
\end{figure}

Trend prototypes (1)-(5) are depression trend prototypes. They represent the severities of depression symptoms trending up. Some of them have deviations from time to time in the upward trend, such as (2)-(4), representing temporary symptom relief and deterioration of depression. This conforms with the typical depression trend \citepsec{bockting_lifetime_2015_a}. Trend prototypes (6)-(10) are non-depression trend prototypes, where (6), (7), (9), and (10) represent trending down and (8) represents fluctuating with no trend. These are not typical depression trends. Each trend prototype is coupled with underlying symptoms. Figure \ref{fig:symptom-results} shows the symptom prototypes that our model learned. The trend prototypes are learned using all the patients' data rather than relying on a single patient's data. Each patient's observed walking test could be at different stages of a trend — some at the rising stage, some at the stable stage, among others. Together they depict a complete trend. Multiple patients could also share the same trend prototype if their symptom severity levels are at the same stage (e.g., all on the rise).

The prototype visualization in Figure \ref{fig:symptom-results} shows the sensor signal of the symptoms. These symptom prototypes are learned by our method across all patients. Each patient may present none, one, or more of these symptoms in their walking patterns. Prior literature suggests that depression walking symptoms can be reflected in gait features, such as walking speed, stride, and speed of the lifting motion of the leg \citepsec{sloman_gait_1982_a,lemke_spatiotemporal_2000_a,nhs_symptoms_2022_a}. When interpreting the prediction of a patient, these gait features can be computed for the symptom prototypes.

\begin{figure}[h]
    \centering
    \includegraphics[width=0.8\textwidth]{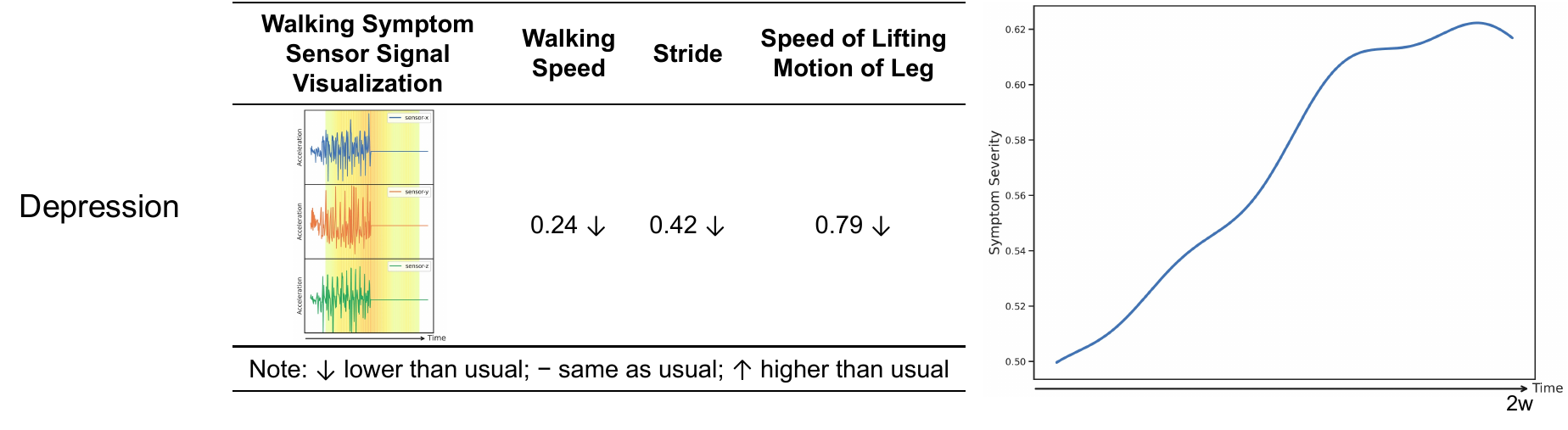}
    \caption{Interpretation of A Depressed Patient}
    \label{fig:interpret-depression}
\end{figure}

\begin{figure}[h]
    \centering
    \includegraphics[width=0.8\textwidth]{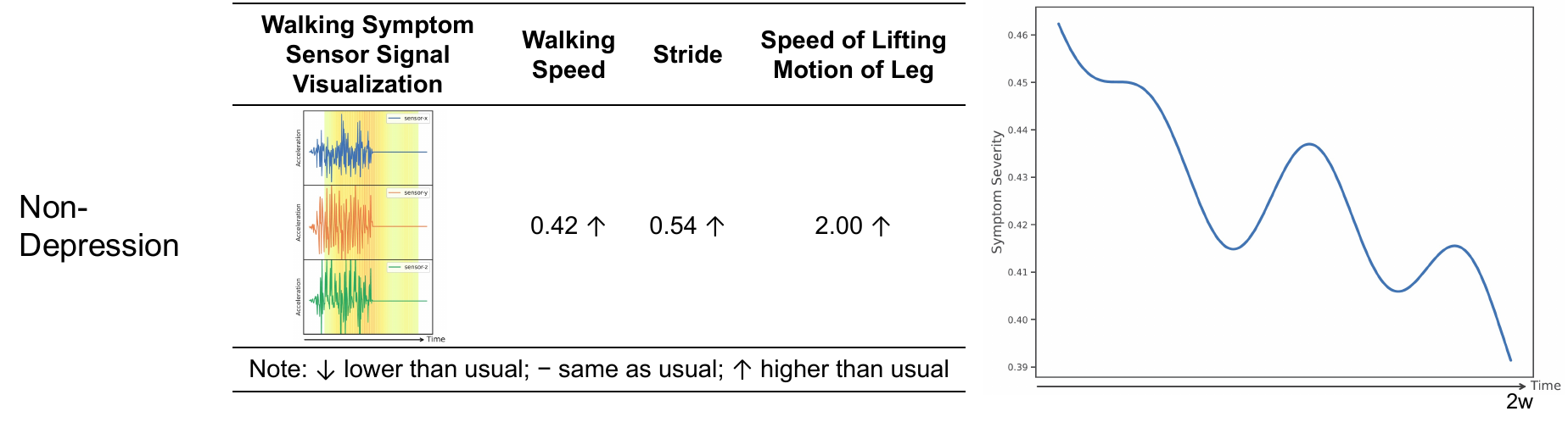}
    \caption{Interpretation of A Non-depressed Patient}
    \label{fig:interpret-nondepression}
\end{figure}

Leveraging the above-learned trend prototypes and symptom prototypes, our model can interpret the prediction of depression for every patient. We randomly select two patients (one depressed and one non-depressed) and showcase TempPNet's interpretation for them. Figure \ref{fig:interpret-depression} shows the interpretation of the depressed patient, and Figure \ref{fig:interpret-nondepression} shows the interpretation of the non-depressed patient. For simplicity, we only show the trend prototype with the highest existing strength and the corresponding symptom prototype with the highest existing strength in these examples. For the symptom prototype, we also compute the gait features using the GaitPy package\footnote{\url{https://pypi.org/project/gaitpy/}} to explain the encoded information from the visualization. The arrows after the gait features denote whether a feature is higher or lower than an average human. They do not imply any trend information (neither go up nor go down).

TempPNet predicts the patient in Figure \ref{fig:interpret-depression} as depressed for two reasons. First, this patient's walking patterns strongly present a walking symptom like in the left part of Figure \ref{fig:interpret-depression}. This walking symptom is manifested as slower-than-usual walking speed\footnote{The usual case is computed as the mean of each gait feature among the non-depressed participants in the dataset.}, shorter stride, and slower lifting motion of the leg. This symptom conforms with the depression physical symptoms in the literature \citepsec{sloman_gait_1982_a,lemke_spatiotemporal_2000_a,nhs_symptoms_2022_a}. Second, the severity of the previously mentioned symptom presents a temporal progression pattern like the right part of Figure \ref{fig:interpret-depression}. TempPNet believes this temporal symptom progression pattern resembles a typical depression progression pattern. According to the depression progression literature \citepsec{bockting_lifetime_2015_a,dattani_mental_2021_a}, this judgment makes sense — this patient's depression walking symptom first worsens rapidly and then peaks, similar to the onset and acute phases in Figure \ref{fig:depression_trend}.

TempPNet predicts the patient in Figure \ref{fig:interpret-nondepression} as non-depressed for two reasons. First, this patient's walking patterns strongly present a walking symptom like in the left part of Figure \ref{fig:interpret-nondepression}. This walking symptom is manifested as faster-than-usual walking speed, longer stride, and faster lifting motion of the leg. This symptom does not resemble the typical depression walking symptoms in the related literature \citepsec{sloman_gait_1982_a,lemke_spatiotemporal_2000_a,nhs_symptoms_2022_a}. Second, the severity of the previously mentioned symptom presents a temporal progression pattern like the right part of Figure \ref{fig:interpret-nondepression}. The symptom severity trends down and has fluctuations in the middle. This trend does not resemble a typical depression trend.

\section{Summary Statistics and Designs of Human Evaluations}\label{apd:user_study_designs}

\begin{figure}[h]
\centering
\begin{subfigure}{0.51\textwidth}
    \includegraphics[width=\textwidth]{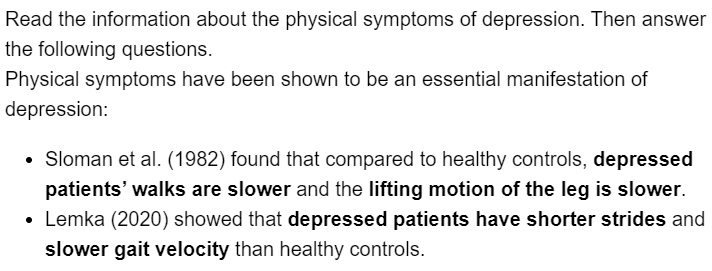}
    \caption{Knowledge Training Reading}
    \label{fig:dep_kldg_reading_nhanes}
\end{subfigure}
\hfill
\begin{subfigure}{0.37\textwidth}
    \includegraphics[width=\textwidth]{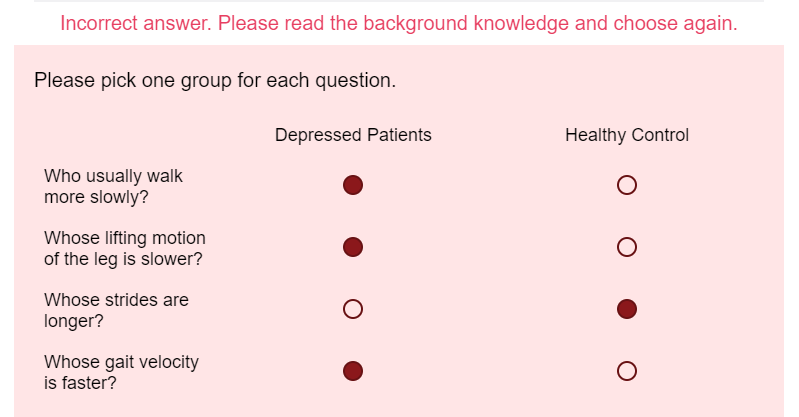}
    \caption{Knowledge Training Test}
    \label{fig:kldg_test_nhanes}
\end{subfigure}
\caption{Depression Knowledge Training}
\label{fig:dep_kldg_training}
\end{figure}

\begin{figure}[H]
\centering
\begin{subfigure}{0.59\textwidth}
    \includegraphics[width=\textwidth]{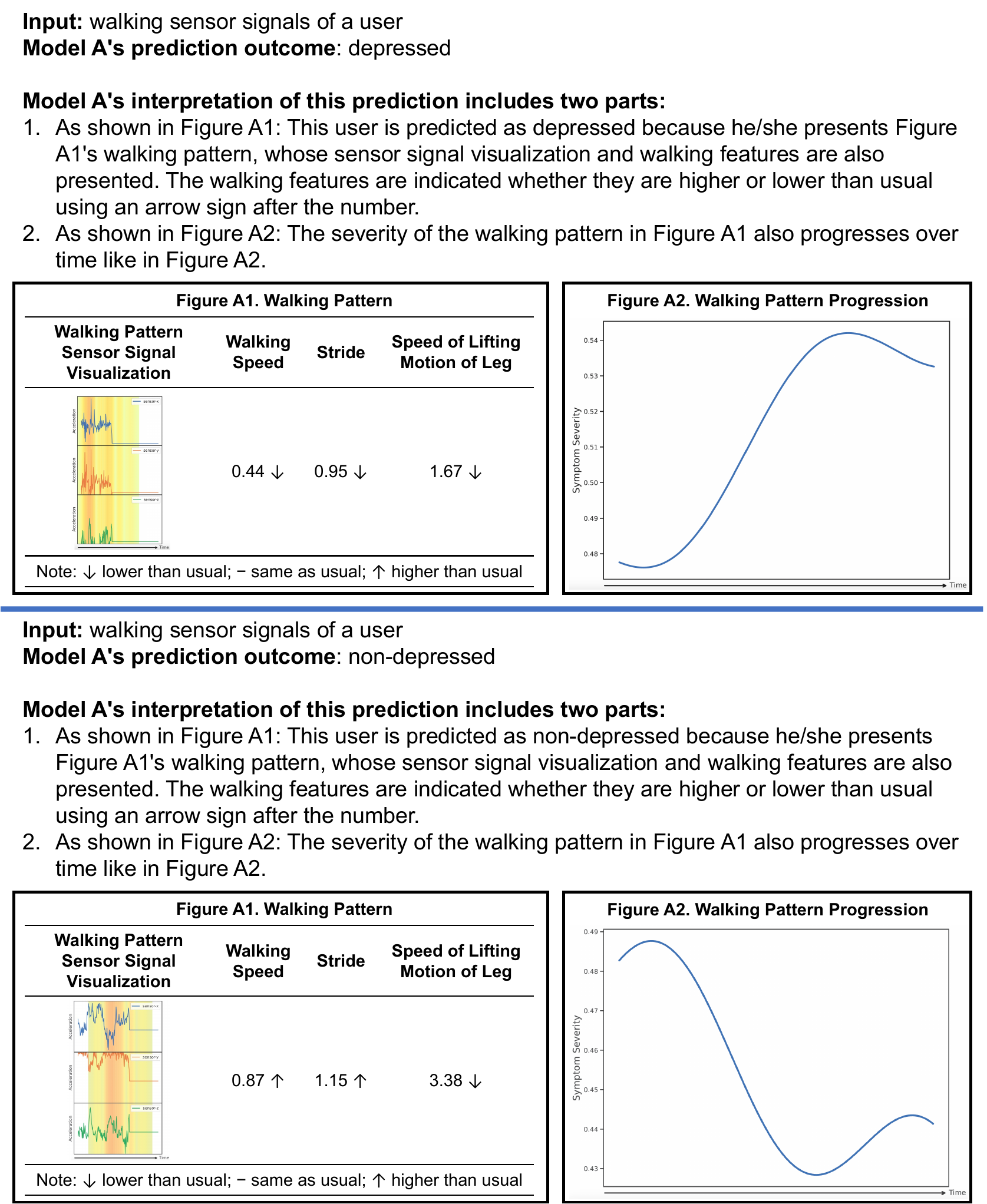}
    \caption{Group TempPNet}
    \label{fig:user-study-temppnet-nhanes}
\end{subfigure}
\begin{subfigure}{0.4\textwidth}
    \includegraphics[width=\textwidth]{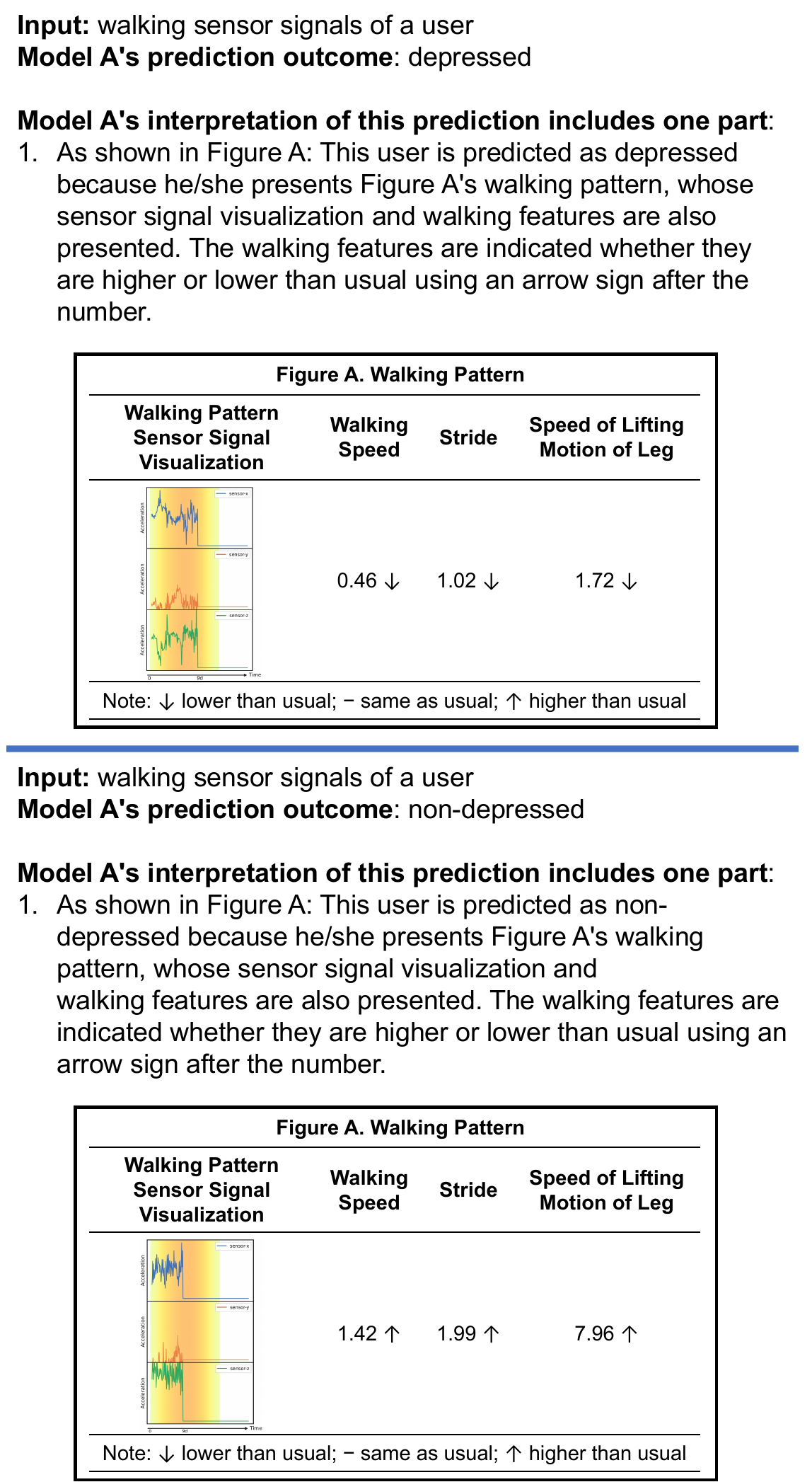}
    \caption{Group Baseline}
    \label{fig:user-study-protopnet-nhanes}
\end{subfigure}
\caption{User Study Groups}
\label{fig:user-study-groups-nhanes}
\end{figure}

The summary statistics and randomization \textit{p}-values of the user study are reported in Tables \ref{tb:summary_stats_cat_nhanes}, \ref{tb:summary_stats_con_nhanes}, and \ref{tb:randomization_check_nhanes}. The knowledge training in the user study is shown in Figure \ref{fig:dep_kldg_training}. The user study groups are shown in Figure \ref{fig:user-study-groups-nhanes}. 

\begin{table}[h]
\centering
\caption{Summary Statistics (Categorical)}
\label{tb:summary_stats_cat_nhanes}
\small
\begin{threeparttable}
\begin{tabular}{L{50pt}L{60pt}C{57pt}L{50pt}L{90pt}C{57pt}}
\toprule
 Variable & Category & Count & Variable & Category & Count \\ 
 \midrule
 Age   & 18 and lower & 1 & Education & College freshman &  2 \\
       & 18-24 & 32 &                 & College junior & 1 \\
       & 25-34 & 30 &                 & College senior & 18 \\
       & 35-44 & 2 &                  & Master & 31  \\
       & 45-54 & 1 &                  & Doctorate & 14 \\ 
Gender & Female & 39 &                &  &  \\  
       & Male & 27 &                  & & \\
 \bottomrule
\end{tabular}
\end{threeparttable}
\end{table}

\begin{table}[h]
\centering
\caption{Summary Statistics (Continuous)}
\label{tb:summary_stats_con_nhanes}
\small
\begin{threeparttable}
\begin{tabular}{L{70pt}C{40pt}C{40pt}C{40pt}C{40pt}C{40pt}C{40pt}}
\toprule
 Statistics & Min & 1st Qu. & Median & Mean & 3rd Qu. & Max \\ \midrule
 Trust in AI & 1.000 & 2.000 & 3.000 & 2.530 & 3.000 & 4.000  \\
 Health Literacy & 1.500 & 2.750 & 3.000 & 2.966 & 3.188 & 4.000 \\
 \bottomrule
\end{tabular}
\end{threeparttable}
\end{table}

\begin{table}[h]
\centering
\caption{Randomization Checks}
\label{tb:randomization_check_nhanes}
\small
\begin{threeparttable}
\begin{tabular}{L{40pt}C{40pt}C{40pt}C{40pt}C{70pt}C{70pt}}
\toprule
  & Age & Education & Gender  & Trust in AI & Health Literacy \\ \midrule
 P-value & 0.776 & 0.632 & 0.749  & 0.243 & 0.127 \\
 \bottomrule
\end{tabular}
\end{threeparttable}
\end{table}

\clearpage

\section{Implications for Other Business Areas} \label{apd:managerial_for_other_area}

\textbf{Mobile analytics:} The advent of IoT and mobile apps has enabled innovative approaches to collect granular and real-time data to assess various human behaviors \citepsec{chen_2012_business_a}. To harness such data volume and granularity, our method allows mobile business analytics researchers to systematically extract interpretable patterns for dynamic physical activities and assess user behaviors based on those patterns. For instance, driving behavior is of great interest to auto insurance companies to personalize insurance premiums. They have been using sensing technology in conjunction with mobile apps to detect careless drivers. Examples include Geico's DriveEasy, Progressive's Snapshot, and StateFarm's Drive Safe \& Save. Our method can help them detect the risk of each driver while providing an interpretation of what driving behavior and its temporal trend attribute to such driving risks. With such understanding, auto insurance companies can offer not only their customers reasons for the premium increase or decrease but also recommendations for correcting specific driving behaviors.

\textbf{Health information technology:} Our proposed method is particularly useful for those models that rely on snapshot information to make a prediction but neglect the temporal changes of such information. For example, one closely related HIT area that TempPNet can be generalized to is Parkinson's disease (PD) management. Similar to our study, wearable sensor data can be collected to reflect the motion symptoms of PD. The symptom prototype of our method can detect typical PD walking symptoms, such as smaller steps, slower speed, less trunk movement, and a narrow base of support.\footnote{\url{https://www.parkinson.org/understanding-parkinsons/symptoms/movement-symptoms/trouble-moving}} 
PD patients' symptoms may also form a trend. The trend prototype of our method is able to detect such a trend, predict PD severity, and interpret such a prediction. This new capability enables health organizations and startups to work together to manage PD symptoms timely. 
Consider another HIT area, as sensor data naturally resemble image data, TempPNet can be used to process time-series images and videos. For example, imaging results, such as X-ray and MRI, from routine doctor's visits and physicals can be processed by our model. We can pinpoint the abnormal patterns from each imaging result as well as the risky trend over time. Recent HIT studies also examine patient engagement in health education YouTube videos \citepsec{liu_2020_go_a}. The symptom prototype in our method can be adapted to recognize typical objects in each video frame, and the trend prototype can be used to capture the temporal changes of these objects, such as shape and angle changes, location moves, and context shifts. This information is essential to understand what type of information is more effective to engage patients on video platforms.

\textbf{Investment portfolio choice:} In finance and accounting, portfolio managers often rely on expected return and volatility (e.g., Sharpe Ratio) to select stocks. Our method is able to supplement this process. The time-series 10-K and 10-Q documents reveal a company's financial health, business environment, and strategic development. Apart from the commonly used accounting measures, these temporal textual data can be utilized to predict the return and volatility of stocks as well. Our method can also disclose the typical text contents that appear in these documents (e.g., certain business foci, investment areas, and competitor dynamics) and its temporal progression that attribute to a low return and high volatility prediction.

\textbf{Social media analytics:} Recent social media analytics studies in IS have investigated user behaviors such as medication nonadherence \citepsec{xie_understanding_2022_a} and emotions \citepsec{chau_2020_finding_a}. These social media data naturally form a temporal pattern where our method can play a pivotal role. For instance, a user's historical social media posts can be fed into our model to detect their emotional distress. The symptom prototype in our method can discover the typical phrases (e.g., major life events) in their posts that are mostly related to their emotional distress. The trend prototype can further depict the temporal progress of such events as well as how it contributes to the decision.

\clearpage

\bibliographystylesec{informs2014}
\bibliographysec{Project-IDDvSD-ap.bib}

\end{appendices}

\end{document}